\documentclass[pdflatex,sn-nature]{sn-jnl}

\usepackage{graphicx}%
\usepackage{multirow}%
\usepackage{amsmath,amssymb,amsfonts}%
\usepackage{amsthm}%
\usepackage{mathrsfs}%
\usepackage[title]{appendix}%
\usepackage{xcolor}%
\usepackage{textcomp}%
\usepackage{manyfoot}%
\usepackage{booktabs}%
\usepackage{algorithm}%
\usepackage{algorithmicx}%
\usepackage{algpseudocode}%
\usepackage{listings}%
\usepackage{placeins}
\usepackage{caption}
\usepackage{longtable}

\newcommand{\siTessera}{Supplementary~\ref{note:tessera}}

\newcommand{\siModelParams}{Supplementary~\ref{note:model_params}}
\newcommand{\siEmbedStruct}{Supplementary~\ref{note:crosscorr}}
\newcommand{\siSpatial}{Supplementary~\ref{note:spatialcv}}
\newcommand{\siYear}{Supplementary~\ref{note:year}}
\newcommand{\siSpectral}{Supplementary~\ref{note:spectral}}
\newcommand{\siOutliers}{Supplementary~\ref{note:higherror}}
\newcommand{\siPatch}{Supplementary~\ref{note:patchsize}}
\newcommand{\siPaired}{Supplementary~\ref{note:table1}}
\newcommand{\siTableOne}{Supplementary Table~\ref{tab:mainresults}}
\newcommand{\siFigOutliers}{Supplementary Fig.~\ref{fig:sensitivitycurve}}

\theoremstyle{thmstyleone}

\theoremstyle{thmstyletwo}

\theoremstyle{thmstylethree}

\begin{document}

\makeatletter
\let\std@addcontentsline\addcontentsline
\renewcommand{\addcontentsline}[3]{}
\makeatother

\makeatletter
\renewcommand*\l@section[2]{%
  \ifnum \c@tocdepth >\z@
    \addpenalty\@secpenalty
    \addvspace{1.0em \@plus\p@}%
    \setlength\@tempdima{2.8em}%
    \begingroup
      \parindent \z@ \rightskip \@pnumwidth
      \parfillskip -\@pnumwidth
      \leavevmode \bfseries
      \advance\leftskip\@tempdima
      \hskip -\leftskip
      #1\nobreak\hfil
      \nobreak\hb@xt@\@pnumwidth{\hss #2%
                                 \kern-\p@\kern\p@}\par
    \endgroup
  \fi}
\renewcommand*\l@subsection{\@dottedtocline{2}{2.8em}{3.5em}}
\makeatother

\title[Below-ground Fungal Biodiversity Can be Monitored Using Self-Supervised Learning Satellite Features]{Below-ground Fungal Biodiversity Can be Monitored Using Self-Supervised Learning Satellite Features}


\author*[1]{\fnm{Robin} \sur{Young}}\email{ray25@cam.ac.uk} \author[2]{\fnm{Michael E.} \sur{Van Nuland}} \author[2,3]{\fnm{E. Toby} \sur{Kiers}} \author[4]{\fnm{Tom{\'a}{\v s}} \sur{V{\v e}trovsk{\'y}}} \author[5]{\fnm{Petr} \sur{Kohout}} \author[4]{\fnm{Petr} \sur{Baldrian}} \author*[1]{\fnm{Srinivasan} \sur{Keshav}}\email{sk818@cam.ac.uk}

\affil*[1]{\orgdiv{Department of Computer Science and Technology}, \orgname{University of Cambridge}, \orgaddress{\city{Cambridge}, \state{Cambridgeshire}, \country{UK}}} \affil[2]{\orgname{Society for the Protection of Underground Networks}, \orgaddress{\city{Dover}, \state{DE}, \country{US}}} \affil[3]{\orgdiv{Amsterdam Institute for Life and Environment (A-LIFE), Section Ecology \& Evolution}, \orgname{Vrije Universiteit Amsterdam}, \orgaddress{\city{Amsterdam}, \country{The Netherlands}}} \affil[4]{\orgdiv{Institute of Microbiology}, \orgname{Czech Academy of Sciences}, \orgaddress{\street{Videnska 1083}, \city{Prague}, \postcode{14200}, \country{Czech Republic}}} \affil[5]{\orgdiv{Laboratory of Microbial Ecology and Biogeography, Institute of Microbiology}, \orgname{Czech Academy of Sciences}, \orgaddress{\street{Videnska 1083}, \city{Prague}, \postcode{14200}, \country{Czech Republic}}}


\abstract{Mycorrhizal fungi are vital to terrestrial ecosystem functioning. Yet monitoring their biodiversity at landscape scales is often unfeasible due to time and cost constraints. Current predictions suggest that 90\% of mycorrhizal diversity hotspots remain unprotected, opening questions of how to broadly and effectively map underground fungal communities. We show that self-supervised learning (SSL) applied to satellite imagery can predict below-ground ectomycorrhizal fungal richness across diverse environments. Our models explain over half the variance in species richness across $\sim$12,000 field samples spanning Europe and Asia. SSL-derived features are the most informative tested predictor group, and outperform each of the established climate, soil, and land cover baselines. We achieve a 10,000-fold increase in spatial resolution over existing techniques, moving from 1km landscape averages to 10m habitat-scale observations. As satellite observations are dynamic rather than static, this enables temporal monitoring of below-ground biodiversity at landscape scales for the first time. We apply this approach across two UK National Park woodlands, where ancient forests are predicted to have high but declining ectomycorrhizal diversity, marking key areas for further field verification. These results establish SSL satellite features as a scalable tool for extending sparse field observations to continuous, high-resolution biodiversity maps for monitoring the invisible half of terrestrial ecosystems.}




\maketitle

\section{Introduction}
\label{intro}

Mycorrhizal fungi associate with over 80\% of plant species and are fundamental to terrestrial ecosystem functioning, from carbon and nutrient cycling to the regulation of plant community composition \citep{vanderHeijden2015mycorrhizal, Brundrett2018evolutionary, Zhang2025Mycelium, Lang2023ForestSD}. Plants allocate an estimated 13 billion tons of $\mathrm{CO_2}$ to mycorrhizal networks annually, making these organisms a key entry point of carbon into soil systems \citep{Hawkins2023MycorrhizalMA}. Yet monitoring mycorrhizal fungal communities at scale remains difficult as traditional approaches depend on direct soil sampling followed by costly molecular analysis \citep{Tedersoo2022, Nilsson2019}, producing datasets that are spatially sparse and logistically demanding \citep{Davison2015GlobalAO, Basset2025AI}. Existing predictive models rely on coarse-resolution ($\geq$1km) environmental datasets such as WorldClim and SoilGrids \citep{Vtrovsk2019AMO, VanNuland2025, Mikryukov2023}, which can identify broad biogeographic patterns \citep{Guerra2022global} but cannot capture fine-scale habitat variation or temporal dynamics.

Remote sensing offers a potential solution by providing spatially continuous, high-resolution observations of the environmental conditions that shape below-ground communities. This is because there are above-ground vegetation properties that are directly measurable by optical and radar sensors, and these properties are intrinsically tied to below-ground processes through plant-fungal interactions \citep{CastroSanchez2024TreeAM, Li2015AbovegroundbelowgroundBL, CavenderBares2021RemotelyDA, DeLuca2024MycorrhizalFR, VzquezSantos2024ArbuscularMF}. Previous studies have shown that canopy spectral properties can distinguish between mycorrhizal categories \citep{Fisher2016TreeMA}, correlate with fungal richness \citep{Thers2017LidarderivedVA, Lang2023ForestSD}, and that plant genotype signatures detectable from space track below-ground microbial community composition \citep{Madritch2014ImagingSL}. However, leveraging this approach at scale faces technical challenges, including cloud contamination of optical imagery \citep{Torresani2024ReviewingTS, Lisaius2024UsingBT, Chu2021LongTN}, the need to extract subtle temporal dynamics, and the difficulty of linking above-ground spectral signatures to complex below-ground ecological processes \citep{Gholizadeh2018RemoteSO}. The recent emergence of self-supervised learning (SSL) geospatial foundation models capable of learning rich representations directly from vast quantities of unlabeled, cloud-corrupted, multimodal satellite time series \citep{bommasani2022opportunitiesrisksfoundationmodels, Jakubik2023FoundationMF, Wang2022SSL, Lisaius2024UsingBT} offers new opportunities to address these challenges. Yet the application of SSL models to biodiversity prediction, particularly below-ground, remains largely unexplored.

Prior remote-sensing work on below-ground biodiversity has established that above-ground optical and structural signals carry information about fungal communities. Past work suggests that canopy spectra can separate the major two mycorrhizal types \citep{Fisher2016TreeMA, Thers2017LidarderivedVA, Lang2023ForestSD}, and imaging spectroscopy can track below-ground microbial composition through plant genotype \citep{Madritch2014ImagingSL}. Those studies establish that relationships exist, but are often restricted to small extents, to single sensors or airborne campaigns, and to hand-chosen indices. Separately, large-scale prediction of mycorrhizal richness has been demonstrated from 1km climate and soil surfaces \citep{VanNuland2025}, at a resolution set by those surfaces rather than by the ecology. What has not been tested if whether the two can be combined, whether a learned representation of the full multi-modal satellite time series predicts below-ground richness at continental extent and at the 10m resolution of the underlying imagery, without hand-designed features and without an airborne campaign.

In this study, we present the first application of SSL geospatial foundation models to predict below-ground fungal biodiversity from satellite time series. We focus on ectomycorrhizal fungi (EcM), which form root symbioses with trees across temperate and boreal forests \citep{Smith2010MycorrhizalSymbiosis, Brundrett2009MycorrhizalAssociations} and represent one of the most ecologically consequential mycorrhizal guilds. Approximately 60\% of all tree stems on Earth form EcM relationships \citep{Steidinger2019}, channelling $\sim$9$\mathrm{GtCO_2eyr^{-1}}$ into fungal systems \citep{Hawkins2023MycorrhizalMA}. These symbioses may be increasingly vulnerable to climate change \citep{Steidinger2020, VanNuland2024}, yet 90\% of current EcM diversity hotspots are unprotected \citep{VanNuland2025}. EcM fungi are also well-suited to remote sensing approaches: their effects on plant physiology, nutrient status, and stress tolerance manifest in detectable changes to vegetation spectral properties and phenological dynamics \citep{Wang2018OpticalDiversity, Schweiger2018PlantSpectralDiversity}.

We use Tessera, a Barlow Twins-based SSL model designed for cloud-corrupted satellite time series \citep{feng2025tesseratemporalembeddingssurface}, to extract features from Sentinel-1 and Sentinel-2 imagery, and predict EcM fungal richness using $\sim$12,000 field samples from the GlobalFungi database \citep{Vetrovsky2020GlobalFungi} spanning Europe and Asia. We test the hypothesis that the orders-of-magnitude increase in resolution afforded by SSL features from 10m satellite imagery provides a stronger predictive signal for mycorrhizal biodiversity than the coarser, static datasets that have traditionally dominated the field. Recent work demonstrated that EcM richness can be modeled from 1km climate and soil data \citep{VanNuland2025}; we show that SSL features achieve comparable predictive performance with two key advantages. First, a 10,000-fold increase in spatial resolution, moving from landscape-level averages to habitat-scale heterogeneity. Second, the elimination of manual feature engineering: a single learned representation from raw satellite observations carries enough information to model ectomycorrhizal fungi biodiversity, without the acquisition and alignment that would be required from manually crafting features from several climate, soil, and land cover datasets.

\begin{figure*}[ht]
\includegraphics[width=\textwidth, height=\textheight, keepaspectratio]{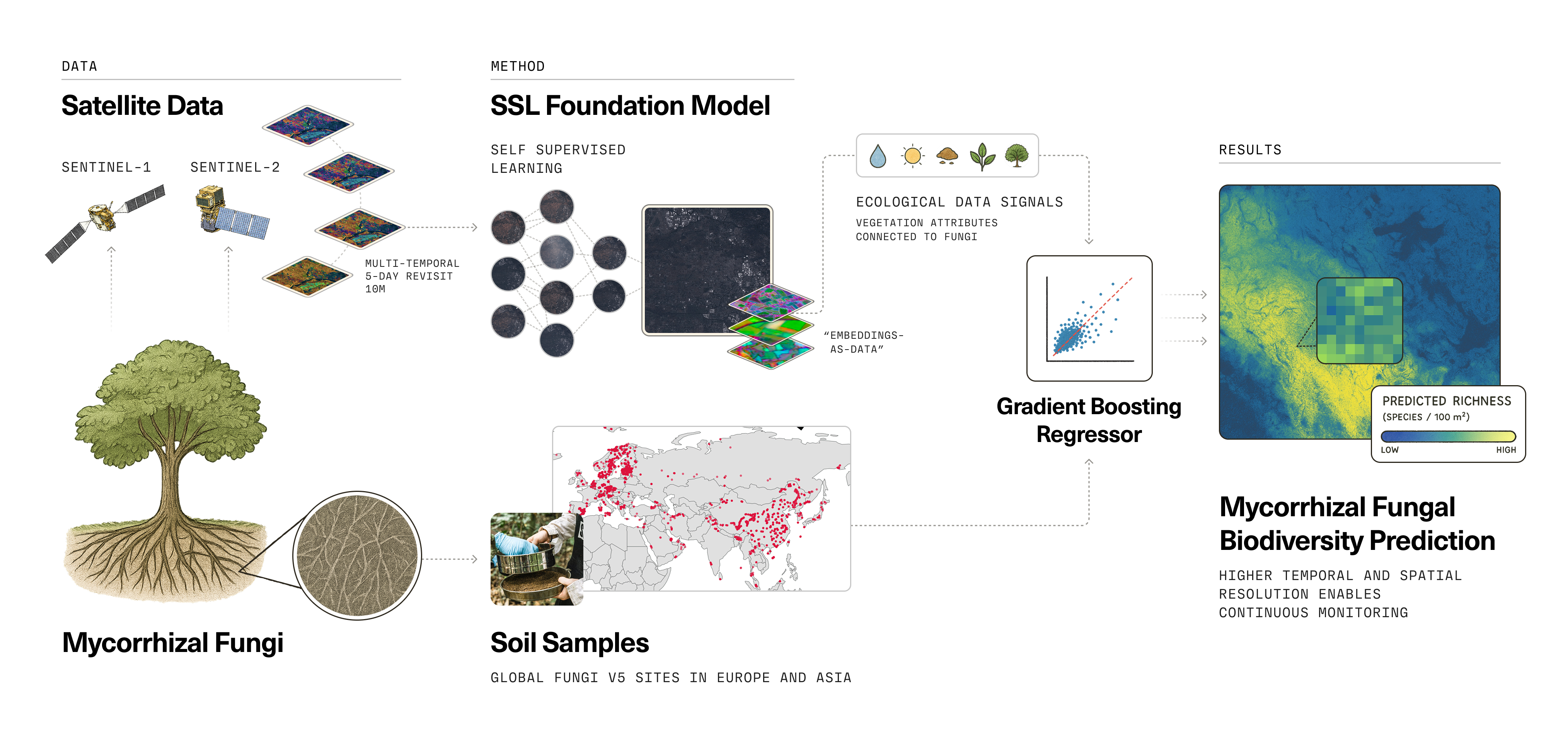}
\caption{Overview of the satellite-to-fungi biodiversity prediction framework. Ground-truth ectomycorrhizal fungal richness estimates, derived from metabarcoding of soil eDNA samples across Europe and Asia, are paired with co-located satellite time series from Sentinel-1 (radar) and Sentinel-2 (optical) imagery. A self-supervised learning foundation model (Tessera) processes the multi-temporal, multi-modal satellite data to produce high-dimensional embeddings that encode vegetation dynamics, phenology, and environmental conditions without requiring hand-crafted features. These learned representations, along with optional traditional environmental predictors (climate, soil, land cover), are used to train a gradient boosting model to predict fungal species richness. The output is a continuous, high-resolution prediction of below-ground fungal species richness at habitat scale (10m).}
\label{fig:conceptfigure}
\end{figure*}

We systematically evaluate our approach through feature set ablations comparing SSL-derived features against traditional environmental predictors, analyze geographic patterns of prediction errors to identify where and why the model succeeds or fails, and demonstrate a novel temporal monitoring capability by tracking multi-year trends in predicted EcM richness across UK National Park woodlands. An overview of our approach is presented in Figure~\ref{fig:conceptfigure}.

\section{Results}
\label{results}

First, we ask: can satellite-derived features effectively predict mycorrhizal fungal richness across diverse environments? To answer this, we analyze our ground-truth data which consists of EcM fungal richness measurements from the GlobalFungi database (release version 5; \cite{Vetrovsky2020GlobalFungi}). The dataset includes approximately 12,000 soil samples across Europe and Asia, with each sample containing fungal taxa recovered using metabarcoding of soil eDNA (see \cite{Vetrovsky2020GlobalFungi} and the \href{www.globalfungi.com}{GlobalFungi website} for further details on compiling fungal sequences and bioinformatic methods.) Here, we focus on the subset of soil fungi assigned to the EcM lifestyle using the FungalTraits database \citep{Polme2020FungalTraits}. For each sample, we estimated species richness (i.e. the number of unique species present in a sample) using a CHAO rarefaction and extrapolation approach (see Methods), with ``species'' here defined using a 97 percent sequence similarity cutoff. Following previous modeling efforts that pointed out data quality and reporting issues \citep{VanNuland2025}, we filtered samples by biome from RESOLVE Ecoregions \citep{Dinerstein2017} by removing estimated richness values that were more than five times the interquartile range higher than the biome-level median estimate. The final soil fungal dataset contained ECM fungal richness estimates ranging from 0--579.

\subsection*{Satellite Data and SSL Feature Performance}
\label{data}

\begin{figure*}[ht]
\includegraphics[width=\textwidth]{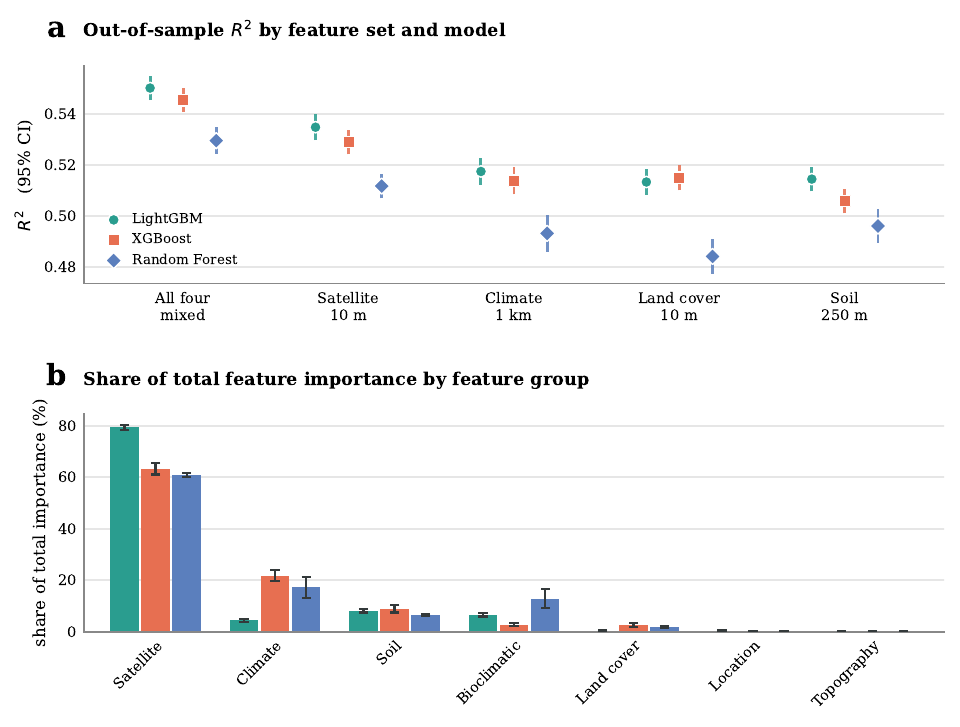}
\caption{Predictive performance and feature importance across models and feature sets. \textbf{(a)} Out-of-sample $R^2$ for each feature set, by model. Points are the mean over 50 runs, error bars are 95\% confidence intervals. \textbf{(b)} Share of total feature importance carried by each feature group, as mean $\pm$ 1 standard deviation over 50 runs.}
\label{fig:mainresults}
\end{figure*}

For each sample location we encoded a year-long time series of Sentinel-1 and Sentinel-2 satellite imagery into a high-dimensional feature vector capturing vegetation dynamics and phenological patterns. These embeddings were used to train three downstream models, LightGBM \citep{Ke2017LightGBM}, XGBoost \citep{Friedman2001GreedyFA} and Random Forest \citep{Breiman2001RF}. Full architectural and training details are given in Methods. Our SSL-based approach achieved strong predictive performance for mycorrhizal richness across the diverse study region. Figure~\ref{fig:mainresults}a summarizes out-of-sample $R^2$ for every model and feature combination across 50 cross-validation runs. The full table, with RMSE, MAE and mean error for all configurations, is given in \siTableOne. The best-performing model (LightGBM with all feature sets) achieved $R^2$ = 0.55 $\pm$ 0.017. Here, we also note the gap between the RMSE (46.22) and the MAE (28.29) (\siTableOne), which indicates that the mean error is being disproportionately influenced by a small number of large prediction failures, which can suggest a heavy-tailed error distribution \citep{Hyndman2006}. We discuss this further in Section~\ref{geospatialerror}.

\subsection*{Environmental Baseline and Feature Importance}
\label{environmental}

Then, we next asked: how do satellite features compare against established environmental predictors? We assembled a set of conventional environmental covariates spanning climate (WorldClim bioclimatic variables at 1\,km), soil (SoilGrids at 250\,m), and land cover (ESA WorldCover at 10\,m), representing the factors known to influence mycorrhizal communities \citep{VanNuland2025, Vtrovsk2019AMO}. We then tested each feature set individually as well as altogether in a full combination. All models shared the same cross-validation seeds, making comparisons paired. Full details are in methods.

Several patterns emerge from the results in terms of predictive precision and efficiency. The SSL satellite features alone outperform every single conventional data source. Because all fifteen configurations share one set of fifty seeds, these comparisons are paired. Satellite-only exceeds climate-only, soil-only and land cover-only in nine of nine model $\times$ baseline comparisons, with every 95\% confidence interval on the paired difference excluding zero (\siPaired). The improvement gained by combining SSL features with conventional predictors is modest ($R^2$ increase of $\sim$0.015). The small increase in performance is likely attributed to unique non-overlapping information from inductive modeling bias or additional sampling information incorporated in the creation of those datasets. Model choice has relatively minor impact compared to feature selection, with LightGBM and XGBoost showing insignificant performance differences.

Analysis of feature importance using each model's reported metrics revealed that SSL-derived satellite features dominated model predictions, accounting for 61\% to 79\% of total feature importance (Fig.~\ref{fig:mainresults}b). Climate and bioclimatic variables together contributed 11\% to 30\% of importance, soil properties 7\% to 9\%, and land cover, geographic coordinates and topography under 3\% each.

Explicit land cover features contributed minimal predictive power (under 3\% importance), despite land cover being a useful predictor of fungal communities in previous studies \citep{Dai2013, Barcelo2019, Labouyrie2023}, and indeed as a standalone feature set in Figure~\ref{fig:mainresults}a. We hypothesize this reflects information overlap, where the SSL model implicitly learns to distinguish land cover types through analysis of temporal spectral dynamics, making explicit land cover classifications less important with respect to the downstream task model as a different modeled product that is also based on Sentinel-2. This could also suggest that the self-supervised learning approach is more information rich than the inductive supervised modeling of land use type, perhaps from vegetation condition and phenology heterogeneity.

One important consideration in interpreting the high importance of SSL satellite features could be the spatial resolution mismatch between data sources. SSL features are extracted at 10m resolution, while traditional environmental predictors operate at coarser resolutions: WorldClim climate data at 1km, and SoilGrids at 250m.

When multiple samples fall within the same coarse-resolution environmental pixel (e.g. multiple sample locations within a single 1km WorldClim grid cell), we can only assign identical environmental values to samples that may actually experience different local conditions. In contrast, SSL features can distinguish between these samples based on fine-scale differences in vegetation dynamics, potentially capturing local environmental variation not resolved by coarse-scale datasets. This effect could explain the apparent importance of SSL features in our models. The SSL features may be more important to models because they resolve the fine-scale environmental heterogeneity that coarse-grained data cannot.

We regard this as a property of the data classes being compared rather than a confound to be removed. Resolution is not separable from the datasets themselves, and a 1km climate surface is coarse because of how it is constructed, and equalizing resolution would mean degrading the satellite features rather than improving the baselines. The like-for-like comparison that is available holds resolution fixed and varies only the representation. Using the same Sentinel-1 and Sentinel-2 imagery, the same 3$\times$3 window and the same protocol, we built a 90-feature spectral baseline of band statistics, vegetation indices, phenology terms and SAR summaries (\siSpectral). The learned embeddings still lead it by a margin comparable to the one by which they lead WorldClim. That the learned features lead a baseline built from the same imagery points to what the representation adds. The indices summarize the time series into seasonal statistics, whereas the embeddings encode it directly, which is the expected direction if the plant-mediated linkage to below-ground communities runs through phenology rather than through any single-date spectral property.

We also examined the latent structure of the SSL embeddings using linear (PCA) dimensionality reduction and their correlations with the WorldClim and SoilGrids variables. Several leading components correlate with established environmental gradients. The variance explained per component and the correlations are discussed in \siEmbedStruct. We interpret these as descriptive suggestions, as what a self-supervised representation encodes is a hard open question for machine learning in general.

\subsection*{Analysis of Bias}

A key question is whether satellite and environmental features show different error patterns. A detailed analysis of model bias, as measured by Mean Error (ME) in \siTableOne, reveals a qualitative difference between the satellite-derived features and environmental predictors. While the overall explanatory power ($R^2$) of the single-feature models was more or less comparable between satellite only and climate only, their systematic error patterns were not.

Boosting models trained exclusively on satellite-derived SSL features were nearly unbiased. For instance, the LightGBM SSL-only model yielded ME values of 0.03 $\pm$ 0.96. These values, centered near zero, indicate that prediction errors were symmetrically distributed, with no systematic tendency to over or underestimate fungal richness across the dataset. This stands in contrast to models trained on environmental data. The climate-only models consistently exhibited a small but consistent negative bias, as did the soil-only models.

This asymmetry in bias may point to a difference in the information content of the data sources. Traditional predictors like WorldClim and SoilGrids operate at coarse spatial resolutions (1km and 250m), providing a homogenized, time-averaged view of the landscape. While effective at capturing broad biogeographic gradients that set a baseline for biodiversity, they may be blind to the fine-grained environmental heterogeneity that can create localized hotspots of high richness. By averaging environmental conditions, they could be effectively erasing the peaks, or within-cell outliers, the model needs to predict, leading to a functional bias. On the other hand, the SSL features are learned from a dynamic time-series of high-resolution (10m) satellite imagery. These features implicitly encode fine-scale information about vegetation phenology, micro-topography, soil moisture dynamics, and disturbance history. This allows the model to identify both the broad constraints on richness and the specific, localized conditions that contribute to it.

\begin{figure*}[ht]
\includegraphics[width=\textwidth, height=\textheight, keepaspectratio]{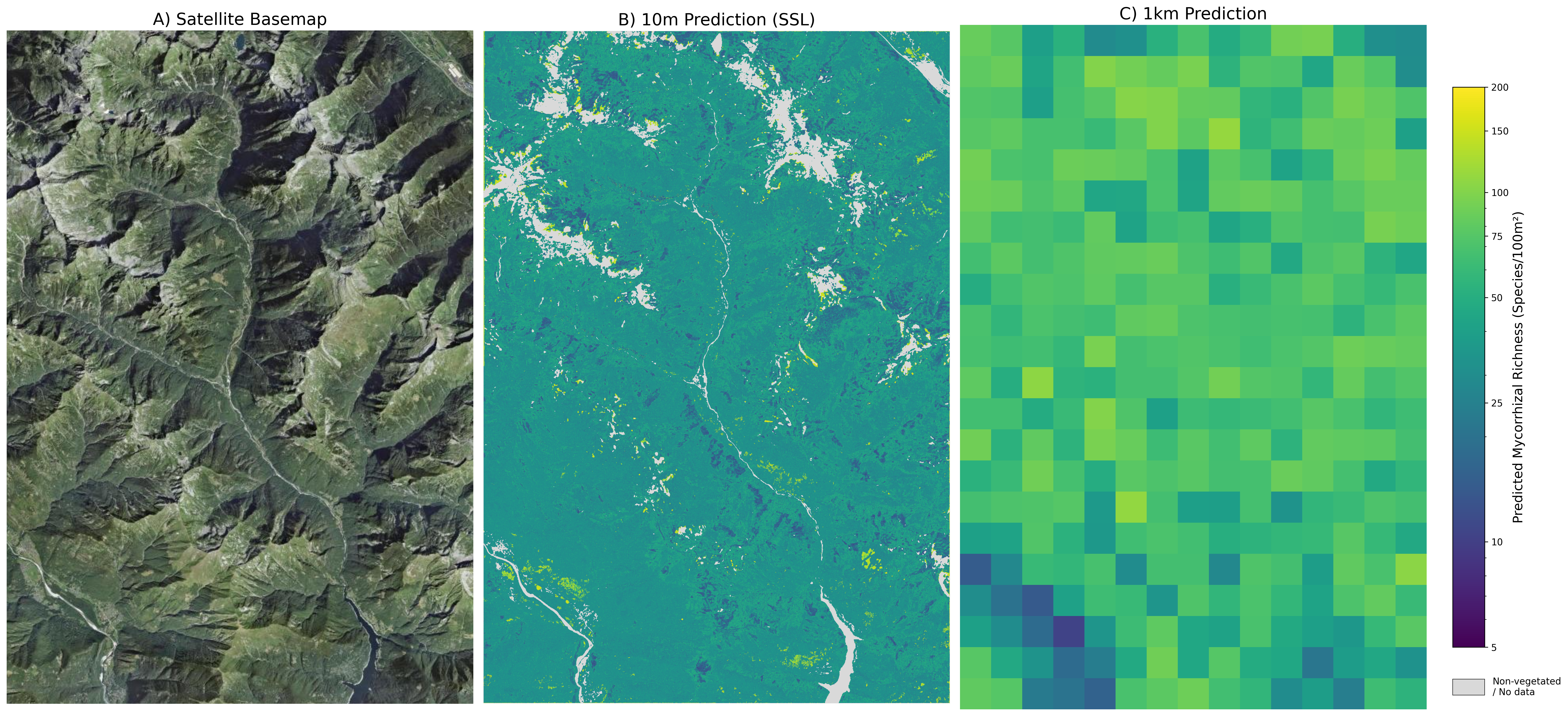}
\caption{Visual comparison of model predictions for fungal richness in a mountainous landscape. (A) Satellite basemap provides the landscape context. (B) The high-resolution (10m) prediction from the SSL model with water or glacial areas masked. (C) The coarse-resolution (1km) prediction from the climate-only model.}
\label{fig:resdiff}
\end{figure*}

To illustrate the practical implications of this higher resolution and predictive power, Figure~\ref{fig:resdiff} provides a visual comparison of the SSL and climate-only models for a topographically complex region in the Swiss Alps.

\subsection*{Error Analysis}
\label{geospatialerror}

We next asked where does the model fail and what drives prediction errors? To answer this question, we conducted spatial error analysis. Absolute prediction errors were mapped and analyzed in relation to environmental characteristics, sample density, and biogeographic factors. We identified regions of systematic model failure and investigated potential ecological explanations for poor performance.

A diagnostic analysis of the prediction errors suggests a pronounced heavy tail, indicating that a small number of samples contribute disproportionately to the overall error. While the majority of predictions exhibit relatively low errors, a small fraction of samples show extremely large prediction errors exceeding 300-400. These outliers create a pronounced right-skewed error distribution that inflates quadratic-based metrics like RMSE while having lesser impact on linear-based measures like MAE. This pattern is consistent across all algorithms, indicating systematic model limitations in extrapolating to ecologically extreme sites.

\begin{figure*}[ht]
\includegraphics[width=\textwidth, height=\textheight, keepaspectratio]{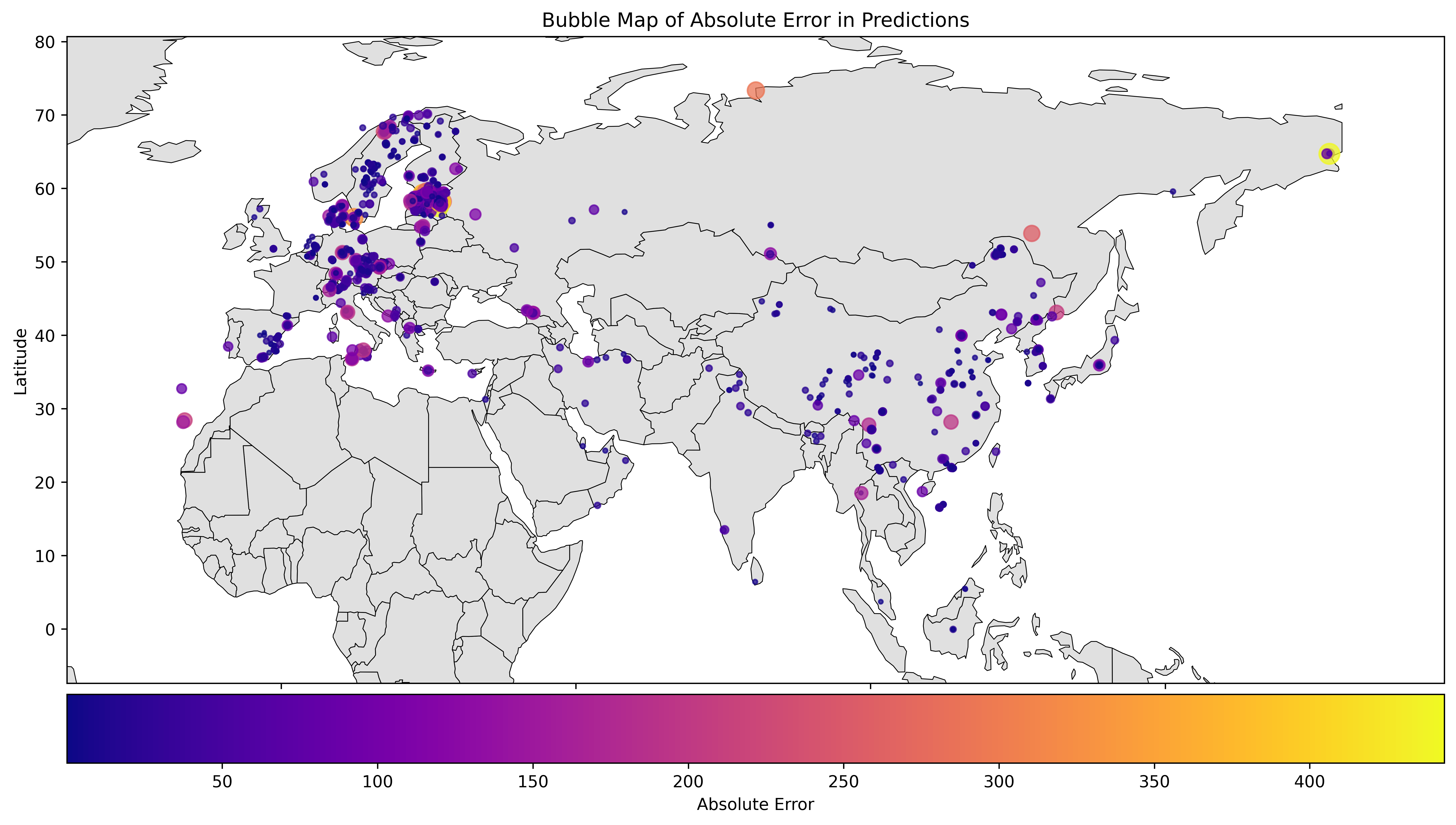}
\caption{Map showing geographic distribution of absolute prediction errors of satellite SSL only model, with point sizes and colors representing error magnitude}
\label{fig:errordistbubble}
\end{figure*}

Where are these points? As the geographic analysis shows (Fig.~\ref{fig:errordistbubble}), these are not randomly distributed; they appear to be systematically located in ecologically extreme and under-represented environments like arctic regions and seasonal wetlands. This suggests that the model struggles with out-of-distribution extrapolation.

The model performed poorly in several specific regions, notably in high-arctic regions of Siberia and isolated samples in general, where errors likely stem from sparse training data and extreme climatic constraints. A sensitivity analysis (\siFigOutliers) reveals that these failures are concentrated in a small subset of samples; removing the 2\% highest-error predictions raises the mean $R^2$ from 0.55 to 0.63, with diminishing returns thereafter. This indicates that the model performs consistently across the vast majority of conditions, and its primary weakness is a well-defined boundary condition arising from out-of-distribution environments, as corroborated by the geographic clustering of errors (Fig.~\ref{fig:errordistbubble}). The model is thus a reliable tool for temperate and well-sampled biomes, while simultaneously flagging regions where additional field sampling would be most valuable.

Characterizing that subset directly, the decile of samples with the largest absolute error carries 39.7\% of all error in the model. Those samples sit in sparser neighborhoods than the rest, a mean of 8.0 other samples within 1km against 87.3 (Cohen's $d = -0.54$), and lie further from the training distribution in embedding space ($d = +0.83$). Both effects hold whether outliers are defined by absolute error or by error relative to observed richness (\siOutliers). Flagging the decile with the lowest local sampling density recovers 1.67 times its expected share of total absolute error, and the decile most novel in embedding space 1.65 times.

\subsection*{Spatio-Temporal Analysis}

\begin{figure*}[ht]
\includegraphics[width=\textwidth, height=\textheight, keepaspectratio]{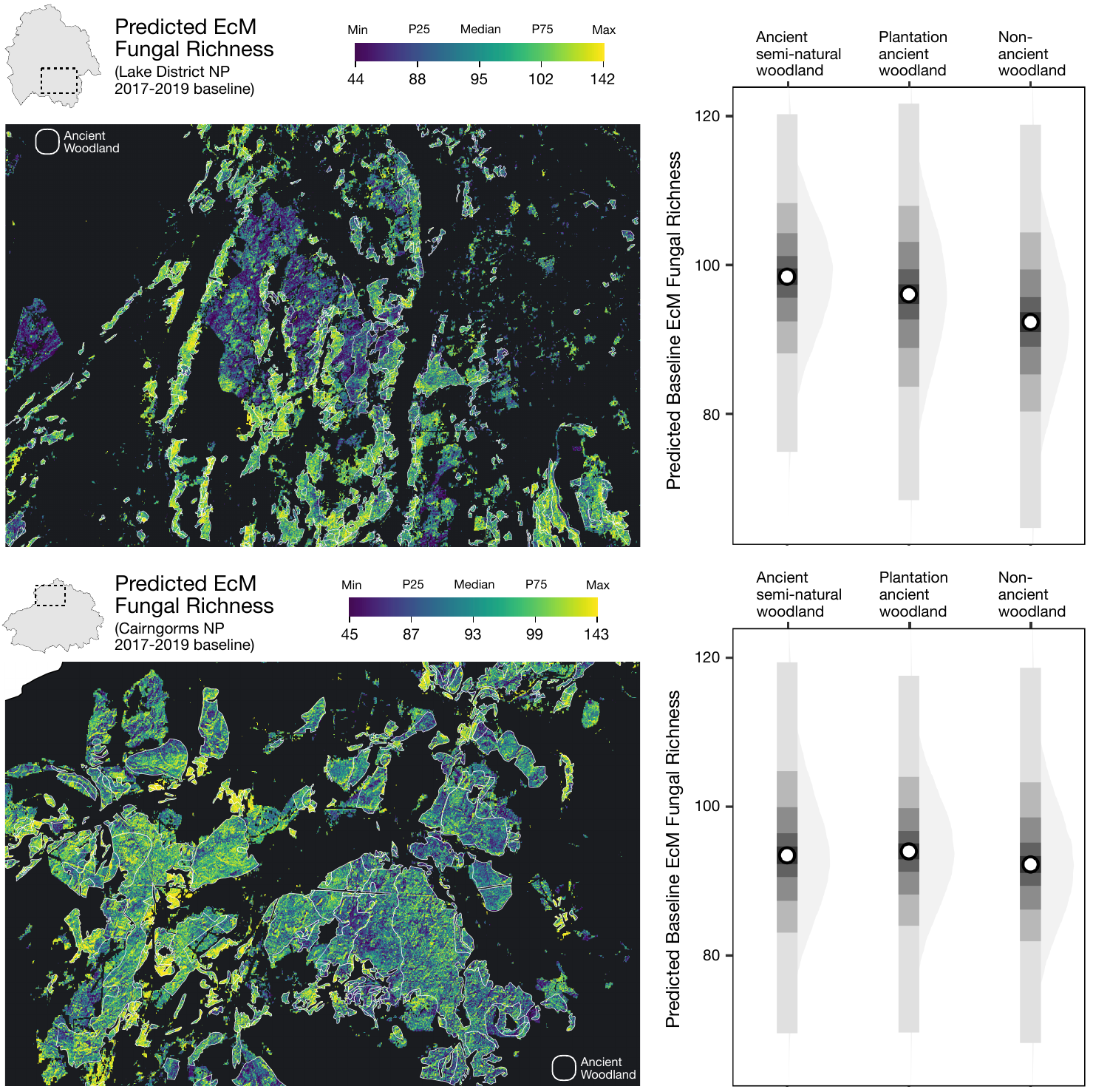}
\caption{Modeled baseline ectomycorrhizal (EcM) fungal richness. Maps show predicted baseline EcM fungal richness (2017–2019 mean) for woodland areas within the Lake District (top) and Cairngorms (bottom) National Parks. White outlines delineate designated Ancient Woodland boundaries. Distribution plots (right) show the predicted baseline richness across all forested pixels, separated by historical forest type. Grey bars indicate the central 99\%, 75\%, 50\%, 25\%, and 10\% of pixels (light to dark) and white circles show the median.}
\label{fig:temporal_baseline_map}
\end{figure*}

What new capabilities might this provide for underground biodiversity monitoring? There are two advantages compared to models built on fixed time decadal averages like WorldClim: 1) spatially continuous modeling of recent mycorrhizal diversity baselines for landscapes with no history of soil fungal sequencing data, and 2) the ability to move beyond static biodiversity maps to dynamic monitoring with annual updates against the modeled baselines (ideally with ground-truth data for validation in key areas). To demonstrate these capabilities, we conducted a spatio-temporal analysis of predicted EcM fungal diversity across forest landscapes in two UK National Parks, the Lake District and the Cairngorms. We derived a putative historical baseline of EcM fungal richness in these areas using the 2017-2019 average richness predictions in each forest pixel, which were then combined with their 8-year linear regression slopes between 2017-2024 to evaluate trends over time. For the purposes of this example, we operationally classified forested pixels into five discrete categories based on their predicted baseline richness and their temporal trajectories (high-stable, high-decreasing, intermediate/stable, low-increasing, low-decreasing). We then quantified the proportional area of these categories across ancient semi-natural woodlands, plantation ancient woodlands, and non-ancient forests using vegetation cover data from The UK Centre for Ecology and Hydrology and ancient woodland boundaries from Natural England and Nature Scot.

\begin{figure*}[htbp]
\includegraphics[width=0.85\textwidth, keepaspectratio]{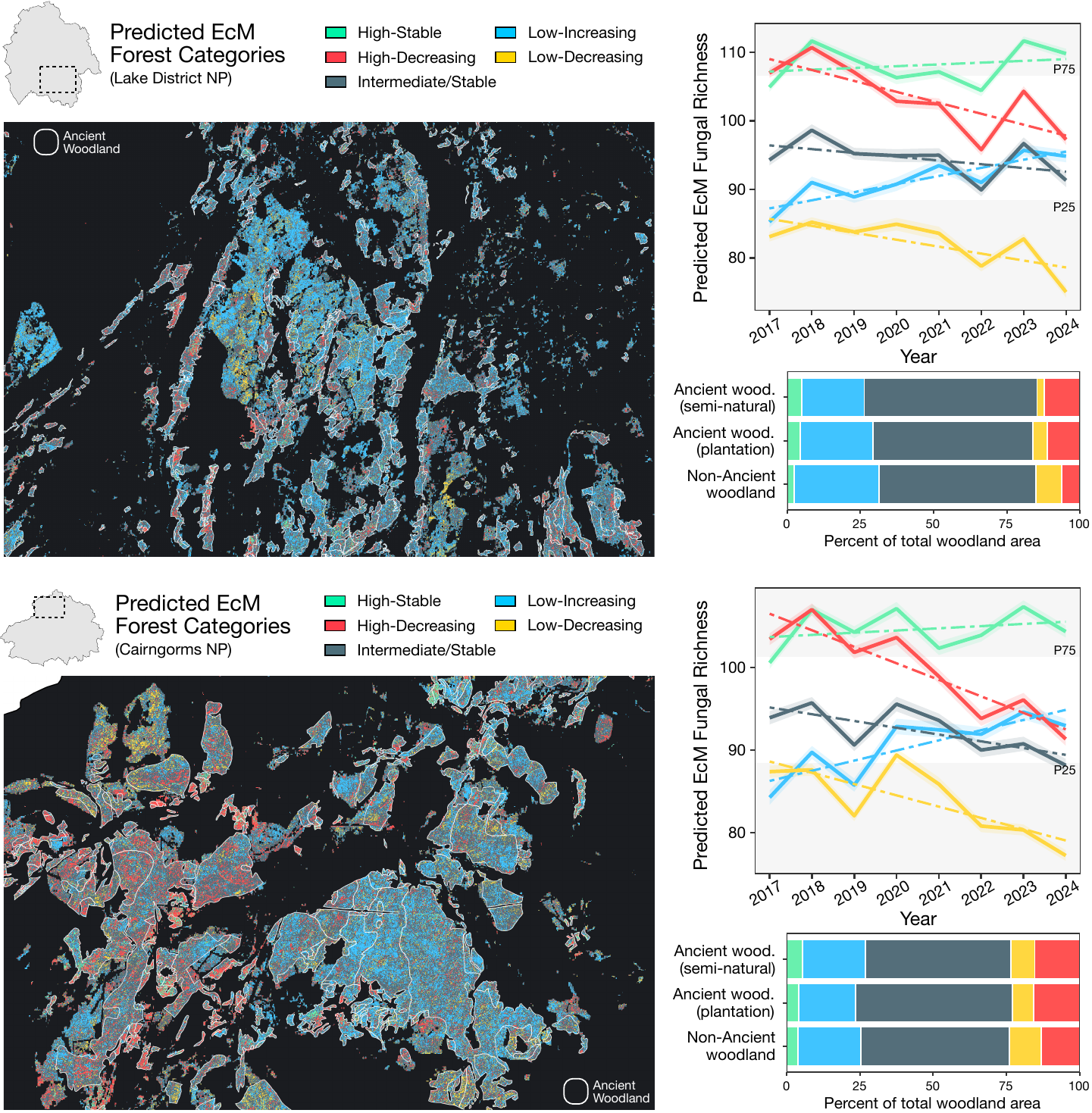}
\caption{Example spatio-temporal analysis of predicted EcM fungal richness trajectories. Maps display predicted EcM forest categories for woodland areas within the Lake District (top) and Cairngorms (bottom) National Parks. White outlines delineate designated Ancient Woodland boundaries. Forested pixels were classified into five categories based on their 2017–2019 baseline richness percentiles and 8-year regression trends in predicted richness: High-stable, High-decreasing, Low-increasing, Low-decreasing, and Intermediate/stable. Line plots track the 2017–2024 annual mean predicted richness (solid lines) and linear trends (dashed lines) for each category, with grey shading indicating values below the 25th and above the 75th baseline richness percentiles for a given park. Stacked bar charts show the proportional extent of each category, separated by historical forest type.}
\label{fig:temporal_trajectory_map}
\end{figure*}

In the Lake District, ancient semi-natural woodlands are predicted to have slightly higher baseline (2017–2019 mean) EcM fungal richness than on plantations or non-ancient forest (Fig~\ref{fig:temporal_baseline_map}). This could be because of forest composition, since the Lake District's older natural woodlands are predominantly upland oak–birch forests (95\% broadleaf, versus 58\% of non-ancient forest), and EcM community diversity also tends to increase with forest age \citep{tomao2020how, Lang2023ForestSD}. In the Cairngorms, the predicted EcM richness baselines do not differ between forest types, possibly because the plant host types are similar: these ancient woodlands are often Caledonian pinewood and non-ancient forest are mainly conifer plantations (73\% and 87\% conifer forest cover, respectively). When EcM richness baselines and trends were combined (Fig.~\ref{fig:temporal_trajectory_map}), the majority of the forested landscape (45-50\%) in both parks is considered within the intermediate/stable category (e.g. average EcM fungal richness predictions for the region with relatively stable year-over-year dynamics). However, about 20,000 hectares of forest across the two parks fell into the high-baseline diversity classes where field sampling would be most important: high-decreasing (13-22\% of forest area) or high-stable (6-13\%). Both high-baseline classes were concentrated in ancient woodland. In the Lake District, for example, 22\% of ancient semi-natural woodland was classified as high-decreasing and 13\% as high-stable, roughly double the shares in non-ancient forest (13\% and 6\%, respectively). Conversely, a substantial proportion of non-ancient forest was classified as low-increasing, with 26\% forest area in the Lake District and 20\% in the Cairngorms. Rising predicted richness in younger forest could reflect enhanced woodland creation and management efforts across the UK, such as recent tree planting initiatives and actions promoting natural regeneration (e.g. Lake District Nature Recovery Plan and Cairngorms National Park Forest Strategy), although this cannot be distinguished here from canopy maturation in existing plantations.

We note that this analysis is based on model predictions rather than observations (i.e., there is no repeated fungal sampling at the landscape scale that can validate the baseline levels and changes in EcM diversity directly). This is not uncommon in restoration, where reference conditions are often assembled from multiple lines of evidence, including predictive models, and treated as a working hypothesis to be refined with further data \citep{gann2019international}. A modeled fungal baseline offers the same kind of starting point for the majority of these forests where EcM diversity has never been measured. As a result, these predicted fungal diversity patterns are best interpreted as spatially explicit hypotheses to prioritize field sampling, such as how a high-stable forest classification might mark an important EcM fungal ``refugia'' or high-decreasing areas flagged to monitor for early signs of ecological change. By leveraging the enhanced spatial and temporal resolution of remote-sensing frameworks, this approach shows that belowground fungal diversity can be predicted continuously, at a resolution useful for forest management, and updated over time, pointing to specific places where fungal data collection and conservation action are most valuable.

\section{Discussion}
\label{discussion}

\subsection*{SSL Satellite Features as Environmental Proxies}

The predictive power of Tessera SSL features suggests that self-supervised satellite features can effectively capture environmental conditions driving mycorrhizal fungal biodiversity patterns. This finding aligns with ecological data: mycorrhizal associations influence plant physiology, nutrient status, stress tolerance, and phenological timing \citep{CastroSanchez2024TreeAM, VzquezSantos2024ArbuscularMF, DeLuca2024MycorrhizalFR, Averill2019global}, all of which can manifest in detectable changes to vegetation spectral properties and seasonal dynamics.

This learned representation offers several advantages over traditional environmental predictors. Satellite-derived features capture environmental conditions at the actual scale of ecological processes (10m resolution) rather than relying on interpolated climate surfaces or coarse soil maps. The temporal dimension of satellite data with weekly revisits could enable detection of seasonal dynamics, stress events, and phenological patterns that static environmental layers cannot capture. Then, SSL features are learned directly from data rather than based on assumptions about which environmental factors matter most, potentially revealing unexpected ecological relationships.

The minimal benefit of adding explicit environmental features suggests that a single learned representation can stand in for the process of acquiring, aligning, and processing several coarser datasets. This can potentially simplify and speed up the process for biodiversity modeling while improving predictive power.

Several ecological pathways potentially contribute to the satellite-fungal biodiversity relationship. Plant species composition, which influences mycorrhizal communities \citep{Li2015AbovegroundbelowgroundBL, Li2018SoilMB}, is potentially detectable through species-specific spectral signatures and phenological patterns. Plant health and stress status, which are influenced by mycorrhizal associations and detectable through vegetation indices or canopy spectral signatures, may provide bidirectional signals about fungal community composition \citep{Sousa2021, Fisher2016TreeMA}. The physical complexity of canopy structure has also been shown to predict EcM richness from airborne observations of forests \citep{Lang2023ForestSD}. Plant diversity and leaf functional traits can be detected from remote spectral data \citep{Wang2020, Schweiger2022}, and mycorrhizal fungi help shape the diversity/composition of plant communities and the morphology, chemistry, and physiology of plant leaves \citep{Averill2019global, Tedersoo2020}. Datasets showing that remotely-sensed plant productivity metrics can be used to predict soil microbial diversity and activity trends are accumulating \citep{Sorensen2025ExploringCH, CavenderBares2021RemotelyDA}.

Understanding these mechanisms more deeply will be important for improving model reliability and interpretability. Future work might explore the ecological interpretation of SSL features through controlled experiments, analysis of feature-environment relationships, and integration with mechanistic understanding of plant-fungal interactions.

\subsection*{Model Limitations}

While our approach achieved strong overall performance, systematic failures in specific environments provide important insights into model limitations and the broader challenges of scaling biodiversity predictions. Failures in arctic environments and isolated islands likely reflect both limited training data representation and genuine ecological differences. Arctic fungal communities may be dominated by different ecological processes (e.g. permafrost dynamics, extreme seasonality) that are not well-represented in temperate training data. Likewise island ecosystems might have endemic species and unique ecological dynamics that cannot be predicted from continental training data \citep{Delavaux2021mycorrhizal}. The characterization of high-error samples reported above supports this reading. The samples the model handles worst are both sparsely surrounded and novel with respect to the training distribution, which is what a coverage limitation looks like rather than a modeling one.

These failure modes highlight the importance of training data coverage for machine learning approaches to biodiversity prediction. Unlike mechanistic models that can potentially extrapolate based on ecological principles, data-driven models are fundamentally limited by the environmental conditions represented in training data such as Siberia and isolated islands far from other sampled locations. This limitation has important implications for global biodiversity monitoring: comprehensive geographic and environmental sampling will be essential for developing robust, transferable predictive models for global mapping efforts such as the \href{https://www.spun.earth/underground-atlas/mycorrhizal-biodiversity}{Underground Atlas}.

Unlike process-based models where the contribution of soil pH or temperature is explicit, SSL features integrate vegetation status, moisture, and topography into high-dimensional embeddings. While this allows the model to capture complex, non-linear interactions, it limits our ability to isolate individual drivers of biodiversity change. Additionally, while the model implicitly learns many environmental gradients, it lacks explicit semantic constraints. Without explicit masking, the model may generate spurious predictions in ecologically impossible locations, such as water bodies. While SSL features are powerful continuous predictors, they should be used in conjunction with standard masks (e.g. masking water and urban areas) to ensure logical consistency in final map products.

\subsection*{Future Directions}

Several research directions emerge from this work that could further advance the application of SSL approaches to below-ground biodiversity monitoring. Extending the analysis to other taxonomic groups (other mycorrhizal fungi, bacteria) would test the generality of satellite-based biodiversity prediction and identify the ecological conditions where such approaches are most effective.

Our results suggest that SSL satellite features could serve as a valuable complement to field sampling efforts in global biodiversity identifying, monitoring, and protection programs, especially in the context of mycorrhizal fungi where 90\% of current biodiversity hotspots are predicted to be unprotected \citep{VanNuland2025}. The ability to predict fungal richness with reasonable accuracy ($R^2$ $\sim$0.54 under random cross-validation, and $\sim$0.30 under spatially blocked cross-validation with a realized train--test separation of roughly 2km; \siSpatial) across diverse environments provides a means to extend sparse field observations spatially and potentially guide targeted sampling efforts. Our contribution here is validating a new feature set that is globally available at 10m rather than a model that extrapolates globally. Where field samples exist, the resolution at which below-ground richness can be modeled is no longer set by the environmental covariates, and where they do not, the binding constraint is sampling coverage rather than remote sensing.

The filtered performance analysis ($R^2$ = 0.63 after removing 2\% worst predictions) indicates that the approach is performant for typical environmental conditions, suggesting its utility for broad-scale biodiversity assessment and monitoring in well-sampled biomes. The geographic error patterns provide clear guidance on where additional field sampling would be most valuable, such as environmentally extreme regions, undersampled biomes, and areas with unique ecological dynamics such as the aformentioned Siberian Arctic or isolated islands.

While our current approach demonstrates the utility of SSL features for broad-scale biodiversity prediction, a promising avenue for future research lies in alternative venues for addressing the fundamental challenge of sparse sampling within specific regions of interest. Neural Processes (NPs) \citep{Garnelo2018CNP, kim2018attentive} represent a compelling approach for this challenge, as they are explicitly designed to learn distributions over functions from sparse context points while providing principled uncertainty quantification.

In the context of biodiversity monitoring, NPs could be conditioned on SSL features extracted from sparse sample locations within a defined region \citep{young2026interpolation}, then used to generate continuous, high-resolution predictions of fungal richness across the entire landscape. Unlike traditional interpolation methods, NPs would provide better pixel-level uncertainty estimates that could guide targeted sampling strategies, identifying locations where additional field data would be most valuable for improving model predictions. The data sparsity challenge extends beyond mycorrhizal fungi to many aspects of biodiversity monitoring, where the spatial and temporal scales of ecological processes often exceed the practical limits of field sampling \citep{Hortal2015SevenShortfalls, Chave2013PatternScaleReview, Levin1992PatternScale}.

Finally, developing mechanistically informed downstream task or continued pretraining architectures for geospatial foundation models that incorporate ecological knowledge about plant-fungal interactions could improve both performance and interpretability. Such approaches might bridge the gap between data-driven prediction and ecological understanding, leading to more robust and trustworthy biodiversity models.

\section{Methods}
\label{sec:methods}

This section provides a description of the software pipeline developed for this study. The pipeline transforms raw point-based biodiversity data into a final, model-ready feature matrix through three stages: (1) Raw ecological and satellite data acquisition; (2) Feature extraction using a pretrained self-supervised learning model; and (3) Generation of environmental baseline predictors for comparative analysis.

\subsection{Fungal Data and Richness Estimates}
\label{app:fungaldata}

Fungal occurrence records were generated by data-mining of published ITS sequencing studies collected in the GlobalFungi database \citep{Vetrovsky2020GlobalFungi}, which includes the ITS region from full-length sequences in the Global Soil Mycobiome consortium database \citep{Tedersoo2021}. For EcM occurrences used in this study, we considered only soil samples, and samples sequenced for the ITS2 region deposited in the GlobalFungi dataset, since this marker is the most represented in the database, is less biased by length variability compared to ITS1 \citep{Tedersoo2016, Yang2018}, and ITS1 samples failed technical validation in a recent fungal geospatial modeling analysis \citep{VanNuland2025}. Briefly, raw sequences and metadata from 437 ITS studies (representing the 5th release of the GlobalFungi database) were processed using a bioinformatic pipeline with sequence quality checks, extraction of the ITS2 fungal region using ITSx v1.1.2 \citep{BengtssonPalme2013}, and clustering into operational taxonomic units (OTU) at 97\% similarity level with subsequent exclusion of global singletons and chimeric sequences using USEARCH v11.0.0667 \citep{Edgar2013}. Singletons were excluded from clustering, and were later binned to the clusters formed from the previous USEARCH step. We used BLASTN search against UNITE version 8.3, released 10.5.2021 to assign taxonomy to non-singleton OTU (default BLAST parameters used). Representative sequences were considered to belong to the closest BLAST hit genera in the case of $>$92\% similarity and $>$95\% coverage, and OTUs represented by sequences with e-value $>$ e$^{-50}$ were excluded. EcM fungi were subset from the resulting OTU table with taxonomy assignments by comparing against the FungalTraits database v.1.1 \citep{Polme2020FungalTraits}.

For this analysis, we subset the GlobalFungi v5 ITS dataset to only samples with confirmed EcM fungal taxa present within Europe ($n = 18{,}322$ samples) and Asia ($n = 12{,}354$ samples) given the high sample density and coverage in these regions for comparing with spectral data. To assess the number of EcM fungal species (OTUs) within our samples, we employed analytic estimators for species richness rarefaction and extrapolation \citep{Chao2014}. This method generates species accumulation curves based on sequencing depth, allowing for the calculation of diversity estimates and 95\% confidence intervals at the curves' asymptotes via the iNEXT R package \citep{Hsieh2016}. We used the default settings of extrapolation endpoints defined as double the sequencing depth for each specific sample. This rarefaction and extrapolation framework facilitates a dependable comparison of mycorrhizal patterns across different research projects, regardless of variations in sequencing technology, primer selection, or sequencing intensity---though some fluctuations in sequencing error rates may remain. The richness estimates extrapolated in this study align with prior research on global mycorrhizal fungi diversity \citep{Toussaint2020, Mikryukov2023}, and richness values per sample proved highly stable across clustering similarity thresholds of 96\%, 97\%, and 98\% \citep{VanNuland2025}.

The GlobalFungi database aggregates studies that differ in sampling design, extraction protocol, sequencing platform and primer choice, and this heterogeneity is a recognized source of variation in the response. The residual effect of project-level protocol variation on this dataset has been examined previously \citep{VanNuland2025}. Six project-specific variables (sample area, extraction mass, sequencing platform, target region, sample type, and 37 primer sets) were tested for systematic disagreement between observed and predicted richness, and only a small number of levels failed that test and were excluded. The remaining protocol variation showed no detectable systematic bias. No study-level design variable enters any model reported here as a predictor, so whatever heterogeneity remains is carried as noise in the data rather than as structure that a model can exploit.

\subsection{Tessera Model and Feature Extraction}

We used the Tessera geospatial foundation model \citep{feng2025tesseratemporalembeddingssurface}, a Barlow Twins-based self-supervised learning architecture that learns robust representations from cloud-corrupted, multimodal satellite time series. For each fungal sample location, we extracted a year-long time series (2024) of Sentinel-1 (VV and VH polarizations) and Sentinel-2 (10 spectral bands) imagery within a 3$\times$3 pixel window (30$\times$30m at 10m resolution). The model processes these multimodal time series through dual transformer-based encoders whose outputs are fused into a 128-dimensional embedding per pixel. Full architectural details, the self-supervised training objective, and the satellite data acquisition and inference pipeline are described in \siTessera, \ref{note:patches}, \ref{note:satellite_data}, \ref{note:ssl_features}.

The spatial extent captures local environmental conditions while minimizing georeferencing uncertainties. The Tessera model processes these multimodal time series where each pixel's spectral-temporal signature is embedded into a high-dimensional feature vector that encodes complex vegetation dynamics, phenological patterns, and environmental conditions. The 3$\times$3 window is chosen to be wide enough to absorb the geolocation error of the Sentinel-1 and Sentinel-2 products and of the recorded sample coordinates, and it is of the order of a single tree crown, which is the relevant scale for an obligate symbiont of trees, while remaining narrow enough not to average across adjacent habitat. Empirically we find no statistically significant differences for pixel windows larger than 3$\times$3. (\siPatch).

Samples were collected across a range of years, while embeddings were extracted from a single year of imagery. To test whether that mismatch matters we repeated the analysis on 2022, 2023 and 2024 embeddings for the 8,139 European samples present in all three years, so that the comparison is within-sample and cannot be explained by composition. Cross-sectional skill is effectively invariant to the choice, and we find that within-year $R^2$ is 0.4905, 0.4995 and 0.5000, a spread of 0.0096 against a seed-to-seed standard deviation of 0.0171. (\siYear).

SSL embedding extraction was performed on a single NVIDIA Tesla T4 GPU. Downstream model training, evaluation, and wall-to-wall 10m mapping were conducted on a dual-socket AMD EPYC 7763 server (128 cores, 1TB RAM). The core training and evaluation pipeline runs on standard desktop hardware, and the large memory configuration was required only for large scale 10m raster generation.

To optimize computational efficiency for sparse sample data, we implemented a pixel-focused inference pipeline rather than processing entire Sentinel tiles. After processing the raw satellite data and acquiring the high-dimensional embeddings, we applied dimensionality reduction \citep{McInnes2018} to compress the 3x3x128-dimensional SSL features to 256 dimensions to reduce noise and improve downstream model training speed. The reduction is fitted inside each training split, and is applied to the satellite embeddings only.

\subsection{Baseline Environmental Predictor Features}
\label{app:baselinefeatures}

How do satellite features compare against established environmental predictors? To evaluate the relative performance, we assembled a representative suite of traditional environmental predictors representing the major abiotic factors known to influence mycorrhizal fungal communities \citep{VanNuland2025, Vtrovsk2019AMO}. Specific variables used are detailed in SI Table~\ref{tab:env_variables}.

This feature set represents the standard approach in contemporary biodiversity modeling, encompassing the major environmental variables known to structure ecological communities: climate (energy and water availability), soils (nutrient availability and growing conditions), topography (local environmental modification), land cover (habitat type and disturbance), and geography (spatial constraints and unmeasured factors). The spatial resolutions range from 10m (land cover) to 1km (climate), reflecting the different scales at which environmental conditions vary and data are available.

\textbf{Climate Variables:} We extracted variables from WorldClim 2.1 \citep{Fick2017WorldClim2}. These climatic variables from WorldClim or other similar modeled datasets are widely used in species distribution modeling and macroecological studies due to their demonstrated relationships with biodiversity patterns and their global availability at consistent spatial resolution \citep{ElithLeathwick2009}. WorldClim is a global climate dataset available at 30-arcsecond (roughly 1km) spatial resolution containing long-term climate data averages for the period 1970-2000. The climatic variables include average temperature, minimum temperature, maximum temperature, precipitation, solar radiation, water vapor pressure, and wind speed. The better known Bioclimatic variables are derived to capture biologically meaningful aspects of climate variation that influence species distributions and ecosystem processes.

\textbf{Topographic Variables:} Topography may influence mycorrhizal communities indirectly through effects on microclimate, soil development, hydrology, and plant community composition. Elevation captures broad climatic gradients and biogeographic effects, slope affects drainage and erosion patterns, and aspect influences solar radiation exposure and moisture regimes. These variables are useful in heterogeneous landscapes where local topographic effects may override broader climatic patterns. Topographic features were calculated from WorldClim's 30-arcsecond digital elevation model using standard procedures with the richdem library in Python \citep{RichDEM}. Elevation (meters above sea level) was extracted, while slope (degrees) and aspect (degrees from north) were calculated from the extracted elevation data using neighborhood analysis of elevation gradients \citep{Horn1981hill}.

\textbf{Soil Properties:} Soil pH is particularly important as it affects nutrient solubility and fungal growth \citep{Rousk2010}, while organic carbon content indicates substrate availability for decomposer communities \citep{Hodge2001}. Soil texture (clay/sand/silt ratios) influences water and nutrient retention, root penetration, and fungal hyphal growth patterns. To that end, we obtained soil data from SoilGrids 2.0 \citep{soilgrids}, a global soil dataset providing predictions at 250m resolution based on machine learning models trained on soil profile databases. We extracted chemical and physical soil properties at two depth intervals (0-5cm, 5-15cm) to capture vertical soil heterogeneity of soil pH (in $H_2O$), soil organic carbon content, clay content, sand content, silt content, and bulk density. These soil properties influence mycorrhizal fungi by affecting nutrient availability, soil structure, water retention, and chemical conditions \citep{vanderHeijden2008}.

\textbf{Land Cover Classification:} Land cover directly influences mycorrhizal communities through plant host availability, land management practices, soil disturbance regimes, and habitat connectivity. Different land cover types support distinct plant communities with specific mycorrhizal associations, while anthropogenic land uses (agriculture, urban development) can alter soil fungal communities through disturbance, fertilization, and host plant changes. Land cover classes were acquired from ESA WorldCover 2021 \citep{Zanaga2022WorldCover}, a global land cover modeled map product at 10m resolution based on Sentinel-1 and Sentinel-2 data. The classification includes 11 classes: tree cover, shrubland, grassland, cropland, built-up areas, bare/sparse vegetation, snow and ice, permanent water bodies, herbaceous wetland, mangroves, and moss and lichen. We one-hot encoded these categorical variables to create binary indicators for each land cover class.

\textbf{Geographic Coordinates:} Geographic distance often explains variation in community composition due to dispersal constraints, evolutionary history, and unmeasured environmental factors that vary spatially. Including coordinates helps distinguish between environmental filtering (species responding to local conditions) and spatial processes (species responding to geographic proximity or barriers). Raw latitude and longitude coordinates were included as continuous variables to account for unmeasured spatial gradients, dispersal limitations, and historical biogeographic effects not captured by environmental variables.

\subsection{Final Modeling}
The SSL features and environmental baseline features were combined into a final feature matrix. The ablation studies and model evaluations were then performed with the models and hyperparameters. This is described in more detail in \siModelParams.

We used LightGBM \citep{Ke2017LightGBM}, a gradient boosting model, as our primary downstream model for predicting mycorrhizal fungi richness from environmental features. Gradient boosting models are particularly well-suited for medium-sized tabular datasets like ours due to their ability to handle mixed data types, capture complex non-linear relationships, and provide interpretable feature importance estimates.

To obtain robust performance estimates and quantify variability, we employed a cross-validation procedure with 50 independent training runs using 70/10/20 train/validation/test splits with early stopping. This approach provides reliable estimates of both mean performance and variability across different data partitions. We also evaluated under spatially explicit cross-validation, with blocks and single-linkage groups sized from the published autocorrelation ranges for ectomycorrhizal communities (8.5m \citep{tedersoo2010}) and soil fungi generally (2--3m, up to 17m for large genets \citep{birch2023}). Because sampling is clustered into discrete sites, a nominal 25m block produces a median train--test separation of about 2km. Satellite-only performance under that separation is $R^2 = 0.30$, and the three spatial schemes agree within 0.020 (\siSpatial).

Our objective was to test the raw predictive power of the input features rather than achieve marginal gains through exhaustive hyperparameter optimization. We evaluated different algorithms and feature combinations to assess the relative contributions of different data sources. Our experimental design encompassed algorithm comparison across three gradient boosting and ensemble methods (LightGBM, XGBoost \citep{Friedman2001GreedyFA}, Random Forest \citep{Breiman2001RF}) to ensure robustness of findings across different modeling approaches. For all models, we used standard, default hyperparameters (see \siModelParams). Feature importance was calculated by split count for the gradient boosting models (LightGBM and XGBoost), and Mean Decrease in Impurity for Random Forest. Model performance was assessed using coefficient of determination ($R^2$), mean error (ME, indicating model bias), mean absolute error (MAE), and root mean squared error (RMSE).

Feature set ablations systematically tested the predictive power of individual data sources: (1) satellite-only using Tessera-derived SSL features, (2) climate-only using WorldClim climatic and bioclimatic variables, (3) soil-only using SoilGrids properties, and (4) land cover-only using ESA WorldCover classifications. Additionally, we evaluated the full ensemble model incorporating all feature sets: satellite, climate, soil, land cover, topography, and geographic coordinates.

This ablation study design enables direct comparison of SSL satellite features against each traditional data source individually, as well as assessment of the added value of combining SSL features with conventional environmental predictors. All models were trained and evaluated using identical cross-validation procedures to ensure fair comparison across feature combinations and algorithms.

\subsection{Spatio-Temporal Analysis Implementation}
\label{app:spatiotemporal}

\subsubsection{Spatial Masking and Preprocessing}

Annual predictions of ectomycorrhizal (EcM) fungal richness (2017–2024) were aligned and masked to forested areas within the Lake District and Cairngorms National Parks. The primary forest mask was generated using 2017 UK vegetation cover classifications, supplemented by regional Ancient Woodland Inventories (Natural England and NatureScot). This allowed every forested pixel to be categorized into one of three distinct forest cover histories: ancient semi-natural woodland (ASNW), plantation ancient woodland (PAWS), or non-ancient forest. The proportion of broadleaf versus coniferous cover within each forest history class was calculated from the same UK vegetation cover classification. Finally, to account for localized sensor artifacts and persistent cloud-cover gaps specific to the 2024 Cairngorms National Park remote-sensing data, missing values (NA) were procedurally gap-filled using corresponding 2023 pixel estimates prior to regression modeling, which resulted in 16.5\% of the approximately 55,750 total forested hectares imputed to ensure a spatially complete, gap-free time-series analysis across the entire forested area.

\subsubsection{Temporal Regression and Baseline Thresholds}

To quantify longitudinal shifts in fungal diversity, pixel-wise linear regressions were computed across the 8-year raster stack to determine the annual rate of change (slope) for every forested pixel. A historical biodiversity baseline was established for each park by calculating the mean pixel richness from 2017 to 2019. From this baseline, park-specific 25th, 50th, and 75th diversity percentiles were derived to establish thresholds for high and low baseline richness. Forested pixels were classified into five mutually exclusive categories based on a combination of their historical baseline percentiles and their 8-year regression slopes, utilizing a stability threshold of $\pm$0.5 species per year: High-decreasing = High baseline (>75th percentile) experiencing active loss (slope < -0.5); High-stable = High baseline (>75th percentile) remaining stable or gaining (slope $\geq$ -0.5); Low-increasing = Moderate-to-low baseline (<50th percentile) experiencing active recovery (slope > 0.5); Low-decreasing = Low baseline (<25th percentile) continuing to actively lose species (slope < -0.5); and Intermediate/stable = Intermediate-diversity forests (50th–75th percentiles), alongside lower-diversity forests (<50th percentile) exhibiting stable temporal dynamics (slopes between -0.5 and 0.5).

\subsubsection{Zonal Statistics and Temporal Trajectories}

To evaluate landscape-level conservation status, the proportional extent of each forest category was mathematically quantified. The absolute area of each category was first calculated by tallying the total pixel count per zone and multiplying by the spatial resolution of the satellite-derived predictions (10 m x 10 m, or 0.01 hectares per pixel). These areal estimates were subsequently stratified across the three distinct forest histories (ASNW, PAWS, and non-ancient forest) using the designated spatial masks. The relative proportion of each category was then determined by dividing its absolute area by the total spatial extent of its respective forest history class. To compare baseline conditions among forest types, the 2017–2019 mean predicted richness of all forested pixels was summarized (median and interquartile ranges) within each forest history class per park. To visualize temporal trends, up to 500 representative pixels were randomly sampled from each of the five EcM trajectory categories. Annual richness means and 95\% confidence intervals were calculated for these cohorts and plotted against the global 25th and 75th richness percentiles to contextualize underground ecosystem divergence among these categories.

\backmatter





\bmhead{Data availability} \href{https://www.worldclim.org/data/worldclim21.html}{WorldClim} \citep{Fick2017WorldClim2}, \href{https://www.isric.org/explore/soilgrids}{SoilGrids} \citep{soilgrids}, \href{https://esa-worldcover.org/en/data-access}{ESA WorldCover} \citep{Zanaga2022WorldCover}, \href{https://www.globalfungi.com}{GlobalFungi} database (release version 5; \cite{Vetrovsky2020GlobalFungi}), and \href{https://ecoregions.appspot.com/}{RESOLVE} \citep{Dinerstein2017} are publicly available datasets. Sentinel 1 and Sentinel 2 data access are available through \href{https://planetarycomputer.microsoft.com/dataset/sentinel-2-l2a}{Microsoft Planetary Computer}. UK land cover classifications from the \href{https://www.ceh.ac.uk/data/ukceh-land-cover-maps}{UK Centre for Ecology and Hydrology}, ancient woodland boundaries from \href{https://naturalengland-defra.opendata.arcgis.com/datasets/Defra::ancient-woodland-revised-england-completed-counties-1/about}{Natural England} and \href{https://opendata.nature.scot/datasets/snh::ancient-woodland-inventory/explore}{NatureScot}, and National Park boundaries are publicly available through web access. Tessera embeddings are available through the \href{https://github.com/ucam-eo/geotessera}{geotessera} Python package.
\bmhead{Materials availability} Not applicable
\bmhead{Code availability} \href{https://github.com/rbnyng/spunpaper}{Available on Github}




\section{Declarations}

\begin{itemize}
\item \textbf{Funding} For funding support, we thank the Jeremy and Hannelore Grantham Environmental Trust, the Paul Allen Family Foundation, Bezos Earth Fund, Schmidt Family Foundation, Hefner Foundation, Cambridge Trust, and the Quadrature Climate Foundation, NWO-VICI [202.012] (ETK), NWO-Spinoza [SP.2023.2].
\item \textbf{Conflict of interest/Competing interests} Not applicable
\item \textbf{Ethics approval and consent to participate} Not applicable
\item \textbf{Consent for publication} Not applicable
\item \textbf{Author contribution} \textbf{Robin Young}: Conceptualization, Methodology, Software, Formal analysis, Visualization, Writing -- original draft. \textbf{Michael E. Van Nuland}: Methodology, Formal analysis, Visualization, Writing – review \& editing, Software \textbf{E. Toby Kiers}: Writing – review \& editing, Conceptualization \textbf{Tom{\'a}{\v s} V{\v e}trovsk{\'y}}: Writing – review \& editing \textbf{Petr Kohout}: Writing – review \& editing \textbf{Petr Baldrian}: Writing – review \& editing \textbf{Srinivasan Keshav}: Writing – review \& editing, Supervision, Conceptualization
\end{itemize}

\clearpage
\makeatletter
\let\addcontentsline\std@addcontentsline
\makeatother

\setcounter{figure}{0}
\setcounter{table}{0}
\setcounter{equation}{0}
\setcounter{section}{0}
\renewcommand{\thefigure}{S\arabic{figure}}
\renewcommand{\thetable}{S\arabic{table}}
\renewcommand{\theequation}{S\arabic{equation}}
\renewcommand{\thesection}{S\arabic{section}}

\begin{center}
{\LARGE\bfseries Supplementary Information}\\[0.8em] {\large Below-ground fungal biodiversity can be monitored using self-supervised learning satellite features}\\[0.6em] {\normalsize Robin Young, Michael E.\ Van Nuland, E.\ Toby Kiers, Tom\'{a}\v{s} V\v{e}trovsk\'{y}, Petr Kohout, Petr Baldrian, and Srinivasan Keshav}\\[0.4em] {\small Correspondence: Robin Young (\texttt{ray25@cam.ac.uk})}
\end{center}

\vspace{1.5em}
\hrule
\vspace{1.5em}

\tableofcontents

\newpage

\section{Tessera model details}
\label{note:tessera}

This section describes, at a high level, the data format and the model architecture of the Tessera foundation model, which is described in detail by \cite{feng2025tesseratemporalembeddingssurface}. The input data consists of Sentinel-2 optical satellite imagery (Level-2A bottom-of-atmosphere) and Sentinel-1 Synthetic Aperture Radar (SAR, Radiometrically Terrain Corrected).

\subsection{Data representation}

We consider remote sensing data with $C$ channels (spectral bands or polarizations). Each data tile $R_t$ at time $t$ is represented as a 3D array with dimensions:

\begin{equation}
R_t \in \mathbb{R}^{W \times H \times C}
\end{equation}

where $W$ is the width (longitude dimension), $H$ is the height (latitude dimension), and $C$ is the number of spectral channels.

Each tile is accompanied by a corresponding binary mask $V_t$ of dimensions:

\begin{equation}
V_t \in \{0,1\}^{W \times H}
\end{equation}

where $V_t(i,j) = 0$ indicates clouding or missing data for the pixel at spatial coordinates $(i,j)$, and $V_t(i,j) = 1$ indicates valid data.

\subsection{Temporal data stacking}

We stack spatially aligned tiles over a time period that spans $T$ time steps ($t = 0, 1, \ldots, T-1$). The temporal data stack is defined as:

\begin{equation}
\mathbf{D} = [R_0, R_1, \ldots, R_{T-1}]
\end{equation}

\begin{equation}
\mathbf{M} = [V_0, V_1, \ldots, V_{T-1}]
\end{equation}

\subsection{Time series extraction and d-pixel definition}

For a given spatial location $(i,j)$ and spectral channel $c$, the time series $S_{i,j,c}$ represents all channel $c$ values at coordinates $(i,j)$ over the entire time period:

\begin{equation}
S_{i,j,c} = [R_0(i,j,c), R_1(i,j,c), \ldots, R_{T-1}(i,j,c)]
\end{equation}

We define a \textit{d-pixel} $P_{i,j}$ as the collection of all spectral channels by timesteps at a given spatial location $(i,j)$:

\begin{equation}
P_{i,j}(c) = S(i,j, c)
\end{equation}

In other words, the d-pixel provides all spectral values (Sentinel-2) or backscatter values (Sentinel-1) at a given point over time. Note that d-pixels can be sparse and are accompanied by a mask vector $m_{i,j}$ of size $T$ that indicates the timesteps for which there are valid data, with a value 1 indicating that the corresponding row in $P_{i,j}$ is valid.

\subsection{d-pixel data structure}
\label{note:d-pixel}

We represent each 10m pixel in the time series of images from annual multispectral (Sentinel-2) and SAR backscatter (Sentinel-1) as a ``d-pixel'' array. In this array, each row corresponds to a chronologically ordered observation date, and each column is a specific spectral band or polarization. This representation accommodates data gaps encountered in remote sensing due to cloud cover (for optical data) or other atmospheric interference and acquisition irregularities. Observations identified as invalid (e.g. via Sentinel-2 scene classification layers for clouds, or missing Sentinel-1 acquisitions) are masked and are handled by downstream model components, for instance, by being ignored in the temporal sampling process.

\subsection{Dual-encoder architecture}

Given the distinct nature of Sentinel-1 SAR and Sentinel-2 MSI data, Tessera~\citep{feng2025tesseratemporalembeddingssurface} employs two separate, parallel transformer-based encoder branches.

\begin{itemize}
 \item \textbf{Sentinel-2 MSI Encoder}: This branch processes a time series of 10 spectral bands (excluding the 60m bands used for the detection of water vapour and clouds) from Sentinel-2. We used blue (B2), green (B3), red (B4), red edges 1--3 (B5, B6, B7), near-infrared (B8, B8A), and shortwave infrared (B11, B12).
 \item \textbf{Sentinel-1 SAR Encoder}: This branch processes a time series of 2 polarizations from Sentinel-1 (VV and VH).
\end{itemize}

Each encoder begins by embedding the input features (spectral bands or polarizations) for each time step through a two-layer MLP to a model width of 512. To preserve sequence order and incorporate temporal context, learnable positional encodings based on the Day-of-Year (DOY) of each observation are added to these embeddings. The core of each encoder consists of a stack of 8 standard Transformer blocks with a feed-forward width of 4{,}096, featuring multi-head self-attention and feed-forward layers to learn temporal patterns within the data streams.

To derive a single vector summarizing the entire time series for each modality, observations are pooled by attention. A recurrent temporal-context module (a single-layer GRU with 512 hidden units) contextualizes each timestep within the sequence, and a learned query assigns the pooling weights. The resulting modality-specific embeddings (one from the S1 encoder, one from the S2 encoder) are concatenated to 1{,}024 dimensions and projected to the 128-dimensional pixel embedding by a single linear layer.

\subsection{Projector network}
\label{note:projector}

The fused embedding from the dual-encoder stage is subsequently fed into a large projector network. This projector is a six-layer MLP. Its architecture comprises an input layer mapping the fused embeddings to 16,384 dimensions, followed by four hidden layers of 16,384 dimensions each, and a final linear output layer. Each of the first five layers is a fully-connected linear layer followed by Batch Normalization and a ReLU non-linear activation function, making the network deep and highly non-linear. This significant expansion in dimensionality is crucial, as suggested by the original Barlow Twins work~\cite{BTzbontar}, to enable effective redundancy reduction during the self-supervised loss computation. The final output of the projector is then used in the loss calculation. For downstream tasks, we used the 128-dimensional output from the fusion MLP (before the projector) as the final pixel embedding. The Tessera encoder (up to the fusion MLP) has approximately 88 million parameters, 44 million in each modality branch, while the projector accounts for the remaining 1.34 billion of the model's $\sim$1.4 billion parameters.

\subsection{Self-supervised training}
\label{note:ssl_training}

The Tessera model is trained using a modified Barlow Twins objective function~\cite{BTzbontar}. For this objective, two distorted views, denoted as $Y_A$ and $Y_B$, are generated for each input d-pixel. In Tessera, these views are created by independently running the temporal sampling and preprocessing pipeline twice for the Sentinel-1 and Sentinel-2 data associated with a given d-pixel. This process involves:

\begin{enumerate}
 \item For each view, independent sampling of a fixed number of valid observation dates from the annual Sentinel-2 time series (10 spectral bands).
 \item For each view, independent sampling of a fixed number of valid observation dates from the annual Sentinel-1 time series (2 polarizations).
\end{enumerate}

These views represent different, valid, but inherently incomplete glimpses of the pixel's true temporal-spectral evolution, akin to observing the same location through intermittent cloud cover or from different satellite passes at different times. The model learns by reconciling these partial views. The inherent differences between the Sentinel-1 SAR and Sentinel-2 MSI modalities further provide diverse perspectives on the same underlying physical processes. Thus, our augmentations are fundamentally about sampling from the available, inherently incomplete information streams, rather than artificially distorting a complete input.

The network processes these two views ($Y_A, Y_B$) through the dual-encoder and the projector to produce batch-normalized embeddings $Z_A$ and $Z_B$. The standard Barlow Twins loss function, $\mathcal{L}_{BT}$, is defined as~\cite{BTzbontar}:

\begin{equation} \label{eq:barlow_twins_loss}
\mathcal{L}_{BT} = \sum_i (1 - C_{ii})^2 + \lambda_{BT} \sum_i \sum_{j \neq i} C_{ij}^2
\end{equation}

Here, $C$ is the cross-correlation matrix computed between the batch-normalized embeddings $Z_A$ and $Z_B$. The indices $i$ and $j$ iterate over the dimensions of the embedding vectors. The first term (invariance term) encourages similar embeddings for different views of the same input ($C_{ii} \to 1$). The second term (redundancy reduction term) promotes information efficiency by minimizing the correlation between different embedding dimensions ($C_{ij} \to 0$ for $i \neq j$), weighted by $\lambda_{BT}$.

To further enhance the robustness of the model and mitigate overfitting, Tessera incorporates an additional mix-up regularization term $\mathcal{L}_{MIX}$, inspired by~\cite{Bandara2023MixCoAM}. For each training batch, this involves shuffling one set of views (e.g. $Y_B$) along the batch dimension to create $Y_S = \text{Shuffle}(Y_B)$, then generating a mixed view $Y_M = \alpha_{mix}Y_A + (1-\alpha_{mix})Y_S$. The mixing coefficient $\alpha_{mix}$ is sampled from a uniform distribution on the unit interval, $\alpha_{mix} \sim U(0, 1)$. The embeddings $Z_M$ and $Z_S$ are obtained from their respective views. The mix-up loss penalizes deviations from the assumption that a linear interpolation in the input space corresponds to a linear interpolation in the embedding space.

\begin{align}
C_{target}^{MA} &= \alpha_{mix}(Z_{A})^{T}Z_{A}+(1-\alpha_{mix})(Z_{S})^{T}Z_{A} \\
C_{target}^{MS} &= \alpha_{mix}(Z_{A})^{T}Z_{S}+(1-\alpha_{mix})(Z_{S})^{T}Z_{S} \\
\mathcal{L}_{MIX} &= \frac{1}{2}\bigl(\|C^{MA}-C_{target}^{MA}\|_{F}^{2}+\|C^{MS}-C_{target}^{MS}\|_{F}^{2}\bigr)
\end{align}

where $C^{MA}=(Z_{M})^{T}Z_{A}$ and $C^{MS}=(Z_{M})^{T}Z_{S}$ are the actual cross-correlation matrices from the model's outputs. The total loss function optimized during training is a weighted sum:

\begin{equation} \label{eq:total_loss}
\mathcal{L}_{total} = \mathcal{L}_{BT} + \lambda_{mix}\mathcal{L}_{MIX}
\end{equation}

where $\lambda_{mix}$ controls the strength of the mix-up regularization.

\subsection{Trained configuration}
\label{note:ssl_hparams}

The description above is of the Tessera architecture in general. Table~\ref{tab:ssl_config} gives the specific configuration of the checkpoint whose embeddings are used throughout this work. More details on the specific architecture of the model and testing against a broad range of ecological tasks can be found in~\citep{feng2025tesseratemporalembeddingssurface} and are not restated here.

\begin{table}[ht]
\centering
\caption{Configuration of the trained Tessera checkpoint used in this work. Pretraining settings are those of the released Tessera v1.0 model (Table 8 of \protect\citealp{feng2025tesseratemporalembeddingssurface}); the remaining rows describe our feature-extraction pipeline.}
\label{tab:ssl_config}
\begin{tabular}{p{5.2cm} p{3.0cm} p{6.0cm}}
\toprule
\textbf{Setting} & \textbf{Value} & \textbf{Notes} \\
\midrule
\multicolumn{3}{l}{\textit{Representation}} \\
Embedding dimension & 128 & per 10m pixel, per year \\
Encoder width $d_{model}$ & 512 & per modality branch \\
Transformer blocks & 8 & per modality branch, feed-forward width 4{,}096 \\
Modality fusion & concatenation & Sentinel-1 and Sentinel-2 encoder outputs \\
Projector hidden width & 16{,}384 & discarded after pretraining \\
Projector output width & 16{,}384 & dimension at which the Barlow Twins loss is applied \\
Parameters & 88M / 1.34B & encoder / projector \\
\addlinespace
\multicolumn{3}{l}{\textit{Temporal sampling}} \\
Sampled timesteps, Sentinel-2 & 40 & drawn independently per view \\
Sampled timesteps, Sentinel-1 & 40 & drawn independently per view \\
Minimum valid timesteps & 20 & d-pixels below this threshold are excluded \\
\addlinespace
\multicolumn{3}{l}{\textit{Objective}} \\
Barlow Twins $\lambda_{BT}$ & $5\times10^{-3}$ & off-diagonal weight in Eq.~\ref{eq:barlow_twins_loss} \\
Mix-up $\lambda_{mix}$ & 1.0 & weight of $\mathcal{L}_{MIX}$ in Eq.~\ref{eq:total_loss} \\
\addlinespace
\multicolumn{3}{l}{\textit{Optimization}} \\
Optimizer & AdamW & \\
Weight decay & $1\times10^{-6}$ & \\
Gradient clipping norm & 2.0 & \\
Base learning rate & 0.002 & linear warmup over the first 10\%, then cosine decay \\
Batch size & 2{,}048 & \\
Epochs & 1 & single pass over the pretraining corpus \\
\bottomrule
\end{tabular}
\end{table}

\FloatBarrier

\section{Ground-truth sample patch generation}
\label{note:patches}

The initial input consists of approximately 12,000 ectomycorrhizal (ECM) fungal richness samples with geographic coordinates. To prepare these for satellite data processing, each point was converted into a small, georeferenced raster patch.

\begin{itemize}
 \item \textbf{Process:} A $3 \times 3$ pixel patch ($30\text{m} \times 30\text{m}$) was created to account for local spatial context and any possible coordinate uncertainty, centred on each sample's coordinates.
 \item \textbf{Projection:} Coordinates were projected into their local Universal Transverse Mercator (UTM) zone for accurate geometric measurements.
\end{itemize}

\section{Satellite data acquisition and stacking}
\label{note:satellite_data}

For each sample patch, a one-year time-series of Sentinel-1 and Sentinel-2 imagery was acquired and processed into the spatio-temporal data format described above, ready for the inference stage. The acquisition window is the calendar year from which embeddings are extracted, 1 January to 31 December inclusive.

\subsection{Sentinel-2 optical data}

\begin{itemize}
 \item \textbf{Source \& parameters:} Sentinel-2 L2A data were sourced from Microsoft Planetary Computer's STAC catalog, filtered for images with $<$90\% cloud cover. This scene-level filter only discards scenes that are near-certainly unusable. Per-pixel validity is decided by the scene classification layer below.
 \item \textbf{Bands acquired (10m):} B02 (Blue), B03 (Green), B04 (Red), B05 (Red Edge 1), B06 (Red Edge 2), B07 (Red Edge 3), B08 (NIR), B8A (Narrow NIR), B11 (SWIR 1), B12 (SWIR 2), and the Scene Classification Layer (SCL) for masking. The six bands with a native 20m ground sampling distance (B05, B06, B07, B8A, B11, B12) were resampled to the 10m grid by nearest neighbor, so that no band is smoothed across pixel boundaries.
 \item \textbf{Radiometric harmonization:} Scenes acquired after 25 January 2022 are produced under processing baseline 04.00, which applies a $+1{,}000$ count offset to bottom-of-atmosphere reflectance. This offset was subtracted from all such scenes so that the full series is on a single radiometric scale. Without this step the pre- and post-2022 portions of a series are not comparable.
 \item \textbf{Pixel validity:} A pixel-date is treated as invalid, and excluded from the series rather than set to zero, where its SCL class is 0 (no data), 1 (saturated or defective), 2 (cast shadow), 3 (cloud shadow), 8 (cloud, medium probability) or 9 (cloud, high probability). Classes 10 (thin cirrus) and 11 (snow or ice) are retained, as both carry surface information at the seasonal scale the representation is built to encode. Class 6 (water) is likewise retained. Water is instead handled by masking at the map-production stage, as discussed in the main text.
 \item \textbf{Same-date coverage:} Where more than one scene covered a patch on a given date, typically at the overlap between adjacent orbit swaths, each pixel takes its value from the first scene in which that pixel is valid, so that a date contributes exactly one observation per pixel. Dates on which fewer than 5\% of patch pixels are valid were discarded entirely.
\end{itemize}

\subsection{Sentinel-1 SAR data}

\begin{itemize}
 \item \textbf{Source \& parameters:} Sentinel-1 Radiometric Terrain Corrected (RTC) data were sourced from Microsoft Planetary Computer, including both ascending and descending orbits. The product is gamma-nought backscatter, terrain-corrected against the Copernicus 30m DEM by the data provider. No further radiometric or geometric correction was applied on our side, and no speckle filter was applied. Ascending and descending acquisitions are retained as separate observations rather than combined, so that the encoder sees the viewing geometry as part of the temporal signal.
 \item \textbf{Bands acquired (10m):} VV and VH polarizations were processed. Amplitude values were converted to a decibel (dB) scale. Specifically, values are transformed as $20\log_{10}(x)$, offset by $+50$ and scaled by $200$ before storage as 16-bit integers, a fixed affine reparameterization that preserves the ordering and relative spacing of backscatter values. The same 5\% minimum-coverage rule and the same first-valid-scene rule described for Sentinel-2 apply.
\end{itemize}

\subsection{Output}

The output of this stage is a collection of directories, one for each sample. Each directory holds the per-date, per-band rasters for that patch, which are then assembled into the d-pixel representation of Supplementary Note~\ref{note:d-pixel}. For each of the nine pixels in the $3\times3$ patch, the valid observations are ordered by acquisition date and written as three arrays. \texttt{bands.npy} holds the observation values with shape (dates, pixels, channels), \texttt{masks.npy} the per-observation validity flags with shape (dates, pixels), and \texttt{doys.npy} the day-of-year of each observation with shape (dates,), which supplies the positional encoding described in Supplementary Note~\ref{note:tessera}. Sentinel-1 and Sentinel-2 are stored as separate array sets, since the two encoders consume them independently and their acquisition dates do not coincide. Invalid observations are carried through the mask rather than removed, so that array shapes are uniform across samples.

\section{SSL feature extraction}
\label{note:ssl_features}

The spatiotemporal d-pixels from the previous stage were processed through an inference pipeline to generate the final SSL feature embeddings.

\subsection{SSL model checkpoint}

\begin{itemize}
 \item \textbf{Model:} We used the pretrained model checkpoint for the Tessera architecture as described in \cite{feng2025tesseratemporalembeddingssurface}.
 \item \textbf{Source:} The Tessera v1 \href{https://drive.google.com/drive/folders/18RPptbUkCIgUfw1aMdMeOrFML_ZVMszn}{checkpoint} is publicly available with further instructions on how to set it up for inference at the Tessera GitHub repository at \url{https://github.com/ucam-eo/tessera}.
 \item \textbf{Architecture:} The model consists of two transformer backbones for processing Sentinel-1 and Sentinel-2 time-series, respectively, described above.
\end{itemize}

\subsection{Inference process}

An inference pipeline was developed to apply the Tessera model to the approximately 12,000 d-pixels.

\begin{itemize}
 \item \textbf{Workflow:} A shell script manages the inference workflow. It is designed for execution on a cluster machine, distributing the workload and skipping already-completed tiles for fault tolerance.
 \item \textbf{Logic:} The script loads the pre-trained model and processes each d-pixel. For each d-pixel in a sample's $3 \times 3$ patch, the script performs the following steps:
 \begin{enumerate}
 \item Randomly samples 40 time-steps from the valid Sentinel-2 observations.
 \item Randomly samples 40 time-steps from the valid Sentinel-1 observations.
 \item Feeds these sampled time-series into the respective S2 and S1 encoder backbones of the Tessera model.
 \item Fuses the resulting representations using concatenation.
 \item Passes the fused vector through a final layer to produce a high-dimensional embedding.
 \end{enumerate}
 \item The final output is a collection of \texttt{.npy} files, where each file corresponds to a sample and contains a $3 \times 3 \times 128$ feature tensor. This tensor is flattened to a vector for downstream modeling.
\end{itemize}

\section{Environmental variables}
\label{note:env_variables}

\begin{table}[ht]
\centering
\caption{Summary of environmental predictor variables used as baseline features for comparison with SSL satellite features.}
\label{tab:env_variables}
\small
\begin{tabular}{p{2.8cm} p{2.5cm} p{1.8cm} p{5.5cm}}
\toprule
\textbf{Category} & \textbf{Source} & \textbf{Resolution} & \textbf{Variables \& processing} \\
\midrule
Climate &
WorldClim 2.1~\citep{Fick2017WorldClim2} &
$\sim$1km &
19 bioclimatic variables (BIO1--BIO19); annual mean, min, and max of monthly \texttt{tavg}, \texttt{tmin}, \texttt{tmax}, \texttt{prec}, \texttt{srad}, \texttt{vapr}, \texttt{wind} \\
\addlinespace
Topography &
WorldClim 2.1 DEM~\citep{Fick2017WorldClim2} &
$\sim$1km &
Elevation (ma.s.l.), slope (\textdegree), and aspect (\textdegree) derived via \texttt{richdem}~\citep{RichDEM} \\
\addlinespace
Soil properties &
SoilGrids 2.0~\citep{soilgrids} &
250m &
pH, organic carbon, nitrogen, clay/sand/silt fractions, CEC, and bulk density at 0--5cm and 5--15cm depths; WRB soil class (one-hot encoded) \\
\addlinespace
Land cover &
ESA WorldCover 2021~\citep{Zanaga2022WorldCover} &
10m &
11 land cover classes (one-hot encoded) \\
\addlinespace
Geography &
--- &
--- &
Latitude and longitude (continuous) \\
\bottomrule
\end{tabular}
\end{table}

\FloatBarrier

\section{Model implementation and hyperparameters}
\label{note:model_params}

All models were implemented in Python. This section details the specific libraries and hyperparameters used for the results presented in the main manuscript. Our objective was to use robust, standard configurations to fairly evaluate the information content of the feature sets, rather than achieve marginal performance gains through extensive hyperparameter optimization.

\subsection{LightGBM (\texttt{lgb.LGBMRegressor})}

Parameters: \texttt{n\_estimators}=1000; \texttt{learning\_rate}=0.05; \texttt{random\_state} set to the run's random seed; \texttt{n\_jobs}=$-1$; \texttt{early\_stopping}=15. All other parameters were left at their library defaults.

\subsection{XGBoost (\texttt{xgb.XGBRegressor})}

Parameters: \texttt{n\_estimators}=1000; \texttt{learning\_rate}=0.05; \texttt{random\_state} set to the run's random seed; \texttt{n\_jobs}=$-1$; \texttt{early\_stopping\_rounds}=15. All other parameters were left at their library defaults.

\subsection{Random Forest (\texttt{sklearn.ensemble.RandomForestRegressor})}

Parameters: \texttt{n\_estimators}=100; \texttt{random\_state} set to the run's random seed; \texttt{n\_jobs}=$-1$. All other parameters were left at their library defaults.

\section{Dimensionality reduction of the satellite features}
\label{note:dimreduction}

The satellite block is reduced before it enters the downstream model. We note that this reduction is distinct from the principal component analysis of correlation of embeddings to environmental features below. The reduction described here is part of the predictive pipeline, whereas the decomposition there is a descriptive analysis of embedding structure and is not used for prediction.

Each sample's $3 \times 3 \times 128$ embedding patch is flattened to 1{,}152 values and reduced to 256 components with UMAP~\citep{McInnes2018}, using \texttt{umap-learn} with \texttt{n\_neighbors}~$=15$, \texttt{min\_dist}~$=0.1$, a Euclidean metric and a fixed random state, preceded by feature-wise standardization. The reduction is applied to the satellite block only. Climate, soil, land cover, topographic and coordinate features enter the model unreduced. The reducer and the standardizer are fitted on the training split of each run and then applied unchanged to the validation and test splits. The reducer's random state is held fixed across runs so that the only quantity varying between them is the split.

UMAP is a non-linear manifold embedding and has no explained-variance decomposition, so no variance-explained figure is available for this step. The choice of 256 components was made to keep the satellite block comparable in width to the combined environmental blocks rather than by a variance criterion. For the reader who wants a variance accounting of the embedding, the linear decomposition in Supplementary~\ref{note:crosscorr} provides one, and we find that 7 components reach 50\% of embedding variance, 34 reach 80\%, and 50 reach 85.5\%.

\section{PCA feature correlations}
\label{note:crosscorr}

The representation learning model applied to the satellite data is unsupervised, so it is reasonable to ask what the resulting components correspond to. This note reports, for the leading ten principal components of the embedding, the variance each explains, its correlation with every environmental variable together with bootstrap confidence intervals, and a resampling check on how variable those associations are.

\subsection{Variance explained}

\begin{table}[ht]
\centering
\caption{Variance explained by the leading principal components of the SSL embedding.}
\label{tab:varexplained}
\begin{tabular}{lcccccccccc}
\toprule
\textbf{Component} & PC1 & PC2 & PC3 & PC4 & PC5 & PC6 & PC7 & PC8 & PC9 & PC10 \\
\midrule
Explained variance (\%) & 14.4 & 12.0 & 7.0 & 6.3 & 4.8 & 3.4 & 3.1 & 2.9 & 2.5 & 2.3 \\
Cumulative (\%) & 14.4 & 26.4 & 33.4 & 39.7 & 44.4 & 47.8 & 50.9 & 53.8 & 56.3 & 58.6 \\
\bottomrule
\end{tabular}
\end{table}

\FloatBarrier

The representation is not dominated by a few directions. Seven components are required to reach 50\% of the variance, 34 to reach 80\%, and 50 components reach 85.5\%. The ten components in Table~\ref{tab:pca_correlations} account for 58.6\% between them.

{\footnotesize
\setlength{\tabcolsep}{4pt}
\begin{longtable}{l rrrrrrrrrr}
\caption{Pearson correlation between the leading ten principal components of the SSL embedding and every environmental variable. Correlations with $|r| \geq 0.30$ (before rounding) are in bold. Correlations whose 95\% bootstrap confidence interval (2000 bootstraps) includes zero are marked $^{\dagger}$. All others are significant at $\alpha = 0.05$.}
\label{tab:pca_correlations} \\
\toprule
\textbf{Variable} & \textbf{PC1} & \textbf{PC2} & \textbf{PC3} & \textbf{PC4} & \textbf{PC5} & \textbf{PC6} & \textbf{PC7} & \textbf{PC8} & \textbf{PC9} & \textbf{PC10} \\
\midrule
\endfirsthead
\multicolumn{11}{l}{\small\itshape Table \thetable\ continued from previous page} \\
\toprule
\textbf{Variable} & \textbf{PC1} & \textbf{PC2} & \textbf{PC3} & \textbf{PC4} & \textbf{PC5} & \textbf{PC6} & \textbf{PC7} & \textbf{PC8} & \textbf{PC9} & \textbf{PC10} \\
\midrule
\endhead
\midrule
\multicolumn{11}{r}{\small\itshape continued on next page} \\
\endfoot
\bottomrule
\endlastfoot
\multicolumn{11}{l}{\textit{Topography}} \\
Elevation & 0.16 & \textbf{0.38} & 0.03 & $-$0.05 & \textbf{$-$0.50} & 0.09 & \textbf{0.30} & \textbf{$-$0.37} & 0.14 & 0.04 \\
\addlinespace
\multicolumn{11}{l}{\textit{Temperature}} \\
Bio 1 (Ann.\ Mean Temp) & $-$0.30 & $-$0.22 & \textbf{$-$0.76} & 0.14 & 0.00$^{\dagger}$ & 0.05 & 0.16 & 0.07 & 0.00$^{\dagger}$ & 0.06 \\
Bio 2 (Mean Diurnal Range) & 0.20 & $-$0.05 & $-$0.16 & $-$0.05 & $-$0.02$^{\dagger}$ & $-$0.02$^{\dagger}$ & \textbf{0.39} & $-$0.10 & 0.03 & 0.05 \\
Bio 3 (Isothermality) & $-$0.03 & 0.09 & \textbf{$-$0.61} & $-$0.03 & $-$0.24 & $-$0.04 & 0.27 & $-$0.21 & 0.04 & 0.19 \\
Bio 4 (Temp Seasonality) & 0.19 & $-$0.17 & \textbf{0.60} & $-$0.04 & 0.29 & 0.02$^{\dagger}$ & $-$0.01$^{\dagger}$ & 0.13 & $-$0.04 & $-$0.15 \\
Bio 5 (Max Temp Warmest Mo.) & $-$0.14 & \textbf{$-$0.35} & \textbf{$-$0.52} & 0.12 & 0.17 & 0.06 & 0.27 & 0.13 & $-$0.01$^{\dagger}$ & $-$0.02$^{\dagger}$ \\
Bio 6 (Min Temp Coldest Mo.) & \textbf{$-$0.32} & $-$0.10 & \textbf{$-$0.77} & 0.13 & $-$0.11 & 0.03 & 0.08 & 0.01$^{\dagger}$ & 0.02$^{\dagger}$ & 0.10 \\
Bio 7 (Temp Annual Range) & 0.27 & $-$0.16 & \textbf{0.50} & $-$0.05 & 0.27 & 0.01$^{\dagger}$ & 0.12 & 0.09 & $-$0.03 & $-$0.13 \\
Bio 8 (Mean Temp Wettest Qtr) & $-$0.14 & \textbf{$-$0.34} & 0.12 & 0.08 & 0.19 & 0.11 & $-$0.01$^{\dagger}$ & $-$0.06 & $-$0.09 & $-$0.02$^{\dagger}$ \\
Bio 9 (Mean Temp Driest Qtr) & $-$0.05 & $-$0.03 & \textbf{$-$0.65} & 0.10 & $-$0.19 & 0.00$^{\dagger}$ & 0.28 & 0.08 & 0.01$^{\dagger}$ & $-$0.01$^{\dagger}$ \\
Bio 10 (Mean Temp Warmest Qtr) & $-$0.23 & \textbf{$-$0.35} & \textbf{$-$0.58} & 0.15 & 0.16 & 0.06 & 0.20 & 0.17 & $-$0.03 & $-$0.01$^{\dagger}$ \\
Bio 11 (Mean Temp Coldest Qtr) & $-$0.27 & $-$0.09 & \textbf{$-$0.78} & 0.12 & $-$0.11 & 0.03 & 0.13 & 0.01$^{\dagger}$ & 0.01$^{\dagger}$ & 0.09 \\
Avg Temp (Mean) & $-$0.30 & $-$0.22 & \textbf{$-$0.76} & 0.14 & 0.00$^{\dagger}$ & 0.05 & 0.16 & 0.07 & 0.00$^{\dagger}$ & 0.06 \\
Avg Temp (Max) & $-$0.19 & \textbf{$-$0.35} & \textbf{$-$0.53} & 0.15 & 0.18 & 0.07 & 0.22 & 0.18 & $-$0.03 & $-$0.03$^{\dagger}$ \\
Avg Temp (Min) & $-$0.27 & $-$0.08 & \textbf{$-$0.78} & 0.11 & $-$0.12 & 0.03 & 0.13 & 0.00$^{\dagger}$ & 0.02$^{\dagger}$ & 0.10 \\
Avg Temp (SD) & 0.19 & $-$0.17 & \textbf{0.60} & $-$0.04 & 0.29 & 0.02$^{\dagger}$ & $-$0.01$^{\dagger}$ & 0.13 & $-$0.04 & $-$0.15 \\
Max Temp (Mean) & $-$0.25 & $-$0.22 & \textbf{$-$0.75} & 0.13 & 0.00$^{\dagger}$ & 0.04 & 0.22 & 0.05 & 0.00$^{\dagger}$ & 0.07 \\
Max Temp (Max) & $-$0.14 & \textbf{$-$0.35} & \textbf{$-$0.52} & 0.12 & 0.17 & 0.06 & 0.27 & 0.13 & $-$0.01$^{\dagger}$ & $-$0.02$^{\dagger}$ \\
Max Temp (Min) & $-$0.21 & $-$0.06 & \textbf{$-$0.78} & 0.10 & $-$0.14 & 0.02$^{\dagger}$ & 0.18 & $-$0.01$^{\dagger}$ & 0.01$^{\dagger}$ & 0.11 \\
Max Temp (SD) & 0.10 & $-$0.28 & \textbf{0.49} & 0.00$^{\dagger}$ & \textbf{0.35} & 0.06 & 0.01$^{\dagger}$ & 0.11 & $-$0.01$^{\dagger}$ & $-$0.16 \\
Min Temp (Mean) & \textbf{$-$0.34} & $-$0.21 & \textbf{$-$0.74} & 0.15 & 0.00$^{\dagger}$ & 0.05 & 0.09 & 0.09 & $-$0.01$^{\dagger}$ & 0.05 \\
Min Temp (Max) & $-$0.24 & \textbf{$-$0.32} & \textbf{$-$0.50} & 0.16 & 0.18 & 0.06 & 0.11 & 0.24 & $-$0.06 & $-$0.04 \\
Min Temp (Min) & \textbf{$-$0.32} & $-$0.10 & \textbf{$-$0.77} & 0.13 & $-$0.11 & 0.03 & 0.08 & 0.01$^{\dagger}$ & 0.02$^{\dagger}$ & 0.10 \\
Min Temp (SD) & 0.26 & $-$0.05 & \textbf{0.67} & $-$0.07 & 0.22 & $-$0.02$^{\dagger}$ & $-$0.02$^{\dagger}$ & 0.14 & $-$0.07 & $-$0.13 \\
\addlinespace
\multicolumn{11}{l}{\textit{Precipitation}} \\
Bio 12 (Ann.\ Precipitation) & $-$0.04 & 0.17 & $-$0.07 & $-$0.04 & $-$0.25 & 0.09 & 0.03 & $-$0.30 & 0.10 & 0.05 \\
Bio 13 (Precip Wettest Mo.) & 0.01$^{\dagger}$ & 0.15 & 0.08 & $-$0.04 & $-$0.27 & 0.10 & 0.09 & $-$0.27 & 0.05 & 0.05 \\
Bio 14 (Precip Driest Mo.) & $-$0.13 & 0.18 & $-$0.10 & $-$0.03 & $-$0.17 & 0.07 & $-$0.08 & \textbf{$-$0.31} & 0.07 & 0.17 \\
Bio 15 (Precip Seasonality) & 0.15 & $-$0.11 & 0.26 & 0.02$^{\dagger}$ & $-$0.03 & 0.04 & 0.21 & 0.16 & $-$0.03 & $-$0.17 \\
Bio 16 (Precip Wettest Qtr) & $-$0.01$^{\dagger}$ & 0.13 & 0.04 & $-$0.03 & $-$0.26 & 0.12 & 0.07 & $-$0.28 & 0.08 & 0.01$^{\dagger}$ \\
Bio 17 (Precip Driest Qtr) & $-$0.10 & 0.19 & $-$0.14 & $-$0.04 & $-$0.19 & 0.05 & $-$0.05 & \textbf{$-$0.31} & 0.08 & 0.16 \\
Bio 18 (Precip Warmest Qtr) & $-$0.09 & 0.07 & 0.21 & $-$0.07 & $-$0.10 & 0.13 & $-$0.12 & \textbf{$-$0.37} & 0.06 & 0.10 \\
Bio 19 (Precip Coldest Qtr) & 0.01$^{\dagger}$ & 0.20 & $-$0.22 & $-$0.05 & $-$0.29 & 0.00$^{\dagger}$ & 0.09 & $-$0.18 & 0.08 & 0.03$^{\dagger}$ \\
Precip (Mean) & $-$0.04 & 0.17 & $-$0.07 & $-$0.04 & $-$0.25 & 0.09 & 0.03 & $-$0.30 & 0.10 & 0.05 \\
Precip (Max) & 0.01$^{\dagger}$ & 0.15 & 0.08 & $-$0.04 & $-$0.27 & 0.10 & 0.09 & $-$0.27 & 0.05 & 0.05 \\
Precip (Min) & $-$0.13 & 0.18 & $-$0.10 & $-$0.03 & $-$0.17 & 0.07 & $-$0.08 & \textbf{$-$0.31} & 0.07 & 0.17 \\
Precip (SD) & 0.10 & $-$0.01$^{\dagger}$ & 0.21 & 0.00$^{\dagger}$ & $-$0.20 & 0.12 & 0.19 & $-$0.07 & 0.03 & $-$0.17 \\
\addlinespace
\multicolumn{11}{l}{\textit{Solar radiation}} \\
Solar Rad. (Mean) & 0.00$^{\dagger}$ & 0.08 & \textbf{$-$0.48} & 0.10 & \textbf{$-$0.39} & 0.14 & \textbf{0.48} & $-$0.21 & 0.12 & $-$0.02$^{\dagger}$ \\
Solar Rad. (Max) & 0.15 & $-$0.08 & $-$0.18 & 0.15 & $-$0.18 & 0.16 & \textbf{0.42} & 0.00$^{\dagger}$ & 0.08 & $-$0.21 \\
Solar Rad. (Min) & 0.02$^{\dagger}$ & 0.16 & \textbf{$-$0.49} & 0.05 & \textbf{$-$0.43} & 0.10 & \textbf{0.47} & $-$0.25 & 0.11 & 0.04 \\
Solar Rad. (SD) & 0.12 & $-$0.28 & \textbf{0.39} & 0.11 & \textbf{0.31} & 0.05 & $-$0.11 & \textbf{0.31} & $-$0.06 & $-$0.26 \\
\addlinespace
\multicolumn{11}{l}{\textit{Vapor pressure}} \\
Vapor Pr. (Mean) & \textbf{$-$0.31} & $-$0.21 & \textbf{$-$0.73} & 0.12 & 0.05 & $-$0.01$^{\dagger}$ & 0.08 & 0.13 & $-$0.05 & 0.05 \\
Vapor Pr. (Max) & $-$0.29 & \textbf{$-$0.37} & \textbf{$-$0.46} & 0.14 & 0.24 & 0.05 & $-$0.02$^{\dagger}$ & 0.19 & $-$0.02$^{\dagger}$ & $-$0.12 \\
Vapor Pr. (Min) & $-$0.22 & $-$0.06 & \textbf{$-$0.77} & 0.09 & $-$0.10 & $-$0.07 & 0.13 & 0.09 & $-$0.05 & 0.12 \\
Vapor Pr. (SD) & $-$0.24 & \textbf{$-$0.41} & 0.03 & 0.11 & \textbf{0.37} & 0.13 & $-$0.13 & 0.16 & 0.01$^{\dagger}$ & $-$0.20 \\
\addlinespace
\multicolumn{11}{l}{\textit{Wind speed}} \\
Wind Speed (Mean) & $-$0.11 & 0.15 & $-$0.23 & 0.07 & $-$0.16 & $-$0.03 & $-$0.09 & 0.16 & 0.03 & $-$0.10 \\
Wind Speed (Max) & $-$0.10 & 0.17 & $-$0.24 & 0.03 & $-$0.17 & $-$0.05 & $-$0.07 & 0.17 & 0.03 & $-$0.13 \\
Wind Speed (Min) & $-$0.11 & 0.14 & $-$0.22 & 0.11 & $-$0.15 & $-$0.03 & $-$0.09 & 0.16 & 0.02$^{\dagger}$ & $-$0.06 \\
Wind Speed (SD) & $-$0.07 & 0.15 & $-$0.21 & $-$0.06 & $-$0.15 & $-$0.06 & $-$0.02$^{\dagger}$ & 0.17 & 0.02$^{\dagger}$ & $-$0.21 \\
\addlinespace
\multicolumn{11}{l}{\textit{Soil}} \\
Bulk density (0--5cm) & \textbf{$-$0.38} & $-$0.01$^{\dagger}$ & $-$0.21 & \textbf{0.43} & \textbf{$-$0.38} & 0.02$^{\dagger}$ & 0.23 & $-$0.14 & 0.09 & 0.15 \\
Bulk density (5--15cm) & \textbf{$-$0.35} & $-$0.23 & 0.00$^{\dagger}$ & \textbf{0.47} & $-$0.23 & 0.06 & 0.12 & $-$0.13 & 0.16 & 0.04 \\
Bulk density (depth gradient) & 0.00$^{\dagger}$ & \textbf{0.44} & \textbf{$-$0.39} & $-$0.15 & $-$0.25 & $-$0.07 & 0.20 & 0.01$^{\dagger}$ & $-$0.15 & 0.21 \\
Cation exch.\ capacity (0--5cm) & $-$0.08 & $-$0.22 & \textbf{0.40} & \textbf{0.40} & 0.01$^{\dagger}$ & 0.01$^{\dagger}$ & $-$0.12 & $-$0.16 & 0.23 & 0.05 \\
Cation exch.\ capacity (5--15cm) & $-$0.06 & $-$0.06 & \textbf{0.32} & \textbf{0.37} & $-$0.12 & $-$0.01$^{\dagger}$ & $-$0.05 & $-$0.18 & 0.23 & 0.12 \\
Cation exch.\ capacity (depth gradient) & $-$0.07 & \textbf{$-$0.35} & \textbf{0.36} & 0.30 & 0.18 & 0.04 & $-$0.18 & $-$0.08 & 0.15 & $-$0.05 \\
Clay content (0--5cm) & \textbf{$-$0.37} & $-$0.06 & $-$0.18 & \textbf{0.32} & $-$0.28 & 0.04 & 0.23 & $-$0.23 & 0.04 & 0.06 \\
Clay content (5--15cm) & \textbf{$-$0.38} & $-$0.04 & $-$0.20 & \textbf{0.31} & $-$0.29 & 0.03 & 0.23 & $-$0.22 & 0.02$^{\dagger}$ & 0.07 \\
Clay content (depth gradient) & 0.20 & $-$0.11 & 0.25 & $-$0.07 & 0.15 & 0.07 & $-$0.12 & 0.02$^{\dagger}$ & 0.10 & $-$0.11 \\
Nitrogen (0--5cm) & $-$0.10 & $-$0.27 & \textbf{0.42} & \textbf{0.32} & 0.06 & 0.12 & $-$0.18 & $-$0.15 & 0.24 & $-$0.09 \\
Nitrogen (5--15cm) & 0.00$^{\dagger}$ & $-$0.21 & \textbf{0.42} & 0.24 & $-$0.04 & 0.13 & $-$0.20 & $-$0.18 & 0.26 & $-$0.07 \\
Nitrogen (depth gradient) & $-$0.16 & $-$0.24 & 0.28 & 0.28 & 0.13 & 0.07 & $-$0.11 & $-$0.07 & 0.14 & $-$0.08 \\
pH (H$_2$O) (0--5cm) & $-$0.26 & $-$0.14 & 0.03$^{\dagger}$ & \textbf{0.47} & $-$0.21 & 0.11 & 0.11 & $-$0.15 & 0.17 & $-$0.06 \\
pH (H$_2$O) (5--15cm) & $-$0.26 & $-$0.14 & 0.02$^{\dagger}$ & \textbf{0.47} & $-$0.21 & 0.11 & 0.10 & $-$0.14 & 0.17 & $-$0.06 \\
pH (H$_2$O) (depth gradient) & 0.18 & 0.13 & 0.09 & $-$0.21 & 0.08 & $-$0.03 & 0.01$^{\dagger}$ & 0.00$^{\dagger}$ & 0.00$^{\dagger}$ & 0.06 \\
Sand content (0--5cm) & $-$0.01$^{\dagger}$ & $-$0.24 & 0.29 & \textbf{0.34} & $-$0.02$^{\dagger}$ & 0.08 & $-$0.14 & 0.00$^{\dagger}$ & 0.23 & $-$0.04 \\
Sand content (5--15cm) & $-$0.01$^{\dagger}$ & $-$0.24 & 0.30 & \textbf{0.33} & $-$0.02$^{\dagger}$ & 0.09 & $-$0.15 & 0.00$^{\dagger}$ & 0.23 & $-$0.05 \\
Sand content (depth gradient) & $-$0.07 & 0.12 & $-$0.16 & 0.03 & $-$0.09 & $-$0.20 & 0.19 & 0.04 & $-$0.09 & 0.25 \\
Silt content (0--5cm) & \textbf{$-$0.32} & $-$0.10 & 0.07 & 0.30 & $-$0.26 & $-$0.02$^{\dagger}$ & 0.11 & $-$0.22 & 0.13 & 0.14 \\
Silt content (5--15cm) & \textbf{$-$0.31} & $-$0.10 & 0.07 & 0.30 & $-$0.26 & $-$0.03 & 0.12 & $-$0.22 & 0.13 & 0.16 \\
Silt content (depth gradient) & $-$0.13 & $-$0.02 & $-$0.07 & 0.03 & $-$0.06 & 0.16 & $-$0.11 & $-$0.07 & 0.01$^{\dagger}$ & $-$0.19 \\
Organic carbon (0--5cm) & 0.17 & $-$0.20 & \textbf{0.56} & 0.22 & 0.06 & $-$0.09 & $-$0.16 & $-$0.12 & 0.26 & 0.09 \\
Organic carbon (5--15cm) & 0.18 & $-$0.25 & \textbf{0.53} & 0.17 & 0.04 & $-$0.05 & $-$0.26 & $-$0.05 & 0.24 & 0.01$^{\dagger}$ \\
Organic carbon (depth gradient) & 0.07 & $-$0.03 & \textbf{0.38} & 0.20 & 0.07 & $-$0.13 & 0.08 & $-$0.20 & 0.18 & 0.19 \\
Water content (33kPa) (0--5cm) & $-$0.21 & $-$0.28 & \textbf{0.30} & \textbf{0.42} & $-$0.08 & 0.05 & $-$0.06 & $-$0.14 & 0.22 & 0.02$^{\dagger}$ \\
Water content (33kPa) (5--15cm) & $-$0.20 & $-$0.23 & 0.26 & \textbf{0.44} & $-$0.17 & 0.04 & $-$0.01$^{\dagger}$ & $-$0.19 & 0.23 & 0.08 \\
Water content (33kPa) (depth gradient) & $-$0.17 & \textbf{$-$0.38} & \textbf{0.34} & 0.18 & 0.27 & 0.06 & $-$0.25 & 0.08 & 0.09 & $-$0.19 \\
Water content (1500kPa) (0--5cm) & $-$0.26 & $-$0.09 & 0.05 & \textbf{0.42} & $-$0.29 & 0.04 & 0.13 & $-$0.24 & 0.18 & 0.10 \\
Water content (1500kPa) (5--15cm) & $-$0.27 & $-$0.11 & 0.05 & \textbf{0.41} & $-$0.28 & 0.05 & 0.13 & $-$0.24 & 0.17 & 0.09 \\
Water content (1500kPa) (depth gradient) & 0.12 & 0.30 & 0.03 & 0.03 & $-$0.15 & $-$0.08 & 0.01$^{\dagger}$ & 0.01$^{\dagger}$ & 0.04 & 0.21 \\
\end{longtable}
}

\FloatBarrier

\begin{table}[ht]
\centering \small
\caption{Three strongest environmental correlates of each principal component, with 95\%
bootstrap confidence intervals.}
\label{tab:pca_top}
\begin{tabular}{llccc}
\toprule
\textbf{Component} & \textbf{Strongest correlates} & \textbf{$r$} & \textbf{95\% CI} & \textbf{$p$} \\
\midrule
PC1 & Bulk density (0--5cm) & $-0.381$ & $[-0.400, -0.362]$ & $<10^{-100}$ \\
 & Clay content (5--15cm) & $-0.378$ & $[-0.397, -0.357]$ & $<10^{-100}$ \\
 & Clay content (0--5cm) & $-0.373$ & $[-0.393, -0.353]$ & $<10^{-100}$ \\
\addlinespace
PC2 & Bulk density (depth gradient) & $+0.444$ & $[+0.428, +0.460]$ & $<10^{-100}$ \\
 & Vapor Pr. (SD) & $-0.410$ & $[-0.432, -0.390]$ & $<10^{-100}$ \\
 & Elevation & $+0.377$ & $[+0.357, +0.398]$ & $<10^{-100}$ \\
\addlinespace
PC3 & Bio 11 (Mean Temp Coldest Qtr) & $-0.784$ & $[-0.792, -0.775]$ & $<10^{-100}$ \\
 & Avg Temp (Min) & $-0.783$ & $[-0.791, -0.774]$ & $<10^{-100}$ \\
 & Max Temp (Min) & $-0.778$ & $[-0.786, -0.769]$ & $<10^{-100}$ \\
\addlinespace
PC4 & pH (H$_2$O) (5--15cm) & $+0.473$ & $[+0.453, +0.495]$ & $<10^{-100}$ \\
 & pH (H$_2$O) (0--5cm) & $+0.472$ & $[+0.450, +0.493]$ & $<10^{-100}$ \\
 & Bulk density (5--15cm) & $+0.471$ & $[+0.447, +0.491]$ & $<10^{-100}$ \\
\addlinespace
PC5 & Elevation & $-0.498$ & $[-0.513, -0.483]$ & $<10^{-100}$ \\
 & Solar Rad. (Min) & $-0.434$ & $[-0.452, -0.415]$ & $<10^{-100}$ \\
 & Solar Rad. (Mean) & $-0.394$ & $[-0.414, -0.374]$ & $<10^{-100}$ \\
\addlinespace
PC6 & Sand content (depth gradient) & $-0.197$ & $[-0.216, -0.178]$ & $2.7\times10^{-71}$ \\
 & Silt content (depth gradient) & $+0.165$ & $[+0.144, +0.184]$ & $6.1\times10^{-50}$ \\
 & Solar Rad. (Max) & $+0.161$ & $[+0.142, +0.180]$ & $9.1\times10^{-48}$ \\
\addlinespace
PC7 & Solar Rad. (Mean) & $+0.478$ & $[+0.453, +0.500]$ & $<10^{-100}$ \\
 & Solar Rad. (Min) & $+0.471$ & $[+0.447, +0.493]$ & $<10^{-100}$ \\
 & Solar Rad. (Max) & $+0.421$ & $[+0.397, +0.445]$ & $<10^{-100}$ \\
\addlinespace
PC8 & Bio 18 (Precip Warmest Qtr) & $-0.373$ & $[-0.399, -0.347]$ & $<10^{-100}$ \\
 & Elevation & $-0.373$ & $[-0.400, -0.344]$ & $<10^{-100}$ \\
 & Solar Rad. (SD) & $+0.315$ & $[+0.293, +0.335]$ & $<10^{-100}$ \\
\addlinespace
PC9 & Organic carbon (0--5cm) & $+0.261$ & $[+0.240, +0.283]$ & $<10^{-100}$ \\
 & Nitrogen (5--15cm) & $+0.258$ & $[+0.234, +0.282]$ & $<10^{-100}$ \\
 & Organic carbon (5--15cm) & $+0.242$ & $[+0.221, +0.261]$ & $<10^{-100}$ \\
\addlinespace
PC10 & Solar Rad. (SD) & $-0.259$ & $[-0.282, -0.238]$ & $<10^{-100}$ \\
 & Sand content (depth gradient) & $+0.253$ & $[+0.232, +0.274]$ & $<10^{-100}$ \\
 & Bulk density (depth gradient) & $+0.214$ & $[+0.192, +0.237]$ & $2.9\times10^{-84}$ \\
\bottomrule
\end{tabular}
\end{table}

\FloatBarrier

The components differ in how strongly they align with anything measured. PC3 is the clearest axis in the analysis and tracks winter temperature ($r = -0.784$ with mean temperature of the coldest quarter, 95\% CI $[-0.792, -0.775]$) along with soil organic carbon ($+0.560$). PC4 aligns with soil chemistry, correlating with pH ($+0.473$) and bulk density ($+0.471$) at both depths. PC5 tracks elevation ($-0.498$) and incident radiation ($-0.434$), PC7 tracks solar radiation ($+0.478$), and PC1 and PC2 align most strongly with soil texture and bulk density ($-0.381$ and $+0.444$). PC6, PC9 and PC10 have no correlate above $0.27$ in absolute value and are not interpretable against this set of covariates.

\subsection{Variability under resampling}

We checked how much of the structure in Table~\ref{tab:pca_correlations} persists between resamples by refitting the decomposition independently on random halves of the samples, ten times, and comparing the resulting correlation profile of each component across the two halves. Agreement is high overall, with a median correlation between the two halves' profiles of 0.96 and 72\% of component-by-replicate comparisons exceeding 0.9, but it is not uniform across components.

PC1, PC2, PC3 and PC5 are stable (mean $|r| \geq 0.98$), whereas PC6, PC7 and PC8 are less so (0.74, 0.76 and 0.78). This tracks the spacing of the eigenvalues where the components that reproduce well are those separated from their neighbors, while PC6 through PC8 differ from one another by 0.3--0.5 percentage points of explained variance and are correspondingly free to rotate among themselves. The components carrying the strongest and most stable associations are therefore PC3, PC4, PC5 and PC7.

\section{Spatial cross-validation}
\label{note:spatialcv}

The main text reports performance under random 70/10/20 splits. Random splits allow a test sample to have a near neighbor in the training set, which can inflate apparent skill for any spatially structured target. We therefore repeated the satellite-only evaluation under three spatial cross validation schemes.

Block sizes were taken from the published autocorrelation ranges for the taxon rather than chosen for convenience. Ectomycorrhizal community structure autocorrelates to $8.5 \pm 1.0$m~\citep{tedersoo2010}, and soil fungal communities more generally to 2--3m, extending to roughly 17m for large genets~\citep{birch2023}.

We evaluated spatial blocks at both 25m and 50m, and additionally a grouped scheme in which samples within 25m of one another are joined by single linkage and assigned to the same fold, so that clusters of co-located samples are never split across the train/test boundary.

\begin{table}[ht]
\centering
\caption{Performance under random and spatially explicit cross-validation. LightGBM trained on all samples, five folds $\times$ three seeds. The last column is the median great-circle distance from each test sample to its nearest training sample.}
\label{tab:spatialcv}
\begin{tabular}{lcccc}
\toprule
\textbf{Scheme} & \textbf{$R^2$} & \textbf{MAE} & \textbf{Spearman $\rho$} & \textbf{Median train--test separation} \\
\midrule
Random & 0.507 $\pm$ 0.018 & 29.9 & 0.765 & 0m \\
Grouped, 25m & 0.316 $\pm$ 0.047 & 40.0 & 0.612 & 2,187m \\
Block, 25m & 0.296 $\pm$ 0.057 & 40.8 & 0.610 & 1,920m \\
Block, 50m & 0.298 $\pm$ 0.064 & 41.1 & 0.614 & 2,035m \\
\bottomrule
\end{tabular}
\end{table}

\FloatBarrier

The realized separation is much larger than the nominal block size since sampling is clustered into discrete study sites, so excluding a 25m block in practice excludes an entire site, and the median distance from a test sample to the nearest training sample is therefore about 2km, and the nominal block size understates the severity of the test. We also find that the result is not sensitive to the choice of scheme. The three spatial schemes agree to within 0.020 $R^2$. Fold-to-fold variability roughly triples under spatial cross-validation (standard deviation 0.047--0.064 against 0.018 for random splits). Rank correlation degrades far less than $R^2$ (0.765 to 0.61), indicating that the ordering of sites is largely preserved even where the absolute calibration is not.

\section{Sensitivity to the year of satellite imagery}
\label{note:year}

Field samples in GlobalFungi were collected over a range of years, while embeddings in the main analysis are extracted from a single year of Sentinel-1 and Sentinel-2 imagery. To test whether that mismatch affects the results we extracted embeddings for two other years as well for all European samples.

\begin{table}[ht]
\centering
\caption{Effect of embedding year. Satellite-only, LightGBM, five seeds, on European samples. Within-year trains and tests on the same year.}
\label{tab:yearablation}
\begin{tabular}{lccc}
\toprule
 & \textbf{2022} & \textbf{2023} & \textbf{2024} \\
\midrule
Within-year $R^2$ & 0.4905 $\pm$ 0.0243 & 0.4995 $\pm$ 0.0099 & 0.5000 $\pm$ 0.0172 \\
Within-year MAE & 33.58 & 33.59 & 33.36 \\
\midrule
Mean cosine similarity to 2024 embeddings & 0.872 & 0.892 & --- \\
\bottomrule
\end{tabular}
\end{table}

\FloatBarrier

Cross-sectional skill is effectively invariant to the choice of year. The spread of within-year $R^2$ across the three years is 0.0096, against a seed-to-seed standard deviation of 0.0171 for a single year, that is, changing the imagery year moves performance less than reshuffling the train/test split does.

\section{Patch size ablation}
\label{note:patchsize}

The main analysis uses a 3$\times$3 pixel window (30m $\times$ 30m) centred on each sample. The choice balances two constraints. The window must be wide enough to absorb the combined geolocation error of the Sentinel products and of the recorded sample coordinates, and it is of the order of a single tree crown, which is the relevant scale for an obligate symbiont of trees. It must also remain narrow enough not to average across adjacent habitat.

To test the choice empirically we evaluated windows from 1$\times$1 to 7$\times$7 with a subsample under the protocol used throughout the paper. Smaller windows are exact centre-crops of a single fetched 7$\times$7 cache, so every window is evaluated on identical locations and identical underlying tile data, and every fold assignment is shared across windows. Comparisons are therefore paired, and the difference in per-sample error can be bootstrapped directly. Only the step from a single pixel to 3$\times$3 is significant. Both wider windows improve the point estimate slightly, and both confidence intervals include zero, so spatial context contributes but saturates at the width used.

\begin{table}[ht]
\centering
\caption{Patch-size ablation. All sizes are center-crops of the same 7$\times$7 patch, so the samples and folds are identical across rows and comparisons are paired. 784 locations across five seeds. $\Delta$MAE is against the 3$\times$3 window used throughout the paper.}
\label{tab:patchsize}
\begin{tabular}{lrccc c}
\toprule
\textbf{Window} & \textbf{Features} & \textbf{$R^2$} & \textbf{MAE} & \textbf{Spearman $\rho$} & \textbf{$\Delta$MAE vs 3$\times$3 (95\% CI)} \\
\midrule
1$\times$1 & 128 & 0.219 $\pm$ 0.057 & 40.78 & 0.652 & $+2.87$ $[+1.43, +4.27]$ \\
3$\times$3 & 1{,}152 & 0.303 $\pm$ 0.034 & 37.91 & 0.680 & --- \\
5$\times$5 & 3{,}200 & 0.320 $\pm$ 0.020 & 37.32 & 0.687 & $-0.59$ $[-1.69, +0.53]$ \\
7$\times$7 & 6{,}272 & 0.340 $\pm$ 0.015 & 36.90 & 0.700 & $-1.01$ $[-2.23, +0.20]$ \\
\bottomrule
\end{tabular}
\end{table}

\FloatBarrier

\section{Comparison with a spectral-index baseline}
\label{note:spectral}

Except for WorldCover, the environmental baselines in the main comparison are coarser than 10m, which leaves open whether the advantage of the learned features is due to the representation or simply to spatial resolution. To separate these we built a baseline from the same imagery at the same resolution, so that only the representation differs.

For each sample we computed 90 features per pixel directly from the same Sentinel-1 and Sentinel-2 stacks that the embeddings were extracted from, over the same 3$\times$3 window and under the same protocol and the same ten random seeds. Observations flagged as cloud-affected by the Sentinel-2 scene classification layer are set to missing rather than zero, so they drop out of the statistics instead of biasing them downward. The same applies to absent Sentinel-1 acquisitions. Table~\ref{tab:spectralfeatures} lists the feature families. Because both feature sets are computed from the same sensors, the same pixels, the same acquisition dates and the same window, and are evaluated with the same downstream model on the same folds and seeds, the only quantity that varies between them being the representation.

The phenology terms are non-parametric because fitting a double-logistic curve to recover green-up and senescence dates is unreliable on sparse, cloud-gapped series, which is the same difficulty the embedding model addresses through its temporal sampling, so we summarize the trajectory by the timing of its extremes, its integral and its roughness instead. Where a pixel has no clear observation in a given channel, that channel is imputed with the patch mean.

\begin{table}[ht]
\centering \small
\caption{Composition of the spectral-index baseline, 90 features per pixel. Statistics are computed over the cloud-free observations of a single year, the same year from which the embeddings are extracted.}
\label{tab:spectralfeatures}
\begin{tabular}{p{4.6cm} c p{6.4cm}}
\toprule
\textbf{Family} & \textbf{$n$} & \textbf{Definition} \\
\midrule
Sentinel-2 band statistics & 50 & Mean, median, 10th percentile, 90th percentile and standard deviation of each of the ten bands (B2, B3, B4, B5, B6, B7, B8, B8A, B11, B12) \\
\addlinespace
Vegetation indices & 24 & NDVI, EVI, NDWI and NBR. For each, same five statistics and a seasonal amplitude term defined as the 90th minus the 10th percentile \\
\addlinespace
NDVI phenology & 4 & Day of year of the annual maximum and of the annual minimum, the integrated annual NDVI, and the mean absolute successive difference between consecutive observations as a measure of trajectory roughness \\
\addlinespace
Sentinel-1 backscatter & 12 & Mean, median and standard deviation of VV and VH, computed separately for ascending and descending orbits \\
\bottomrule
\end{tabular}
\end{table}

\FloatBarrier

\begin{table}[ht]
\centering
\caption{Learned embeddings against a spectral-index baseline computed from the same imagery at the same resolution. LightGBM, ten seeds.}
\label{tab:spectral}
\begin{tabular}{lcc}
\toprule
\textbf{Feature set} & \textbf{$R^2$} & \textbf{MAE} \\
\midrule
Tessera embeddings & 0.5094 $\pm$ 0.0154 & 30.06 \\
Spectral indices and statistics & 0.4866 $\pm$ 0.0230 & 30.89 \\
\bottomrule
\end{tabular}
\end{table}

\FloatBarrier

Because the seeds are shared, the comparison is paired. The mean paired difference is $+0.0228$ $R^2$ (95\% confidence interval $[+0.0060, +0.0396]$), the embeddings lead on eight of ten seeds, and a Wilcoxon signed-rank test gives $p = 0.014$. Mean absolute error separates the two on ten of ten seeds, by 0.83 species. Hand-engineered spectral indices are therefore a competitive baseline, and are consistently beaten by approximately the same margin as the embeddings beat WorldClim in the main comparison. This is the expected outcome if the mechanistic linkage runs through vegetation phenology, since the embeddings encode the full time series whereas the indices summarize it.

\section{Cross-continental transfer}
\label{note:transfer}

The models in the main text pool European and Asian samples. To see if models can learn broadly generalized ECM biodiversity, we trained models on one continent and tested on the other and find that transfer fails in both directions. The asymmetry in magnitude reflects the difference in richness distributions. European richness is substantially higher, so a model trained there overpredicts Asian sites by a large absolute margin.

Ectomycorrhizal species pools are structured regionally by host tree availability and biogeographic barriers, so absolute richness in one region carries little information about absolute richness in another, and successful transfer would have been the surprising result. The practical implication is that the pooled model reported here should be understood as interpolating within the environmental and biogeographic range it was fitted on, and that extending it to an unsampled region requires samples from that region rather than transfer alone. This holds for every feature set we test, not only the learned one. Every feature set fails to be useful predictors, so the limitation is therefore a property of the target rather than of the representation.

\begin{table}[ht]
\centering
\caption{Cross-continental transfer for each feature set. all samples, LightGBM, three seeds.}
\label{tab:transfer_by_featureset}
\begin{tabular}{lcccc}
\toprule
& \multicolumn{2}{c}{\textbf{Transferred ($R^2$)}}
& \multicolumn{2}{c}{\textbf{Transferred ($\rho$)}} \\
\cmidrule(lr){2-3}\cmidrule(lr){4-5}
\textbf{Feature set} & eur $\rightarrow$ asia & asia $\rightarrow$ eur
& eur $\rightarrow$ asia & asia $\rightarrow$ eur \\
\midrule
Satellite & $-9.8 \pm 2.3$ & $-0.051 \pm 0.045$ & $+0.04 \pm 0.05$ & $+0.10 \pm 0.07$ \\
Climate & $-8.2 \pm 1.9$ & $-0.221 \pm 0.038$ & $+0.14 \pm 0.02$ & $-0.17 \pm 0.11$ \\
Soil & $-13.3 \pm 3.4$ & $-0.023 \pm 0.029$ & $-0.06 \pm 0.06$ & $+0.22 \pm 0.03$ \\
Land cover & $-14.5 \pm 1.9$ & $-0.280 \pm 0.085$ & $-0.01 \pm 0.06$ & $+0.17 \pm 0.04$ \\
\bottomrule
\end{tabular}
\end{table}

\FloatBarrier

\section{Prediction intervals}
\label{note:intervals}

The predictions in the main text are point estimates. To see whether we can quantify uncertainty we refitted the satellite-only model under pinball loss at the 5th, 50th and 95th percentiles, giving a nominal 90\% interval for every sample. Inference is point-wise, so the same three models applied over a raster produce a wall-to-wall interval layer without further methodology.

Over five seeds and all samples, the nominal 90\% intervals cover $85.4\% \pm 1.3\%$ of held-out observations at a mean width of 118 species (median 103). Per-quantile pinball losses are 2.74, 14.34 and 6.59 for the 5th, 50th and 95th percentiles. The intervals are thus somewhat narrow, by 4.6 percentage points of coverage. Closing that gap would require conformal or held-out recalibration, which is a modeling contribution beyond the scope of this work, as we focus on feature set comparisons rather than model architectures. Interval width correlates $+0.470 \pm 0.006$ with absolute error, so wide intervals do mark the samples the model predicts poorly.

\begin{table}[ht]
\centering
\caption{Interval behavior by quartile of predicted richness.}
\label{tab:intervals}
\begin{tabular}{lcccc}
\toprule
\textbf{Quartile of predicted richness} & \textbf{$n$} & \textbf{Coverage} & \textbf{Mean width} & \textbf{Mean absolute error} \\
\midrule
1 (lowest) & 2,961 & 0.837 & 60.7 & 6.7 \\
2 & 2,961 & 0.863 & 77.4 & 18.4 \\
3 & 2,961 & 0.870 & 131.6 & 34.7 \\
4 (highest) & 2,961 & 0.842 & 202.4 & 71.5 \\
\bottomrule
\end{tabular}
\end{table}

\FloatBarrier

\section{Error structure and outlier sensitivity}
\label{note:higherror}
\label{note:outliers} 

\subsection{Sensitivity of model fit to prediction outliers}

\begin{figure}[htbp]
\centering
\includegraphics[width=0.85\textwidth]{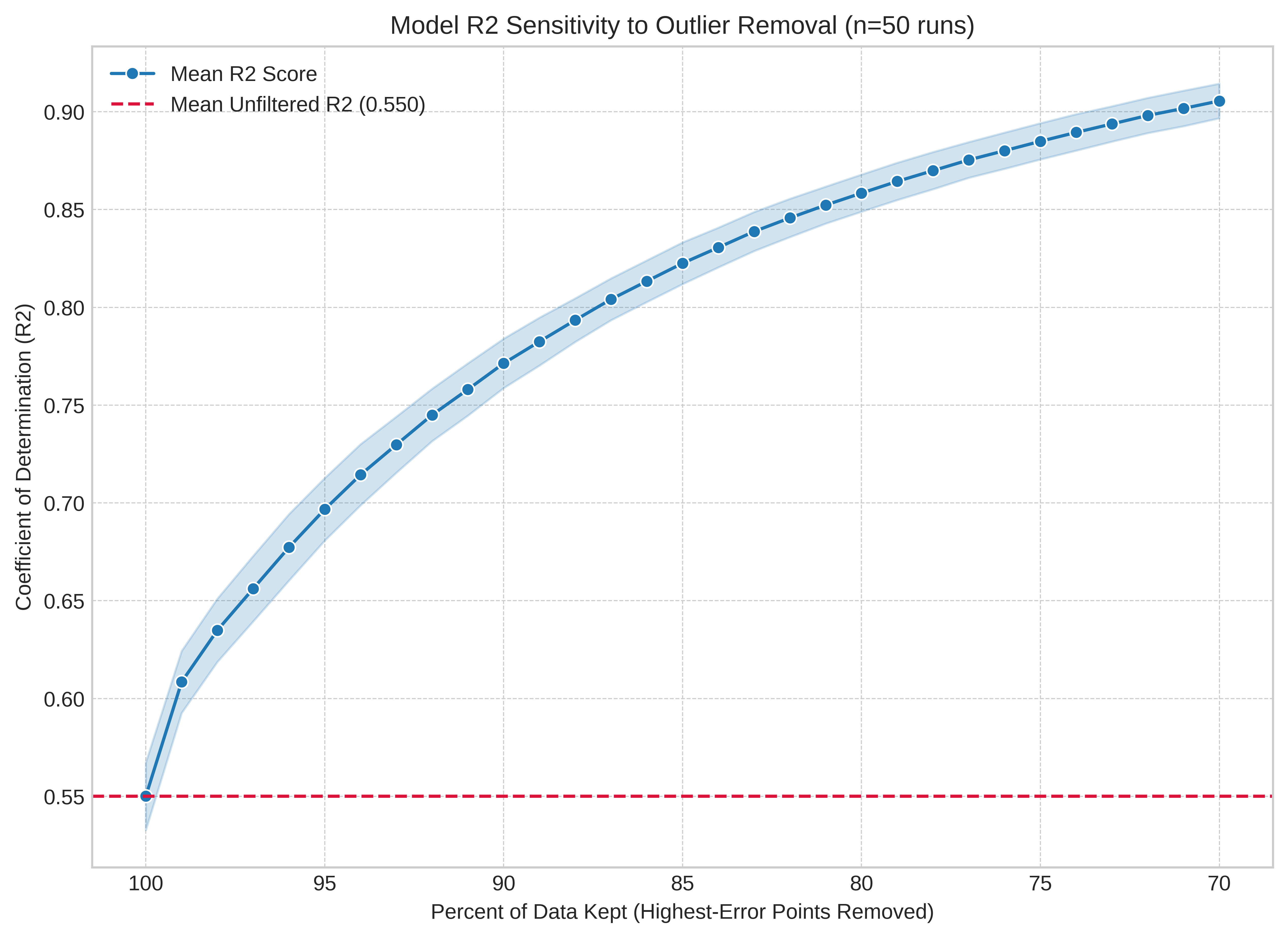}
\caption{Diagnosing the impact of prediction outliers on model performance. This sensitivity analysis plots the model's mean $R^2$ ($y$-axis) as a function of the percentage of the test set retained after progressively removing the highest-error points ($x$-axis). Data are from the best-performing model (LightGBM, all features) over 50 runs; the shaded area shows the standard deviation.}
\label{fig:sensitivitycurve}
\end{figure}

\FloatBarrier

\subsection{Characterization of high-error samples}

Figure~\ref{fig:sensitivitycurve} shows that a small fraction of samples accounts for a disproportionate share of total error. Here we ask what distinguishes those samples, and in particular whether any property observable before ground truth is available can identify them.

Taking the top decile of absolute error ($n = 1{,}185$, carrying 39.7\% of all absolute error), high-error samples differ from the rest in two respects that are robust to how outliers are defined:

\begin{itemize}
\item \textbf{Sampling density.} High-error samples sit in sparser neighborhoods, with a mean of 8.0 other samples within 1km against 87.3 for the remainder (Cohen's $d = -0.54$, 95\% CI $[-0.56, -0.53]$).
\item \textbf{Representation novelty.} They are more distant from the training distribution in embedding space, measured as mean distance to their $k$ nearest training neighbors ($d = +0.83$).
\end{itemize}

Both hold whether outliers are defined by absolute error or by error relative to observed richness, and both are measurable without ground truth. As an applicability-domain diagnostic, we ask what share of total absolute error is recovered by flagging the worst decile under each label-free criterion. A value of 1.0 corresponds to no information.

\begin{table}[ht]
\centering
\caption{Share of total absolute error captured by flagging the top decile under
each diagnostic, expressed as a multiple of the 10\% expected by chance.}
\label{tab:capture}
\begin{tabular}{lc}
\toprule
\textbf{Diagnostic (top decile)} & \textbf{Lift} \\
\midrule
Highest predicted richness & 1.94 \\
Lowest sampling density within 1km & 1.67 \\
Greatest novelty in embedding space & 1.65 \\
Greatest distance to nearest training sample & 1.60 \\
\bottomrule
\end{tabular}
\end{table}

\FloatBarrier

\section{Full ablation results and paired tests}
\label{note:table1}

Supplementary Table~\ref{tab:mainresults} gives the complete results summarized in Figure~\ref{fig:mainresults} of the main text, with root mean squared error, mean absolute error and mean error in addition to $R^2$.

\begin{table}[ht]
\centering \small
\caption{Performance of models and feature set ablations for predicting fungal
biodiversity. Metrics are mean $\pm$ standard deviation over 50 runs varying the random seed of the data split. The best value in each column is in bold. \textbf{Data sources:} Sat = satellite SSL features; Clim = WorldClim; Soil = SoilGrids; WC = ESA WorldCover. \textbf{Metrics:} $R^2$ = coefficient of determination; RMSE = root mean squared error; MAE = mean absolute error; ME = mean error (bias).}
\label{tab:mainresults}
\begin{tabular}{@{}llcccc@{}}
\toprule
\textbf{Data Source(s)} & \textbf{Model} & \textbf{$R^2$} & \textbf{RMSE} & \textbf{MAE} & \textbf{ME (Bias)} \\
\midrule
\multicolumn{6}{c}{\textit{Full Models (4 Feature Sets)}} \\
\midrule
Sat + Clim + Soil + WC & LightGBM & \bfseries 0.550 $\pm$ 0.017 & \bfseries 46.22 $\pm$ 1.54 & 28.29 $\pm$ 0.70 & -0.07 $\pm$ 0.97 \\
 & XGBoost & 0.545 $\pm$ 0.017 & 46.47 $\pm$ 1.58 & 28.55 $\pm$ 0.76 & -0.09 $\pm$ 0.97 \\
 & Random Forest & 0.529 $\pm$ 0.018 & 47.27 $\pm$ 1.50 & 28.34 $\pm$ 0.77 & 1.07 $\pm$ 0.91 \\
\midrule
\multicolumn{6}{c}{\textit{Single-Feature-Set Models (Ablations)}} \\
\midrule
Satellite only & LightGBM & 0.535 $\pm$ 0.018 & 47.01 $\pm$ 1.61 & 29.07 $\pm$ 0.78 & \bfseries 0.03 $\pm$ 0.96 \\
 & XGBoost & 0.529 $\pm$ 0.017 & 47.30 $\pm$ 1.58 & 29.33 $\pm$ 0.78 & 0.10 $\pm$ 0.98 \\
 & Random Forest & 0.512 $\pm$ 0.017 & 48.16 $\pm$ 1.38 & 29.20 $\pm$ 0.72 & 1.43 $\pm$ 0.98 \\
\addlinespace
Climate only & LightGBM & 0.517 $\pm$ 0.018 & 47.56 $\pm$ 1.54 & 28.87 $\pm$ 0.66 & -0.22 $\pm$ 1.14 \\
 & XGBoost & 0.514 $\pm$ 0.019 & 47.75 $\pm$ 1.59 & 29.17 $\pm$ 0.73 & -0.21 $\pm$ 1.13 \\
 & Random Forest & 0.493 $\pm$ 0.026 & 48.74 $\pm$ 1.74 & 28.25 $\pm$ 0.73 & -0.29 $\pm$ 1.19 \\
\addlinespace
Soil only & LightGBM & 0.514 $\pm$ 0.017 & 47.72 $\pm$ 1.58 & 29.24 $\pm$ 0.75 & -0.33 $\pm$ 1.13 \\
 & XGBoost & 0.506 $\pm$ 0.017 & 48.14 $\pm$ 1.65 & 29.51 $\pm$ 0.83 & -0.37 $\pm$ 1.14 \\
 & Random Forest & 0.496 $\pm$ 0.023 & 48.60 $\pm$ 1.73 & 28.34 $\pm$ 0.80 & -0.65 $\pm$ 1.25 \\
\addlinespace
WorldCover only & LightGBM & 0.513 $\pm$ 0.018 & 47.74 $\pm$ 1.54 & 29.06 $\pm$ 0.72 & -0.03 $\pm$ 1.04 \\
 & XGBoost & 0.515 $\pm$ 0.018 & 47.66 $\pm$ 1.59 & 29.04 $\pm$ 0.97 & -0.08 $\pm$ 1.10 \\
 & Random Forest & 0.484 $\pm$ 0.024 & 49.15 $\pm$ 1.71 & \bfseries 28.04 $\pm$ 0.73 & -0.41 $\pm$ 1.14 \\
\bottomrule
\end{tabular}
\end{table}

\FloatBarrier

All fifteen configurations were evaluated on the same vector of 50 random seeds, so comparison between them is paired and the variance contributed by the split cancels.

Satellite-only exceeds every environmental baseline in all nine model $\times$ baseline comparisons, with every confidence interval excluding zero and differences ranging from $+0.014$ to $+0.028$ $R^2$.

\begin{table}[ht]
\centering
\caption{Paired comparison of the satellite-only configuration against each single
environmental baseline. Differences are satellite minus baseline, in $R^2$, over 50 shared seeds.}
\label{tab:paired}
\begin{tabular}{llccc}
\toprule
\textbf{Baseline} & \textbf{Model} & \textbf{Mean difference} & \textbf{95\% CI} & \textbf{Wilcoxon $p$} \\
\midrule
Climate only & LightGBM & $+$0.0174 & $[+0.0110, +0.0238]$ & $8.2\times10^{-7}$ \\
 & XGBoost & $+$0.0153 & $[+0.0082, +0.0223]$ & $9.0\times10^{-5}$ \\
 & Random Forest & $+$0.0185 & $[+0.0106, +0.0265]$ & $4.0\times10^{-5}$ \\
\addlinespace
Soil only & LightGBM & $+$0.0204 & $[+0.0137, +0.0270]$ & $2.0\times10^{-7}$ \\
 & XGBoost & $+$0.0232 & $[+0.0163, +0.0300]$ & $4.2\times10^{-8}$ \\
 & Random Forest & $+$0.0157 & $[+0.0080, +0.0233]$ & $1.7\times10^{-4}$ \\
\addlinespace
WorldCover only & LightGBM & $+$0.0215 & $[+0.0143, +0.0288]$ & $2.8\times10^{-7}$ \\
 & XGBoost & $+$0.0140 & $[+0.0069, +0.0211]$ & $3.0\times10^{-4}$ \\
 & Random Forest & $+$0.0276 & $[+0.0193, +0.0359]$ & $6.9\times10^{-9}$ \\
\bottomrule
\end{tabular}
\end{table}

\FloatBarrier

\bibliography{references}

@ARTICLE{Horn1981hill,
  author={Horn, B.K.P.},
  journal={Proceedings of the IEEE}, 
  title={Hill shading and the reflectance map}, 
  year={1981},
  volume={69},
  number={1},
  pages={14-47},
  doi={10.1109/PROC.1981.11918}}

@article{Fisher2016TreeMA,
author = {Fisher, Joshua B. and Sweeney, Sean and Brzostek, Edward R. and Evans, Tom P. and Johnson, Daniel J. and Myers, Jonathan A. and Bourg, Norman A. and Wolf, Amy T. and Howe, Robert W. and Phillips, Richard P.},
title = {Tree-mycorrhizal associations detected remotely from canopy spectral properties},
journal = {Global Change Biology},
volume = {22},
number = {7},
pages = {2596-2607},
doi = {https://doi.org/10.1111/gcb.13264},
url = {https://onlinelibrary.wiley.com/doi/abs/10.1111/gcb.13264},
eprint = {https://onlinelibrary.wiley.com/doi/pdf/10.1111/gcb.13264},
year = {2016}
}

@article{Davison2015GlobalAO,
  author    = {J. Davison and M. Moora and M. \"Opik and A. Adholeya and L. Ainsaar and A. B\^{a} and S. Burla and A. G. Diedhiou and I. Hiiesalu and T. Jairus and N. C. Johnson and A. Kane and K. Koorem and M. Kochar and C. Ndiaye and M. P\"artel and \"U. Reier and \"U. Saks and R. Singh and M. Vasar and M. Zobel},
  title     = {Global assessment of arbuscular mycorrhizal fungus diversity reveals very low endemism},
  journal   = {Science},
  volume    = {349},
  number    = {6251},
  pages     = {970--973},
  year      = {2015},
  doi       = {10.1126/science.aab1161},
  url       = {https://www.science.org/doi/abs/10.1126/science.aab1161}
}

@article{CastroSanchez2024TreeAM,
author = {Castro Sánchez-Bermejo, Pablo and Monjau, Tilo and Goldmann, Kezia and Ferlian, Olga and Eisenhauer, Nico and Bruelheide, Helge and Ma, Zeqing and Haider, Sylvia},
title = {Tree and mycorrhizal fungal diversity drive intraspecific and intraindividual trait variation in temperate forests: Evidence from a tree diversity experiment},
journal = {Functional Ecology},
volume = {38},
number = {5},
pages = {1089-1103},
doi = {https://doi.org/10.1111/1365-2435.14549},
year = {2024}
}

@article{VanNuland2025,
  author       = {Van Nuland, Michael E. and Averill, Colin and Stewart, Justin D. and Prylutskyi, Oleh and Corrales, Adriana and van Galen, Laura G. and Manley, Bethan F. and Qin, Clara and Lauber, Thomas and Mikryukov, Vladimir and Dulia, Olesia and Furci, Giuliana and Marín, César and Sheldrake, Merlin and Weedon, James T. and Peay, Kabir G. and Cornwallis, Charlie K. and Větrovský, Tomáš and Kohout, Petr and Baldrian, Petr and Tedersoo, Leho and West, Stuart A. and Crowther, Thomas W. and Kiers, E. Toby and SPUN Mapping Consortium and van den Hoogen, Johan},
  title        = {Global hotspots of mycorrhizal fungal richness are poorly protected},
  journal      = {Nature},
  year         = {2025},
  doi          = {10.1038/s41586-025-09277-4},
  url          = {https://doi.org/10.1038/s41586-025-09277-4}
}

@article{Thers2017LidarderivedVA,
	title = {Lidar-derived variables as a proxy for fungal species richness and composition in temperate {Northern} {Europe}},
	volume = {200},
	issn = {00344257},
	url = {https://linkinghub.elsevier.com/retrieve/pii/S0034425717303656},
	doi = {10.1016/j.rse.2017.08.011},
	language = {en},
	journal = {Remote Sensing of Environment},
  author={Henrik Thers and Ane Kirstine Brunbjerg and Thomas L{\ae}ss{\o}e and Rasmus Ejrn{\ae}s and Peder Klith B{\o}cher and Jens‐Christian Svenning},
	month = oct,
	year = {2017},
	pages = {102--113},
}

@article{CavenderBares2021RemotelyDA,
	title = {Remotely detected aboveground plant function predicts belowground processes in two prairie diversity experiments},
	volume = {92},
	issn = {0012-9615, 1557-7015},
	url = {https://esajournals.onlinelibrary.wiley.com/doi/10.1002/ecm.1488},
	doi = {10.1002/ecm.1488},
	language = {en},
	number = {1},
	journal = {Ecological Monographs},
	author = {Cavender‐Bares, Jeannine and Schweiger, Anna K. and Gamon, John A. and Gholizadeh, Hamed and Helzer, Kimberly and Lapadat, Cathleen and Madritch, Michael D. and Townsend, Philip A. and Wang, Zhihui and Hobbie, Sarah E.},
	month = feb,
	year = {2022},
	pages = {e01488},
}

@misc{feng2025tesseratemporalembeddingssurface,
    author    = {Feng, Zhengpeng and Atzberger, Clement and Jaffer, Sadiq and Knezevic, Jovana and Sormunen, Silja and Young, Robin and Lisaius, Madeline C. and Immitzer, Markus and Jackson, Toby and Ball, James and Coomes, David A. and Madhavapeddy, Anil and Blake, Andrew and Keshav, Srinivasan},
    title     = {TESSERA: Temporal Embeddings of Surface Spectra for Earth Representation and Analysis},
    booktitle = {Proceedings of the IEEE/CVF Conference on Computer Vision and Pattern Recognition (CVPR)},
    month     = {June},
    year      = {2026},
    url = {https://openaccess.thecvf.com/content/CVPR2026/papers/Feng_TESSERA_Temporal_Embeddings_of_Surface_Spectra_for_Earth_Representation_and_CVPR_2026_paper.pdf},
    pages     = {34818-34831}
}

@article{Vtrovsk2019AMO,
	title = {A meta-analysis of global fungal distribution reveals climate-driven patterns},
	volume = {10},
	issn = {2041-1723},
	url = {https://www.nature.com/articles/s41467-019-13164-8},
	doi = {10.1038/s41467-019-13164-8},
	language = {en},
	number = {1},
	journal = {Nature Communications},
  author    = {Tom{\'a}{\v{s}} V{\v{e}}trovsk{\'y} and Petr Kohout and Martin Kopeck{\'y} and Anton{\'i}n Machac and Mat{\v{e}}j Man and Barbara Doreen Bahnmann and Vendula Brabcov{\'a} and Jinlyung Choi and Lenka M{\'e}sz{\'a}ro{\v{s}}ov{\'a} and Zander Rainier Human and Cl{\'e}mentine Lepinay and Salvador Llad{\'o} and Rub{\'e}n L{\'o}pez-Mond{\'e}jar and Tijana Martinovi{\'c} and Tereza Ma{\v{s}}{\'i}nov{\'a} and Daniel Kumazawa Morais and Diana Navr{\'a}tilov{\'a} and I{\~n}aki Odriozola and Martina {\v{S}}tursov{\'a} and Karel {\v{S}}vec and Vojt{\v{e}}ch Tl{\'a}skal and Michaela Urbanov{\'a} and Joe Wan and Lucia {\v{Z}}if{\v{c}}{\'a}kov{\'a} and Adina C. Howe and Joshua Ladau and Kabir Gabriel Peay and David Storch and Jan Wild and Petr Baldrian},
	month = nov,
	year = {2019},
	pages = {5142},
}

@article{Lisaius2024UsingBT,
  author={Lisaius, Madeline C and Blake, Andrew and Keshav, Srinivasan and Atzberger, Clement},
  journal={IEEE Journal of Selected Topics in Applied Earth Observations and Remote Sensing}, 
  title={Using Barlow Twins to Create Representations From Cloud-Corrupted Remote Sensing Time Series}, 
  year={2024},
  volume={17},
  number={},
  pages={13162-13168},
  doi={10.1109/JSTARS.2024.3426044}}

@InProceedings{BTzbontar,
  title = 	 {Barlow Twins: Self-Supervised Learning via Redundancy Reduction},
  author =       {Zbontar, Jure and Jing, Li and Misra, Ishan and LeCun, Yann and Deny, Stephane},
  booktitle = 	 {Proceedings of the 38th International Conference on Machine Learning},
  pages = 	 {12310--12320},
  year = 	 {2021},
  editor = 	 {Meila, Marina and Zhang, Tong},
  volume = 	 {139},
  series = 	 {Proceedings of Machine Learning Research},
  month = 	 {18--24 Jul},
  publisher =    {PMLR},
  url = 	 {https://proceedings.mlr.press/v139/zbontar21a.html},
}

@article{Gholizadeh2018RemoteSO,
	title = {Remote sensing of biodiversity: {Soil} correction and data dimension reduction methods improve assessment of $\alpha$-diversity (species richness) in prairie ecosystems},
	volume = {206},
	issn = {00344257},
	doi = {10.1016/j.rse.2017.12.014},
	language = {en},
	journal = {Remote Sensing of Environment},
	author = {Gholizadeh, Hamed and Gamon, John A. and Zygielbaum, Arthur I. and Wang, Ran and Schweiger, Anna K. and Cavender-Bares, Jeannine},
	month = mar,
	year = {2018},
	pages = {240--253},
}

@article{Torresani2024ReviewingTS,
    author    = {Michele Torresani and Christian Rossi and Michela Perrone and Leon T. Hauser and Jean‐Baptiste F{\'e}ret and V{\'\i}t{\v{e}}zslav Moudr{\'y} and Petra {\v{S}}{\'\i}mov{\'a} and Carlo Ricotta and Giles M. Foody and Patrick Kacic and Hannes Feilhauer and Marco Malavasi and Roberto Tognetti and Duccio Rocchini},
    title = {Reviewing the {Spectral} {Variation} {Hypothesis}: {Twenty} years in the tumultuous sea of biodiversity estimation by remote sensing},
    volume = {82},
    issn = {15749541},
    shorttitle = {Reviewing the {Spectral} {Variation} {Hypothesis}},
    doi = {10.1016/j.ecoinf.2024.102702},
    language = {en},
    journal = {Ecological Informatics},
    month = sep,
    year = {2024},
    pages = {102702},
    }

@article{Madritch2014ImagingSL,
	title = {Imaging spectroscopy links aspen genotype with below-ground processes at landscape scales},
	volume = {369},
	issn = {0962-8436, 1471-2970},
	doi = {10.1098/rstb.2013.0194},
	language = {en},
	number = {1643},
	journal = {Philosophical Transactions of the Royal Society B: Biological Sciences},
	author = {Madritch, Michael D. and Kingdon, Clayton C. and Singh, Aditya and Mock, Karen E. and Lindroth, Richard L. and Townsend, Philip A.},
	month = may,
	year = {2014},
	pages = {20130194},
}

@article{Li2015AbovegroundbelowgroundBL,
	title = {Aboveground-belowground biodiversity linkages differ in early and late successional temperate forests},
	volume = {5},
	issn = {2045-2322},
	url = {https://www.nature.com/articles/srep12234},
	doi = {10.1038/srep12234},
	language = {en},
	number = {1},
	journal = {Scientific Reports},
	author = {Li, Hui and Wang, Xugao and Liang, Chao and Hao, Zhanqing and Zhou, Lisha and Ma, Sam and Li, Xiaobin and Yang, Shan and Yao, Fei and Jiang, Yong},
	month = jul,
	year = {2015},
	pages = {12234},
}

@article{Li2018SoilMB,
	title = {Soil microbial beta-diversity is linked with compositional variation in aboveground plant biomass in a semi-arid grassland},
	volume = {423},
	issn = {0032-079X, 1573-5036},
	url = {http://link.springer.com/10.1007/s11104-017-3524-2},
	doi = {10.1007/s11104-017-3524-2},
	language = {en},
	number = {1-2},
	journal = {Plant and Soil},
	author = {Li, Hui and Xu, Zhuwen and Yan, Qingyun and Yang, Shan and Van Nostrand, Joy D. and Wang, Zhirui and He, Zhili and Zhou, Jizhong and Jiang, Yong and Deng, Ye},
	month = feb,
	year = {2018},
	pages = {465--480},
}

@article{Lang2023ForestSD,
  title={Forest structural diversity is linked to soil microbial diversity},
  author={Ashley K. Lang and Elizabeth A. LaRue and Stephanie N. Kivlin and Joseph D. Edwards and Richard P. Phillips and Joey Gallion and Nicole Kong and John D. Parker and Melissa K. McCormick and Grant M Domke and Songlin Fei},
  journal={Ecosphere},
  volume={14},
  number={11},
  pages={e4702},
  year={2023},
  doi={10.1002/ecs2.4702},
  url={https://doi.org/10.1002/ecs2.4702}
}

@article{Jakubik2023FoundationMF,
  title={Foundation Models for Generalist Geospatial Artificial Intelligence},
  author={Johannes Jakubik and Sujit Roy and Christopher Phillips and Paolo Fraccaro and Denys Godwin and Bianca Zadrozny and Daniel Szwarcman and Carlos Gomes and Gabby Nyirjesy and Blair Edwards and Daiki Kimura and Naomi Simumba and Linsong Chu and S. Karthik Mukkavilli and Devyani Lambhate and Kamal Das and Ranjini Bangalore and Dario Oliveira and Michal Muszynski and Kumar Ankur and Muthukumaran Ramasubramanian and Iksha Gurung and Sam Khallaghi and Hanxi Li and Michael Cecil and Maryam Ahmadi and Fatemeh Kordi and Hamed Alemohammad and Manil Maskey and Raghu Kiran Ganti and Kommy Weldemariam and Rahul Ramachandran},
  journal={ArXiv},
  year={2023},
  volume={abs/2310.18660},
  url={https://api.semanticscholar.org/CorpusID:264590307}
}

@article{DeLuca2024MycorrhizalFR,
  title={Mycorrhizal fungi respiration dynamics in relation to gross primary production in a Hungarian dry grassland},
  author={Giulia De Luca and Marianna Papp and Szilvia F{\'o}ti and Katalin Posta and {\'A}d{\'a}m M{\'e}sz{\'a}ros and Krisztina Pint{\'e}r and Zolt{\'a}n Nagy and Evelin Ram{\'o}na P{\'e}li and S{\'a}ndor Fekete and J{\'a}nos Balogh},
  journal={Plant and Soil},
  year={2024},
  doi={10.1007/s11104-024-06557-2},
  url={https://doi.org/10.1007/s11104-024-06557-2}
}

@article{VzquezSantos2024ArbuscularMF,
  title={Arbuscular mycorrhizal fungi affect early phenological stages of three secondary vegetation species in a temperate forest},
  author={Yasmin V{\'a}zquez-Santos and Silvia Castillo-Arg{\"u}ero and No{\'e} Manuel Monta{\~n}o and Francisco Javier Espinosa-Garc{\'i}a and Cesar Mateo Flores-Ort{\'i}z and Yuriana Mart{\'i}nez-Orea},
  journal={Plant Ecology},
  volume={225},
  pages={983--996},
  year={2024},
  doi={10.1007/s11258-024-01448-z},
  url={https://doi.org/10.1007/s11258-024-01448-z}
}

@article{Chu2021LongTN,
  title={Long time-series NDVI reconstruction in cloud-prone regions via spatio-temporal tensor completion},
  author={Chu, Dong and Shen, Huanfeng and Guan, Xiaobin and Chen, James M. and Li, Xinghua and Li, Jie and Zhang, Liangpei},
  journal={Remote Sensing of Environment},
  volume={264},
  pages={112632},
  year={2021},
  doi={10.1016/j.rse.2021.112632},
  url={https://doi.org/10.1016/j.rse.2021.112632}
}

@article{Hawkins2023MycorrhizalMA,
  title={Mycorrhizal mycelium as a global carbon pool},
  author={Heidi-Jayne Hawkins and Rachael I. M. Cargill and Michael E. Van Nuland and Stephen C. Hagen and Katie J. Field and Merlin Sheldrake and Nadejda A. Soudzilovskaia and E. Toby Kiers},
  journal={Current Biology},
  year={2023},
  volume={33},
  number={11},
  pages={R560--R573},
  doi={10.1016/j.cub.2023.02.027},
  url={https://doi.org/10.1016/j.cub.2023.02.027}
}

@article{Breiman2001RF,
author = {Breiman, Leo},
title = {Random Forests},
year = {2001},
issue_date = {October 1 2001},
publisher = {Kluwer Academic Publishers},
address = {USA},
volume = {45},
number = {1},
issn = {0885-6125},
url = {https://doi.org/10.1023/A:1010933404324},
doi = {10.1023/A:1010933404324},
journal = {Mach. Learn.},
month = oct,
pages = {5–32},
numpages = {28}
}

@article{Friedman2001GreedyFA,
	title = {Greedy function approximation: {A} gradient boosting machine.},
	volume = {29},
	issn = {0090-5364},
	shorttitle = {Greedy function approximation},
	doi = {10.1214/aos/1013203451},
	number = {5},
	journal = {The Annals of Statistics},
	author = {Friedman, Jerome H.},
	month = oct,
	year = {2001},
	pages = {1189--1232},
}

@misc{bommasani2022opportunitiesrisksfoundationmodels,
      title={On the Opportunities and Risks of Foundation Models}, 
      author={Rishi Bommasani and Drew A. Hudson and Ehsan Adeli and Russ Altman and Simran Arora and Sydney von Arx and Michael S. Bernstein and Jeannette Bohg and Antoine Bosselut and Emma Brunskill and Erik Brynjolfsson and Shyamal Buch and Dallas Card and Rodrigo Castellon and Niladri Chatterji and Annie Chen and Kathleen Creel and Jared Quincy Davis and Dora Demszky and Chris Donahue and Moussa Doumbouya and Esin Durmus and Stefano Ermon and John Etchemendy and Kawin Ethayarajh and Li Fei-Fei and Chelsea Finn and Trevor Gale and Lauren Gillespie and Karan Goel and Noah Goodman and Shelby Grossman and Neel Guha and Tatsunori Hashimoto and Peter Henderson and John Hewitt and Daniel E. Ho and Jenny Hong and Kyle Hsu and Jing Huang and Thomas Icard and Saahil Jain and Dan Jurafsky and Pratyusha Kalluri and Siddharth Karamcheti and Geoff Keeling and Fereshte Khani and Omar Khattab and Pang Wei Koh and Mark Krass and Ranjay Krishna and Rohith Kuditipudi and Ananya Kumar and Faisal Ladhak and Mina Lee and Tony Lee and Jure Leskovec and Isabelle Levent and Xiang Lisa Li and Xuechen Li and Tengyu Ma and Ali Malik and Christopher D. Manning and Suvir Mirchandani and Eric Mitchell and Zanele Munyikwa and Suraj Nair and Avanika Narayan and Deepak Narayanan and Ben Newman and Allen Nie and Juan Carlos Niebles and Hamed Nilforoshan and Julian Nyarko and Giray Ogut and Laurel Orr and Isabel Papadimitriou and Joon Sung Park and Chris Piech and Eva Portelance and Christopher Potts and Aditi Raghunathan and Rob Reich and Hongyu Ren and Frieda Rong and Yusuf Roohani and Camilo Ruiz and Jack Ryan and Christopher Ré and Dorsa Sadigh and Shiori Sagawa and Keshav Santhanam and Andy Shih and Krishnan Srinivasan and Alex Tamkin and Rohan Taori and Armin W. Thomas and Florian Tramèr and Rose E. Wang and William Wang and Bohan Wu and Jiajun Wu and Yuhuai Wu and Sang Michael Xie and Michihiro Yasunaga and Jiaxuan You and Matei Zaharia and Michael Zhang and Tianyi Zhang and Xikun Zhang and Yuhui Zhang and Lucia Zheng and Kaitlyn Zhou and Percy Liang},
      year={2022},
      eprint={2108.07258},
      archivePrefix={arXiv},
      primaryClass={cs.LG},
      url={https://arxiv.org/abs/2108.07258}, 
}

@article{McInnes2018, doi = {10.21105/joss.00861}, url = {https://doi.org/10.21105/joss.00861}, year = {2018}, publisher = {The Open Journal}, volume = {3}, number = {29}, pages = {861}, author = {McInnes, Leland and Healy, John and Saul, Nathaniel and Großberger, Lukas}, title = {UMAP: Uniform Manifold Approximation and Projection}, journal = {Journal of Open Source Software} }

@ARTICLE{Wang2022SSL,
  author={Wang, Yi and Albrecht, Conrad M. and Braham, Nassim Ait Ali and Mou, Lichao and Zhu, Xiao Xiang},
  journal={IEEE Geoscience and Remote Sensing Magazine}, 
  title={Self-Supervised Learning in Remote Sensing: A review}, 
  year={2022},
  volume={10},
  number={4},
  pages={213-247},
  doi={10.1109/MGRS.2022.3198244}}

@article{Fick2017WorldClim2,
  author    = {Fick, S. E. and Hijmans, R. J.},
  title     = {{WorldClim} 2: new 1-km spatial resolution climate surfaces for global land areas},
  journal   = {International Journal of Climatology},
  year      = {2017},
  volume    = {37},
  pages     = {4302--4315},
  doi       = {10.1002/joc.5086},
  url       = {https://doi.org/10.1002/joc.5086}
}

@misc{Zanaga2022WorldCover,
  author    = {Zanaga, D. and Van De Kerchove, R. and Daems, D. and De Keersmaecker, W. and Brockmann, C. and Kirches, G. and Wevers, J. and Cartus, O. and Santoro, M. and Fritz, S. and Lesiv, M. and Herold, M. and Tsendbazar, N. E. and Xu, P. and Ramoino, F. and Arino, O.},
  title     = {{ESA WorldCover} 10 m 2021 v200},
  year      = {2022},
  doi       = {10.5281/zenodo.7254221},
  url       = {https://doi.org/10.5281/zenodo.7254221},
  note      = {Zenodo dataset}
}

@Article{soilgrids,
AUTHOR = {Poggio, L. and de Sousa, L. M. and Batjes, N. H. and Heuvelink, G. B. M. and Kempen, B. and Ribeiro, E. and Rossiter, D.},
TITLE = {SoilGrids 2.0: producing soil information for the globe with quantified spatial uncertainty},
JOURNAL = {SOIL},
VOLUME = {7},
YEAR = {2021},
NUMBER = {1},
PAGES = {217--240},
URL = {https://soil.copernicus.org/articles/7/217/2021/},
DOI = {10.5194/soil-7-217-2021}
}

@manual{RichDEM,
  title        = {RichDEM: Terrain Analysis Software},
  author       = {Richard Barnes},
  year         = {2016},
  url          = {http://github.com/r-barnes/richdem},
}

@article{Levin1992PatternScale,
  author    = {Levin, S. A.},
  title     = {The Problem of Pattern and Scale in Ecology: The Robert H. MacArthur Award Lecture},
  journal   = {Ecology},
  year      = {1992},
  volume    = {73},
  pages     = {1943--1967},
  doi       = {10.2307/1941447},
  url       = {https://doi.org/10.2307/1941447}
}

@article{Chave2013PatternScaleReview,
  author    = {Chave, J{\'e}r{\^o}me},
  title     = {The problem of pattern and scale in ecology: what have we learned in 20 years?},
  journal   = {Ecology Letters},
  year      = {2013},
  volume    = {16},
  number    = {1},
  pages     = {4--16},
  month     = {May},
  doi       = {10.1111/ele.12048},
  url       = {https://doi.org/10.1111/ele.12048}
}

@article{Hortal2015SevenShortfalls,
  author    = {Joaqu{\'i}n Hortal and Francesco de Bello and Jos{\'e} Alexandre F. Diniz-Filho and Thomas M. Lewinsohn and Jorge M. Lobo and Richard J. Ladle},
  title     = {Seven Shortfalls that Beset Large-Scale Knowledge of Biodiversity},
  journal   = {Annual Review of Ecology, Evolution, and Systematics},
  year      = {2015},
  volume    = {46},
  pages     = {523--549},
  doi       = {10.1146/annurev-ecolsys-112414-054400},
  url       = {https://doi.org/10.1146/annurev-ecolsys-112414-054400},
  note      = {First published online October 28, 2015}
}

@book{Smith2010MycorrhizalSymbiosis,
  author    = {Sally E. Smith and David J. Read},
  title     = {Mycorrhizal Symbiosis},
  edition   = {3},
  publisher = {Academic Press},
  year      = {2010},
  isbn      = {9780080559346},
  pages     = {800},
  url       = {https://www.elsevier.com/books/mycorrhizal-symbiosis/smith/978-0-08-055934-6}
}

@article{Brundrett2009MycorrhizalAssociations,
  author    = {Brundrett, M. C.},
  title     = {Mycorrhizal associations and other means of nutrition of vascular plants: understanding the global diversity of host plants by resolving conflicting information and developing reliable means of diagnosis},
  journal   = {Plant and Soil},
  year      = {2009},
  volume    = {320},
  pages     = {37--77},
  doi       = {10.1007/s11104-008-9877-9},
  url       = {https://doi.org/10.1007/s11104-008-9877-9}
}

@misc{Ke2017LightGBM,
author = {Ke, Guolin and Meng, Qi and Finley, Thomas and Wang, Taifeng and Chen, Wei and Ma, Weidong and Ye, Qiwei and Liu, Tie-Yan},
title = {LightGBM: a highly efficient gradient boosting decision tree},
year = {2017},
isbn = {9781510860964},
publisher = {Curran Associates Inc.},
address = {Red Hook, NY, USA},
booktitle = {Proceedings of the 31st International Conference on Neural Information Processing Systems},
pages = {3149–3157},
numpages = {9},
location = {Long Beach, California, USA},
series = {NIPS'17}
}

@misc{kim2018attentive,
title={Attentive Neural Processes},
author={Hyunjik Kim and Andriy Mnih and Jonathan Schwarz and Marta Garnelo and Ali Eslami and Dan Rosenbaum and Oriol Vinyals and Yee Whye Teh},
booktitle={International Conference on Learning Representations},
year={2019},
url={https://openreview.net/forum?id=SkE6PjC9KX},
}

@InProceedings{Garnelo2018CNP,
  title = 	 {Conditional Neural Processes},
  author =       {Garnelo, Marta and Rosenbaum, Dan and Maddison, Christopher and Ramalho, Tiago and Saxton, David and Shanahan, Murray and Teh, Yee Whye and Rezende, Danilo and Eslami, S. M. Ali},
  booktitle = 	 {Proceedings of the 35th International Conference on Machine Learning},
  pages = 	 {1704--1713},
  year = 	 {2018},
  editor = 	 {Dy, Jennifer and Krause, Andreas},
  volume = 	 {80},
  series = 	 {Proceedings of Machine Learning Research},
  month = 	 {10--15 Jul},
  publisher =    {PMLR},
  url = 	 {https://proceedings.mlr.press/v80/garnelo18a.html},
}

@article{vanderHeijden2008,
  title={The unseen majority: soil microbes as drivers of plant diversity and productivity in terrestrial ecosystems},
  author={van der Heijden, Marcel G.A. and Bardgett, Richard D. and van Straalen, Nico M.},
  journal={Ecology Letters},
  volume={11},
  number={3},
  pages={296--310},
  year={2008},
  publisher={Wiley},
  doi={10.1111/j.1461-0248.2007.01139.x},
  url={https://doi.org/10.1111/j.1461-0248.2007.01139.x}
}

@article{Hodge2001,
  title={Arbuscular mycorrhizal fungi influence decomposition of, but not plant nutrient capture from, glycine patches in soil},
  author={Hodge, Angela},
  journal={New Phytologist},
  volume={151},
  number={3},
  pages={725--734},
  year={2001},
  publisher={Wiley},
  doi={10.1046/j.0028-646x.2001.00200.x},
  url={https://doi.org/10.1046/j.0028-646x.2001.00200.x}
}

@article{Rousk2010,
  title={Soil bacterial and fungal communities across a pH gradient in an arable soil},
  author={Rousk, Johannes and Baath, Erland and Brookes, Philip C and Lauber, Christian L and Lozupone, Catherine and Caporaso, J Gregory and Knight, Rob and Fierer, Noah},
  journal={The ISME Journal},
  volume={4},
  number={10},
  pages={1340--1351},
  year={2010},
  publisher={Nature Publishing Group},
  doi={10.1038/ismej.2010.58},
  url={https://doi.org/10.1038/ismej.2010.58}
}

@article{Schweiger2018PlantSpectralDiversity,
  author    = {Anna K. Schweiger and Jeannine Cavender-Bares and Philip A. Townsend and Sarah E. Hobbie and Michael D. Madritch and Ran Wang and David Tilman and John A. Gamon},
  title     = {Plant spectral diversity integrates functional and phylogenetic components of biodiversity and predicts ecosystem function},
  journal   = {Nature Ecology \& Evolution},
  year      = {2018},
  volume    = {2},
  number    = {6},
  pages     = {976--982},
  doi       = {10.1038/s41559-018-0551-1},
  month     = {June},
}

@article{Wang2018OpticalDiversity,
  author    = {Ran Wang and John A. Gamon and Anna K. Schweiger and Jeannine Cavender-Bares and Philip A. Townsend and Arthur I. Zygielbaum and Shan Kothari},
  title     = {Influence of species richness, evenness, and composition on optical diversity: A simulation study},
  journal   = {Remote Sensing of Environment},
  year      = {2018},
  volume    = {211},
  pages     = {218--228},
  month     = {June},
  doi       = {https://doi.org/10.1016/j.rse.2018.04.010},
}

@article{Barcelo2019,
  author = {Barceló, M. and van Bodegom, P. M. and Soudzilovskaia, N. A.},
  title = {Climate drives the spatial distribution of mycorrhizal host plants in terrestrial ecosystems},
  journal = {Journal of Ecology},
  year = {2019},
  volume = {107},
  number = {6},
  pages = {2564--2573},
  doi = {10.1111/1365-2745.13275},
  url = {https://doi.org/10.1111/1365-2745.13275}
}

@article{Dai2013,
  author = {Dai, M. and Bainard, L. D. and Hamel, C. and Gan, Y. and Lynch, D.},
  title = {Impact of Land Use on Arbuscular Mycorrhizal Fungal Communities in Rural Canada},
  journal = {Applied and Environmental Microbiology},
  year = {2013},
  volume = {79},
  doi = {10.1128/AEM.01333-13},
  url = {https://doi.org/10.1128/AEM.01333-13}
}

@article{Hyndman2006,
  author = {Hyndman, Rob J. and Koehler, Anne B.},
  title = {Another Look at Measures of Forecast Accuracy},
  journal = {International Journal of Forecasting},
  year = {2006},
  volume = {22},
  number = {4},
  pages = {679--688},
  doi = {10.1016/j.ijforecast.2006.03.001},
  url = {https://doi.org/10.1016/j.ijforecast.2006.03.001}
}

@article{Dinerstein2017,
  author = {Eric Dinerstein and David Olson and Anup Joshi and Carly Vynne and Neil D. Burgess and Eric Wikramanayake and Nathan Hahn and Suzanne Palminteri and Prashant Hedao and Reed Noss and Matt Hansen and Harvey Locke and Erle C. Ellis and Benjamin Jones and Charles Victor Barber and Randy Hayes and Cyril Kormos and Vance Martin and Eileen Crist and Wes Sechrest and Lori Price and Jonathan E. M. Baillie and Don Weeden and Kierán Suckling and Crystal Davis and Nigel Sizer and Rebecca Moore and David Thau and Tanya Birch and Peter Potapov and Svetlana Turubanova and Alexandra Tyukavina and Nadia de Souza and Lilian Pintea and José C. Brito and Othman A. Llewellyn and Anthony G. Miller and Annette Patzelt and Shahina A. Ghazanfar and Jonathan Timberlake and Heinz Klöser and Yara Shennan-Farpón and Roeland Kindt and Jens-Peter Barnekow Lillesø and Paulo van Breugel and Lars Graudal and Maianna Voge and Khalaf F. Al-Shammari and Muhammad Saleem},
  title = {An Ecoregion-Based Approach to Protecting Half the Terrestrial Realm},
  journal = {BioScience},
  volume = {67},
  number = {6},
  pages = {534--545},
  year = {2017},
  month = {June},
  doi = {10.1093/biosci/bix014},
  url = {https://doi.org/10.1093/biosci/bix014}
}

@article{ElithLeathwick2009,
  title = {Species Distribution Models: Ecological Explanation and Prediction Across Space and Time},
  author = {Jane Elith and John R. Leathwick},
  journal = {Annual Review of Ecology, Evolution, and Systematics},
  volume = {40},
  pages = {677--697},
  year = {2009},
  month = dec,
  doi = {10.1146/annurev.ecolsys.110308.120159},
  url = {https://doi.org/10.1146/annurev.ecolsys.110308.120159}
}

@article{Vetrovsky2020GlobalFungi,
  author    = {Tom{\'a}{\v s} V{\v e}trovsk{\'y} and Daniel Morais and Petr Kohout and Cl{\'e}mentine Lepinay and Camelia Algora and Sandra Awokunle Holl{\'a} and Barbara Doreen Bahnmann and Kv{\v e}ta B{\'i}lohn{\v e}d{\'a} and Vendula Brabcov{\'a} and Federica D’Al{\`o} and Zander Rainier Human and Mayuko Jomura and Miroslav Kola{\v r}{\'i}k and Jana Kvasni{\v c}kov{\'a} and Salvador Llad{\'o} and Rub{\'e}n L{\'o}pez-Mond{\'e}jar and Tijana Martinovi{\'c} and Tereza Ma{\v s}{\'i}nov{\'a} and Lenka Mesz{\'a}ro{\v s}ov{\'a} and Lenka Michal{\v c}ikov{\'a} and Tereza Michalov{\'a} and Sunil Mundra and Diana Navr{\'a}tilov{\'a} and I{\~n}aki Odriozola and Sarah Pich{\'e}-Choquette and Martina {\v S}tursov{\'a} and Karel {\v S}vec and Vojt{\v e}ch Tl{\'a}skal and Michaela Urbanov{\'a} and Luk{\'a}{\v s} Vlk and Jana Vo{\v r}{\'i}{\v s}kov{\'a} and Lucia {\v Z}if{\v c}{\'a}kov{\'a} and Petr Baldrian},
  title     = {GlobalFungi, a global database of fungal occurrences from high-throughput-sequencing metabarcoding studies},
  journal   = {Scientific Data},
  year      = {2020},
  volume    = {7},
  number    = {1},
  pages     = {228},
  doi       = {10.1038/s41597-020-0567-7},
  url       = {https://doi.org/10.1038/s41597-020-0567-7}
}

@article{Polme2020FungalTraits,
  author    = {Sergei P{\~o}lme and Kessy Abarenkov and R. Henrik Nilsson and Bj{\"o}rn D. Lindahl and Karina Engelbrecht Clemmensen and Havard Kauserud and Nhu Nguyen and Rasmus Kj{\o}ller and Scott T. Bates and Petr Baldrian and Tobias Guldberg Fr{\o}slev and Kristjan Adojaan and Alfredo Vizzini and Ave Suija and Donald Pfister and Hans-Otto Baral and Helle J{\"a}rv and Hugo Madrid and Jenni Nord{\'e}n and Jian-Kui Liu and Julia Pawlowska and Kadri P{\~o}ldmaa and Kadri P{\"a}rtel and Kadri Runnel and Karen Hansen and Karl-Henrik Larsson and Kevin David Hyde and Marcelo Sandoval-Denis and Matthew E. Smith and Merje Toome-Heller and Nalin N. Wijayawardene and Nelson Menolli Jr. and Nicole K. Reynolds and Rein Drenkhan and Sajeewa S. N. Maharachchikumbura and Tatiana B. Gibertoni and Thomas L{\ae}ss{\o}e and William Davis and Yuri Tokarev and Adriana Corrales and Adriene Mayra Soares and Ahto Agan and Alexandre Reis Machado and Andr{\'e}s Arg{\"u}elles-Moyao and Andrew Detheridge and Angelina de Meiras-Ottoni and Annemieke Verbeken and Arun Kumar Dutta and Bao-Kai Cui and C. K. Pradeep and C{\'e}sar Mar{\'i}n and Daniel Stanton and Daniyal Gohar and Dhanushka N. Wanasinghe and Eveli Otsing and Farzad Aslani and Gareth W. Griffith and Thorsten H. Lumbsch and Hans-Peter Grossart and Hossein Masigol and Ina Timling and Inga Hiiesalu and Jane Oja and John Y. Kupagme and J{\'o}zsef Geml and Julieta Alvarez-Manjarrez and Kai Ilves and Kaire Loit and Kalev Adamson and Kazuhide Nara and Kati K{\"u}ngas and Keilor Rojas-Jimenez and Kri{\v s}s Bitenieks and Laszlo Irinyi and L{\'a}szl{\'o} G. Nagy and Liina Soonvald and Li-Wei Zhou and Lysett Wagner and M. Catherine Aime and Maarja {\"O}pik and Mar{\'i}a Isabel Mujica and Martin Metsoja and Martin Ryberg and Martti Vasar and Masao Murata and Matthew P. Nelsen and Michelle Cleary and Milan C. Samarakoon and Mingkwan Doilom and Mohammad Bahram and Niloufar Hagh-Doust and Olesya Dulya and Peter Johnston and Petr Kohout and Qian Chen and Qing Tian and Rajasree Nandi and Rasekh Amiri and Rekhani Hansika Perera and Renata dos Santos Chikowski and Renato L. Mendes-Alvarenga and Roberto Garibay-Orijel and Robin Gielen and Rungtiwa Phookamsak and Ruvishika S. Jayawardena and Saleh Rahimlou and Samantha C. Karunarathna and Saowaluck Tibpromma and Shawn P. Brown and Siim-Kaarel Sepp and Sunil Mundra and Zhu-Hua Luo and Tanay Bose and Tanel Vahter and Tarquin Netherway and Teng Yang and Tom May and Torda Varga and Wei Li and Victor Rafael Matos Coimbra and Virton Rodrigo Targino de Oliveira and Vitor Xavier de Lima and Vladimir S. Mikryukov and Yongzhong Lu and Yosuke Matsuda and Yumiko Miyamoto and Urmas K{\~o}ljalg and Leho Tedersoo},
  title     = {FungalTraits: a user-friendly traits database of fungi and fungus-like stramenopiles},
  journal   = {Fungal Diversity},
  year      = {2020},
  volume    = {105},
  pages     = {1--16},
  doi       = {10.1007/s13225-020-00466-2},
  url       = {https://doi.org/10.1007/s13225-020-00466-2}
}

@article{vanderHeijden2015mycorrhizal,
  title        = {Mycorrhizal ecology and evolution: the past, the present, and the future},
  author       = {van der Heijden, Marcel G. A. and Martin, Francis M. and Selosse, Marc-Andr{\'e} and Sanders, Ian R.},
  journal      = {New Phytologist},
  volume       = {205},
  number       = {4},
  pages        = {1406--1423},
  year         = {2015},
  month        = mar,
  doi          = {10.1111/nph.13288},
  issn         = {1469-8137},
  pmid         = {25639293},
  url          = {https://doi.org/10.1111/nph.13288},
}

@article{Brundrett2018evolutionary,
  title        = {Evolutionary history of mycorrhizal symbioses and global host plant diversity},
  author       = {Brundrett, Mark C. and Tedersoo, Leho},
  journal      = {New Phytologist},
  year         = {2018},
  doi          = {10.1111/nph.14976},
  url          = {https://doi.org/10.1111/nph.14976},
}

@article{Delavaux2021mycorrhizal,
  title        = {Mycorrhizal types influence island biogeography of plants},
  author       = {Delavaux, Camille S. and Weigelt, Patrick and Dawson, Wayne and Essl, Franz and van Kleunen, Mark and K{\"o}nig, Christian and Pergl, Jan and Py{\v{s}}ek, Petr and Stein, Anke and Winter, Marten and Taylor, Amanda and Schultz, Peggy A. and Whittaker, Robert J. and Kreft, Holger and Bever, James D.},
  journal      = {Communications Biology},
  year         = {2021},
  volume       = {4},
  pages        = {1128},
  doi          = {10.1038/s42003-021-02649-2},
  url          = {https://doi.org/10.1038/s42003-021-02649-2},
}

@article{Averill2019global,
  title        = {Global imprint of mycorrhizal fungi on whole-plant nutrient economics},
  author       = {Averill, Colin and Bhatnagar, Jennifer M. and Dietze, Michael C. and Kivlin, Stephanie N.},
  journal      = {Proceedings of the National Academy of Sciences},
  year         = {2019},
  volume       = {116},
  number       = {46},
  pages        = {23163--23168},
  doi          = {10.1073/pnas.1906655116},
  url          = {https://doi.org/10.1073/pnas.1906655116},
}

@article{Basset2025AI,
  title        = {The role of AI-enhanced microscopy in soil biodiversity assessment: Advancing soil security, connectivity and governance with implications for the European Directive on Soil Monitoring and Resilience, and global agendas},
  author       = {Basset, Celine and Zaldo-Aubanell, Quim},
  journal      = {Soil Security},
  year         = {2025},
  pages        = {100203},
  doi          = {10.1016/j.soisec.2025.100203},
  url          = {https://doi.org/10.1016/j.soisec.2025.100203},
}

@article{Guerra2022global,
  title        = {Global hotspots for soil nature conservation},
  author       = {Guerra, Carlos A. and Berdugo, Miguel and Eldridge, David J. and Eisenhauer, Nico and Singh, Brajesh K. and Cui, Haiying and Abades, Sebastian and Alfaro, Fernando D. and Bamigboye, Adebola R. and Bastida, Felipe and Blanco-Pastor, José L. and de los Ríos, Asunción and Durán, Jorge and Grebenc, Tine and Illán, Javier G. and Liu, Yu-Rong and Makhalanyane, Thulani P. and Mamet, Steven and Molina-Montenegro, Marco A. and Moreno, José L. and Mukherjee, Arpan and Nahberger, Tina U. and Peñaloza-Bojacá, Gabriel F. and Plaza, César and Picó, Sergio and Verma, Jay Prakash and Rey, Ana and Rodríguez, Alexandra and Tedersoo, Leho and Teixido, Alberto L. and Torres-Díaz, Cristian and Trivedi, Pankaj and Wang, Juntao and Wang, Ling and Wang, Jianyong and Zaady, Eli and Zhou, Xiaobing and Zhou, Xin-Quan and Delgado-Baquerizo, Manuel},
  journal      = {Nature},
  year         = {2022},
  volume       = {610},
  pages        = {693--698},
  doi          = {10.1038/s41586-022-05292-x},
  url          = {https://doi.org/10.1038/s41586-022-05292-x}
}

@article{Tedersoo2022,
  title        = {Best practices in metabarcoding of fungi: From experimental design to results},
  author       = {Tedersoo, Leho and Bahram, Mohammad and Zinger, Lucie and Nilsson, R Henrik and Kennedy, Peter G and Yang, Teng and Anslan, Sten and Mikryukov, Vladimir},
  journal      = {Molecular Ecology},
  year         = {2022},
  volume       = {31},
  number       = {10},
  pages        = {2769--2795},
  doi          = {10.1111/mec.16460},
  pmid         = {35395127},
  month        = may,
  note         = {Epub 2022 Apr 20}
}

@article{Nilsson2019,
  title        = {Mycobiome diversity: high-throughput sequencing and identification of fungi},
  author       = {Nilsson, R Henrik and Anslan, Sten and Bahram, Mohammad and Wurzbacher, Christian and Baldrian, Petr and Tedersoo, Leho},
  journal      = {Nature Reviews Microbiology},
  year         = {2019},
  volume       = {17},
  number       = {2},
  pages        = {95--109},
  doi          = {10.1038/s41579-018-0116-y},
  pmid         = {30442909},
  month        = jan
}

@article{Mikryukov2023,
  title        = {Connecting the multiple dimensions of global soil fungal diversity},
  author       = {Mikryukov, Vladimir and Dulya, Olesya and Zizka, Alexander and Bahram, Mohammad and Hagh-Doust, Niloufar and Anslan, Sten and Prylutskyi, Oleh and Delgado-Baquerizo, Manuel and Maestre, Fernando T and Nilsson, Henrik and Pärn, Jaan and Öpik, Maarja and Moora, Mari and Zobel, Martin and Espenberg, Mikk and Mander, Ülo and Khalid, Abdul Nasir and Corrales, Adriana and Agan, Ahto and Vasco-Palacios, Aída-M and Saitta, Alessandro and Rinaldi, Andrea and Verbeken, Annemieke and Sulistyo, Bobby and Tamgnoue, Boris and Furneaux, Brendan and Ritter, Camila Duarte and Nyamukondiwa, Casper and Sharp, Cathy and Marín, César and Gohar, Daniyal and Klavina, Darta and Sharmah, Dipon and Dai, Dong-Qin and Nouhra, Eduardo and Biersma, Elisabeth Machteld and Rähn, Elisabeth and Cameron, Erin and De Crop, Eske and Otsing, Eveli and Davydov, Evgeny and Albornoz, Felipe and Brearley, Francis and Buegger, Franz and Zahn, Geoffrey and Bonito, Gregory and Hiiesalu, Inga and Barrio, Isabel and Heilmann-Clausen, Jacob and Ankuda, Jelena and Doležal, Jiri and Kupagme, John and Maciá-Vicente, Jose and Djeugap Fovo, Joseph and Geml, József and Alatalo, Juha and Alvarez-Manjarrez, Julieta and Põldmaa, Kadri and Runnel, Kadri and Adamson, Kalev and Bråthen, Kari-Anne and Pritsch, Karin and Issifou, Kassim Tchan and Armolaitis, Kęstutis and Hyde, Kevin and Newsham, Kevin K and Panksep, Kristel and Lateef, Adebola Azeez and Hansson, Linda and Lamit, Louis and Saba, Malka and Tuomi, Maria and Gryzenhout, Marieka and Bauters, Marijn and Piepenbring, Meike and Wijayawardene, Nalin N and Yorou, Nourou and Kurina, Olavi and Mortimer, Peter and Meidl, Peter and Kohout, Petr and Puusepp, Rasmus and Drenkhan, Rein and Garibay-Orijel, Roberto and Godoy, Roberto and Alkahtani, Saad and Rahimlou, Saleh and Dudov, Sergey and Põlme, Sergei and Ghosh, Soumya and Mundra, Sunil and Ahmed, Talaat and Netherway, Tarquin and Henkel, Terry and Roslin, Tomas and Nteziryayo, Vincent and Fedosov, Vladimir and Onipchenko, Vladimir and Yasanthika, Weeragalle Arachchillage Erandi and Lim, Young and Van Nuland, Michael and Soudzilovskaia, Nadejda and Antonelli, Alexandre and Kõljalg, Urmas and Abarenkov, Kessy and Tedersoo, Leho},
  journal      = {Science Advances},
  year         = {2023},
  volume       = {9},
  number       = {48},
  pages        = {eadj8016},
  doi          = {10.1126/sciadv.adj8016},
  pmid         = {38019923},
  pmcid        = {PMC10686567},
  month        = dec,
  note         = {Epub 2023 Nov 29}
}

@article{Steidinger2019,
  title        = {Climatic controls of decomposition drive the global biogeography of forest-tree symbioses},
  author       = {Steidinger, B. S. and Crowther, T. W. and Liang, J. and Van Nuland, M. E. and Werner, G. D. A. and Reich, P. B. and Nabuurs, G. J. and de-Miguel, S. and Zhou, M. and Picard, N. and Herault, B. and Zhao, X. and Zhang, C. and Routh, D. and Peay, K. G. and the GFBI consortium},
  journal      = {Nature},
  year         = {2019},
  volume       = {569},
  number       = {7756},
  pages        = {404--408},
  doi          = {10.1038/s41586-019-1128-0}
}

@article{Steidinger2020,
  title        = {Ectomycorrhizal fungal diversity predicted to substantially decline due to climate changes in North American Pinaceae forests},
  author       = {Steidinger, B. S. and Bhatnagar, J. M. and Vilgalys, R. and Taylor, J. W. and Qin, C. and Zhu, K. and Bruns, T. D. and Peay, K. G.},
  journal      = {Journal of Biogeography},
  year         = {2020},
  volume       = {47},
  number       = {3},
  pages        = {772--782},
  doi          = {10.1111/jbi.13802}
}

@article{VanNuland2024,
  title        = {Climate mismatches with ectomycorrhizal fungi contribute to migration lag in North American tree range shifts},
  author       = {Van Nuland, M. E. and Qin, C. and Pellitier, P. T. and Zhu, K. and Peay, K. G.},
  journal      = {Proceedings of the National Academy of Sciences},
  year         = {2024},
  volume       = {121},
  number       = {23},
  pages        = {e2308811121},
  doi          = {10.1073/pnas.2308811121}
}

@article{Labouyrie2023,
  title        = {Patterns in soil microbial diversity across Europe},
  author       = {Labouyrie, Maëva and Ballabio, Cristiano and Romero, Ferran and Panagos, Panos and Jones, Arwyn and Schmid, Marc W. and Mikryukov, Vladimir and Dulya, Olesya and Tedersoo, Leho and Bahram, Mohammad and Lugato, Emanuele and van der Heijden, Marcel G. A. and Orgiazzi, Alberto},
  journal      = {Nature Communications},
  year         = {2023},
  volume       = {14},
  number       = {1},
  pages        = {3311},
  doi          = {10.1038/s41467-023-37937-4},
  pmid         = {37291086},
  pmcid        = {PMC10250377},
  month        = jun
}

@article{Sousa2021,
  title        = {Tree Canopies Reflect Mycorrhizal Composition},
  author       = {Sousa, Daniel and Fisher, Joshua B. and Romero Galvan, Fernando and Pavlick, Ryan P. and Cordell, Susan and Giambelluca, Thomas W. and Giardina, Christian P. and Gilbert, Gregory S. and Imran-Narahari, Faith and Litton, Creighton M. and Lutz, James A. and North, Malcolm P. and Orwig, David A. and Ostertag, Rebecca and Sack, Lawren and Phillips, Richard P.},
  journal      = {Geophysical Research Letters},
  year         = {2021},
  volume       = {48},
  number       = {10},
  pages        = {e2021GL092764},
  doi          = {10.1029/2021GL092764},
  month        = may,
  note         = {First published: 19 May 2021}
}

@article{Wang2020,
  title        = {Foliar functional traits from imaging spectroscopy across biomes in eastern North America},
  author       = {Wang, Zhihui and Chlus, Adam and Geygan, Ryan and Ye, Zhiwei and Zheng, Ting and Singh, Aditya and Couture, John J. and Cavender-Bares, Jeannine and Kruger, Eric L. and Townsend, Philip A.},
  journal      = {New Phytologist},
  year         = {2020},
  volume       = {228},
  number       = {2},
  pages        = {494--511},
  doi          = {10.1111/nph.16711},
  note         = {Epub 2020 Jun 23}  
}

@article{Schweiger2022,
  title        = {Plant beta-diversity across biomes captured by imaging spectroscopy},
  author       = {Schweiger, Anna K. and Laliberté, Etienne},
  journal      = {Nature Communications},
  year         = {2022},
  volume       = {13},
  number       = {1},
  pages        = {2767},
  doi          = {10.1038/s41467-022-30369-6}
}

@article{Tedersoo2020,
  title        = {How mycorrhizal associations drive plant population and community biology},
  author       = {Tedersoo, Leho and Bahram, Mohammad and Zobel, Martin},
  journal      = {Science},
  year         = {2020},
  volume       = {367},
  number       = {6480},
  pages        = {eaba1223},
  doi          = {10.1126/science.aba1223}
}

@article{Sorensen2025ExploringCH,
	title = {Exploring crop health and its associations with fungal soil microbiome composition using machine learning applied to remote sensing data},
	volume = {6},
	issn = {2662-4435},
	url = {https://www.nature.com/articles/s43247-025-02330-0},
	doi = {10.1038/s43247-025-02330-0},
	language = {en},
	number = {1},
	journal = {Communications Earth \& Environment},
	author = {Sørensen, Mathies Brinks and Faurdal, David and Schiesaro, Giovanni and Jensen, Emil Damgaard and Jensen, Michael Krogh and Clemmensen, Line Katrine Harder},
	month = may,
	year = {2025},
	pages = {355},
}

@article{Zhang2025Mycelium,
	title = {Mycelium biomass and community composition impact nutrient concentration in arbuscular mycorrhizal fungi at fine spatial scale},
	volume = {39},
	issn = {0269-8463, 1365-2435},
	url = {https://besjournals.onlinelibrary.wiley.com/doi/10.1111/1365-2435.70042},
	doi = {10.1111/1365-2435.70042},
	language = {en},
	number = {6},
	journal = {Functional Ecology},
	author = {Zhang, Xiaojing and Bo, Yushan and Jiang, Liangchao and Wang, Jing and Yu, Lingfei and Fu, Wei and He, Xueli and Dong, Xiaoning and Jiang, Yong and Han, Xingguo and Zhang, Haiyang},
	month = jun,
	year = {2025},
	pages = {1455--1468},
}

@article{Tedersoo2021,
	title = {The {Global} {Soil} {Mycobiome} consortium dataset for boosting fungal diversity research},
	volume = {111},
	issn = {1560-2745, 1878-9129},
	url = {https://link.springer.com/10.1007/s13225-021-00493-7},
	doi = {10.1007/s13225-021-00493-7},
	language = {en},
	number = {1},
	journal = {Fungal Diversity},
	author = {Tedersoo, Leho and Mikryukov, Vladimir and Anslan, Sten and Bahram, Mohammad and Khalid, Abdul Nasir and Corrales, Adriana and Agan, Ahto and Vasco-Palacios, Aída-M. and Saitta, Alessandro and Antonelli, Alexandre and Rinaldi, Andrea C. and Verbeken, Annemieke and Sulistyo, Bobby P. and Tamgnoue, Boris and Furneaux, Brendan and Ritter, Camila Duarte and Nyamukondiwa, Casper and Sharp, Cathy and Marín, César and Dai, D. Q. and Gohar, Daniyal and Sharmah, Dipon and Biersma, Elisabeth Machteld and Cameron, Erin K. and De Crop, Eske and Otsing, Eveli and Davydov, Evgeny A. and Albornoz, Felipe E. and Brearley, Francis Q. and Buegger, Franz and Gates, Genevieve and Zahn, Geoffrey and Bonito, Gregory and Hiiesalu, Indrek and Hiiesalu, Inga and Zettur, Irma and Barrio, Isabel C. and Pärn, Jaan and Heilmann-Clausen, Jacob and Ankuda, Jelena and Kupagme, John Y. and Sarapuu, Joosep and Maciá-Vicente, Jose G. and Fovo, Joseph Djeugap and Geml, József and Alatalo, Juha M. and Alvarez-Manjarrez, Julieta and Monkai, Jutamart and Põldmaa, Kadri and Runnel, Kadri and Adamson, Kalev and Bråthen, Kari A. and Pritsch, Karin and Tchan, Kassim I. and Armolaitis, Kęstutis and Hyde, Kevin D. and Newsham, Kevin K. and Panksep, Kristel and Adebola, Lateef A. and Lamit, Louis J. and Saba, Malka and Da Silva Cáceres, Marcela E. and Tuomi, Maria and Gryzenhout, Marieka and Bauters, Marijn and Bálint, Miklós and Wijayawardene, Nalin and Hagh-Doust, Niloufar and Yorou, Nourou S. and Kurina, Olavi and Mortimer, Peter E. and Meidl, Peter and Nilsson, R. Henrik and Puusepp, Rasmus and Casique-Valdés, Rebeca and Drenkhan, Rein and Garibay-Orijel, Roberto and Godoy, Roberto and Alfarraj, Saleh and Rahimlou, Saleh and Põlme, Sergei and Dudov, Sergey V. and Mundra, Sunil and Ahmed, Talaat and Netherway, Tarquin and Henkel, Terry W. and Roslin, Tomas and Fedosov, Vladimir E. and Onipchenko, Vladimir G. and Yasanthika, W. A. Erandi and Lim, Young Woon and Piepenbring, Meike and Klavina, Darta and Kõljalg, Urmas and Abarenkov, Kessy},
	month = nov,
	year = {2021},
	pages = {573--588},
}

@article{Toussaint2020,
	title = {Asymmetric patterns of global diversity among plants and mycorrhizal fungi},
	volume = {31},
	issn = {1100-9233, 1654-1103},
	url = {https://onlinelibrary.wiley.com/doi/10.1111/jvs.12837},
	doi = {10.1111/jvs.12837},
	language = {en},
	number = {2},
	journal = {Journal of Vegetation Science},
	author = {Toussaint, Aurele and Bueno, Guillermo and Davison, John and Moora, Mari and Tedersoo, Leho and Zobel, Martin and Öpik, Maarja and Pärtel, Meelis},
	editor = {Pugnaire, Francisco},
	month = mar,
	year = {2020},
	pages = {355--366},
}

@article{Tedersoo2016,
	title = {Fungal identification biases in microbiome projects},
	volume = {8},
	issn = {1758-2229},
	url = {https://sfamjournals.onlinelibrary.wiley.com/doi/10.1111/1758-2229.12438},
	doi = {10.1111/1758-2229.12438},
	language = {en},
	number = {5},
	journal = {Environmental Microbiology Reports},
	author = {Tedersoo, Leho and Lindahl, Björn},
	month = oct,
	year = {2016},
	pages = {774--779},
}

@article{Yang2018,
	title = {Evaluation of the ribosomal {DNA} internal transcribed spacer ({ITS}), specifically {ITS1} and {ITS2}, for the analysis of fungal diversity by deep sequencing},
	volume = {13},
	issn = {1932-6203},
	url = {https://dx.plos.org/10.1371/journal.pone.0206428},
	doi = {10.1371/journal.pone.0206428},
	language = {en},
	number = {10},
	journal = {PLOS ONE},
	author = {Yang, Rui-Heng and Su, Jin-He and Shang, Jun-Jun and Wu, Ying-Ying and Li, Yan and Bao, Da-Peng and Yao, Yi-Jian},
	editor = {Cullen, Daniel},
	month = oct,
	year = {2018},
	pages = {e0206428},
}

@article{BengtssonPalme2013,
	title = {Improved software detection and extraction of {ITS1} and {ITS2} from ribosomal {ITS} sequences of fungi and other eukaryotes for analysis of environmental sequencing data},
	volume = {4},
	copyright = {http://onlinelibrary.wiley.com/termsAndConditions\#vor},
	issn = {2041-210X},
	url = {https://besjournals.onlinelibrary.wiley.com/doi/10.1111/2041-210X.12073},
	doi = {10.1111/2041-210X.12073},
	language = {en},
	number = {10},
	journal = {Methods in Ecology and Evolution},
	author = {Bengtsson‐Palme, Johan and Ryberg, Martin and Hartmann, Martin and Branco, Sara and Wang, Zheng and Godhe, Anna and De Wit, Pierre and Sánchez‐García, Marisol and Ebersberger, Ingo and De Sousa, Filipe and Amend, Anthony and Jumpponen, Ari and Unterseher, Martin and Kristiansson, Erik and Abarenkov, Kessy and Bertrand, Yann J. K. and Sanli, Kemal and Eriksson, K. Martin and Vik, Unni and Veldre, Vilmar and Nilsson, R. Henrik},
	editor = {Bunce, Michael},
	month = oct,
	year = {2013},
	pages = {914--919},
}

@article{Edgar2013,
	title = {{UPARSE}: highly accurate {OTU} sequences from microbial amplicon reads},
	volume = {10},
	copyright = {http://www.springer.com/tdm},
	issn = {1548-7091, 1548-7105},
	shorttitle = {{UPARSE}},
	url = {https://www.nature.com/articles/nmeth.2604},
	doi = {10.1038/nmeth.2604},
	language = {en},
	number = {10},
	journal = {Nature Methods},
	author = {Edgar, Robert C},
	month = oct,
	year = {2013},
	pages = {996--998},
}

@article{Chao2014,
	title = {Rarefaction and extrapolation with {Hill} numbers: a framework for sampling and estimation in species diversity studies},
	volume = {84},
	issn = {0012-9615, 1557-7015},
	doi = {10.1890/13-0133.1},
	language = {en},
	number = {1},
	journal = {Ecological Monographs},
	author = {Chao, Anne and Gotelli, Nicholas J. and Hsieh, T. C. and Sander, Elizabeth L. and Ma, K. H. and Colwell, Robert K. and Ellison, Aaron M.},
	month = feb,
	year = {2014},
	pages = {45--67},
}

@article{Hsieh2016,
	title = {{iNEXT}: an {R} package for rarefaction and extrapolation of species diversity ({Hill} numbers)},
	volume = {7},
	issn = {2041-210X},
	doi = {10.1111/2041-210X.12613},
	language = {en},
	number = {12},
	journal = {Methods in Ecology and Evolution},
	author = {Hsieh, T. C. and Ma, K. H. and Chao, Anne},
	editor = {McInerny, Greg},
	month = dec,
	year = {2016},
	pages = {1451--1456},
}

@misc{young2026interpolation,
      title={Interpolation of GEDI Biomass Estimates with Calibrated Uncertainty Quantification}, 
      author={Robin Young and Srinivasan Keshav},
      year={2026},
      eprint={2601.16834},
      archivePrefix={arXiv},
      primaryClass={cs.LG},
      url={https://arxiv.org/abs/2601.16834}, 
}

@article{birch2023,
	title = {Large-diameter trees and deadwood correspond with belowground ectomycorrhizal fungal richness},
	volume = {12},
	issn = {2192-1709},
	url = {https://ecologicalprocesses.springeropen.com/articles/10.1186/s13717-022-00415-8},
	doi = {10.1186/s13717-022-00415-8},
	language = {en},
	number = {1},
	journal = {Ecological Processes},
	author = {Birch, Joseph D. and Lutz, James A. and Struckman, Soren and Miesel, Jessica R. and Karst, Justine},
	month = jan,
	year = {2023},
	pages = {3},
}

@article{tedersoo2010,
	title = {Low diversity and high host preference of ectomycorrhizal fungi in {Western} {Amazonia}, a neotropical biodiversity hotspot},
	volume = {4},
	copyright = {https://academic.oup.com/pages/standard-publication-reuse-rights},
	issn = {1751-7362, 1751-7370},
	url = {https://academic.oup.com/ismej/article/4/4/465/7588015},
	doi = {10.1038/ismej.2009.131},
	language = {en},
	number = {4},
	journal = {The ISME Journal},
	author = {Tedersoo, Leho and Sadam, Ave and Zambrano, Milton and Valencia, Renato and Bahram, Mohammad},
	month = apr,
	year = {2010},
	pages = {465--471},
}

@misc{Bandara2023MixCoAM,
      title={Guarding Barlow Twins Against Overfitting with Mixed Samples},
      author={Wele Gedara Chaminda Bandara and Celso M. De Melo and Vishal M. Patel},
      year={2023},
      eprint={2312.02151},
      archivePrefix={arXiv},
      primaryClass={cs.CV},
      url={https://arxiv.org/abs/2312.02151},
}

@article{tomao2020how,
	title = {How does forest management affect fungal diversity and community composition? {Current} knowledge and future perspectives for the conservation of forest fungi},
	volume = {457},
	issn = {03781127},
	shorttitle = {How does forest management affect fungal diversity and community composition?},
	url = {https://linkinghub.elsevier.com/retrieve/pii/S0378112719313234},
	doi = {10.1016/j.foreco.2019.117678},
	language = {en},
	journal = {Forest Ecology and Management},
	author = {Tomao, Antonio and Antonio Bonet, José and Castaño, Carles and de-Miguel, Sergio},
	month = feb,
	year = {2020},
	pages = {117678},
}

@article{gann2019international,
	title = {International principles and standards for the practice of ecological restoration. {Second} edition},
	volume = {27},
	issn = {1061-2971, 1526-100X},
	url = {https://onlinelibrary.wiley.com/doi/10.1111/rec.13035},
	doi = {10.1111/rec.13035},
	language = {en},
	number = {S1},
	journal = {Restoration Ecology},
	author = {Gann, George D. and McDonald, Tein and Walder, Bethanie and Aronson, James and Nelson, Cara R. and Jonson, Justin and Hallett, James G. and Eisenberg, Cristina and Guariguata, Manuel R. and Liu, Junguo and Hua, Fangyuan and Echeverría, Cristian and Gonzales, Emily and Shaw, Nancy and Decleer, Kris and Dixon, Kingsley W.},
	month = sep,
	year = {2019},
}
\end{document}